\documentclass{article}

\usepackage{graphicx} 

\usepackage{amsfonts}       
\usepackage{nicefrac}       
\usepackage{microtype}      
\usepackage{wrapfig}
\usepackage{caption}
\usepackage{booktabs}

\usepackage{algorithm}
\usepackage{algorithmic}

\usepackage{enumitem}
\usepackage{hyperref}
\hypersetup{colorlinks, citecolor={blue}}

\usepackage{tcolorbox}

\usepackage{tikz}

\usepackage{fullpage}
\usepackage{authblk}
\usepackage[round]{natbib} 

\usepackage{Definitions}
\usepackage{color}

\newcommand{\AlgName}{Adversarial Dynamics Embedding}
\newcommand{\algname}{adversarial dynamics embedding}
\newcommand{\algshort}{ADE}

\renewcommand{\le}{\leqslant}

\renewcommand{\ge}{\geqslant}

\title{Exponential Family Estimation via~\AlgName}

\author{
  $^*$Bo Dai$^1$, \thanks{indicates equal contribution. Email: \texttt{\{bodai, hadai\}@google.com, zhen.liu.2@umontreal.ca}.}
  Zhen Liu$^2$, $^*$Hanjun Dai$^1$, Niao He$^3$, \\
  \vspace{-3mm}
  Arthur Gretton$^4$, Le Song$^{5,6}$, Dale Schuurmans$^{1,7}$\\ 
  \vspace{3mm}
  $^1$Google Research, Brain Team, $^2$Mila, University of Montreal,\\
  $^3$University of Illinois at Urbana Champaign, $^4$University College London,\\
  $^5$Georgia Institute of Technology, $^6$Ant Financial, $^7$University of Alberta
}

\begin{document}

\maketitle

\begin{abstract}
  We present an efficient algorithm for maximum likelihood estimation~(MLE) of  exponential family models, with a general parametrization of the energy function that includes neural networks. We exploit the primal-dual view of the MLE with a \emph{kinetics augmented model} to obtain an estimate associated with an \emph{adversarial} dual sampler. To represent this sampler, we introduce a novel neural architecture, \emph{dynamics embedding}, that generalizes Hamiltonian Monte-Carlo (HMC). The proposed approach inherits the flexibility of HMC while enabling tractable entropy estimation for the augmented model. By learning both a dual sampler and the primal model simultaneously, and sharing parameters between them, we obviate the requirement to design a separate sampling procedure once the model has been trained, leading to more effective learning.
  We show that many existing estimators, such as contrastive divergence, pseudo/composite-likelihood, score matching, minimum Stein discrepancy estimator, non-local contrastive objectives, noise-contrastive estimation, and minimum probability flow, are special cases of the proposed approach, each expressed by a different (fixed) dual sampler. An empirical investigation shows that adapting the sampler during MLE can significantly improve on state-of-the-art estimators\footnote{The code repository is available at~{\href{https://github.com/lzzcd001/ade-code}{https://github.com/lzzcd001/ade-code}}.}. 
\end{abstract}

\section{Introduction}\label{sec:intro}

The exponential family is one of the most important classes of distributions in statistics and machine learning, encompassing undirected graphical models~\citep{WaiJor08} and energy-based models~\citep{LecChoHadRanetal06,WuXieLuZhu18}, which include, for example, Markov random fields~\citep{KinSne80}, conditional random fields~\citep{LafMcCPer01} and language models~\citep{MniTeh12}. Despite the flexibility of this family and the many useful properties it possesses~\citep{Brown86}, most such distributions are intractable because the partition function does not possess an analytic form. This leads to difficulty in evaluating, sampling and learning exponential family models, hindering their application in practice. In this paper, we consider a longstanding question:
\begin{quotation}
  \noindent  
  \emph{Can a simple yet effective algorithm be developed for estimating general exponential family distributions?}
\end{quotation}

There has been extensive prior work addressing this question. Many approaches focus on approximating maximum likelihood estimation (MLE), since it is well studied and known to possess desirable statistical properties, such as consistency, asymptotic unbiasedness, and asymptotic normality~\citep{Brown86}. One prominent example is contrastive divergence~(CD)~\citep{Hinton02} and its variants~\citep{TieHin09,DuMor19}. It approximates the gradient of the log-likelihood by a stochastic estimator that uses samples generated from a few Markov chain Monte Carlo~(MCMC) steps. This approach has two shortcomings: first and foremost, the stochastic gradient is \emph{biased}, which can lead to poor estimates; second, CD and its variants require careful design of the MCMC transition kernel, which can be challenging.

Given these difficulties with MLE, numerous learning criteria have been proposed to avoid the partition function. Pseudo-likelihood estimators~\citep{Besag75} approximate the joint distribution by the product of conditional distributions, each of which only represents the distribution of a single random variable conditioned on the others. However, the the partition function of each factor is still generally intractable.
Score matching~\citep{Hyvarinen05} minimizes the Fisher divergence between the empirical distribution and the model. Unfortunately, it requires third order derivatives for optimization, which becomes prohibitive for large models~\citep{KinLec10,LiSutStrGre18}. 
Noise-contrastive estimation~\citep{GutHyv10} recasts the problem as ratio estimation between the target distribution and a pre-defined auxiliary distribution. However, the auxiliary distribution must cover the support of the data with an analytical expression that still allows efficient sampling; this requirement is difficult to satisfy in practice, particularly in high dimensional settings.
Minimum probability flow~\citep{SohBatDew11} exploits the observation that, ideally, the empirical distribution will be the stationary distribution of transition dynamics defined under an optimal model. The model can then be estimated by matching these two distributions. Even though this idea is inspiring, it is challenging to construct appropriate dynamics that yield efficient learning.

In this paper, we introduce a novel algorithm, \emph{\AlgName~(\algshort)}, that directly approximates the MLE while achieving computational and statistical efficiency. Our development starts with the \emph{primal-dual} view of the MLE~\citep{DaiDaiGreSon18} that provides a natural objective for jointly learning both a sampler and a model, as a remedy for the expensive and biased MCMC steps in the CD algorithm. To parameterize the dual distribution, \citet{DaiDaiGreSon18} applies a naive transport mapping, which makes entropy estimation difficult and requires learning an extra auxiliary model, incurring additional computational and memory cost. 

We overcome these shortcomings by considering a different approach, inspired by the properties of Hamiltonian Monte-Carlo~(HMC)~\citep{Neal11}: 
\begin{itemize}
 \item[{\bf i)}] HMC forms a stationary distribution with \emph{independent} potential and kinetic variables;
 \item[{\bf ii)}] HMC can approximate the exponential family \emph{arbitrarily closely}. 
\end{itemize}
As in HMC, we consider an \emph{augmented model} with latent kinetic variables in~\secref{subsec:argumented_mle}, and introduce a novel neural architecture in~\secref{subsec:dyna_embed}, called~\emph{dynamics embedding}, that mimics sampling and represents the dual distribution via parameters of the primal model. This approach shares with HMC the advantage of a \emph{tractable} entropy function for the augmented model, while enriching the flexibility of sampler without introducing extra parameters. In~\secref{subsec:coupled_learning} we develop a $\max$-$\min$ objective that allows the shared parameters in primal model and dual sampler to be learned simultaneously, which improves computational and sample efficiency. We further show that the proposed estimator subsumes CD, pseudo-likelihood, score matching, non-local contrastive objectives, noise-contrastive estimation, and minimum probability flow as special cases with hand-designed dual samplers in~\secref{sec:related_work}. Finally, in~\secref{sec:experiments} we find that the proposed approach can outperform current state-of-the-art estimators in a series of experiments.

\section{Preliminaries}\label{sec:prelim}

We provide a brief introduction to the technical background that is needed in the derivation of the new algorithm, including exponential family, dynamics-based MCMC sampler, and primal-dual view of MLE.  

\subsection{Exponential Family and Energy-based Model}

The natural form of the exponential family over $\Omega\subset\RR^d$ is defined as 
\begin{equation}\label{eq:exp_family}
\textstyle
p_{f'}\rbr{x} = \exp\rbr{f'(x)-\log p_0\rbr{x} - A_{p_0}\rbr{f'}},\,\, A_{p_0}\rbr{f'} \defeq \log\int_\Omega \exp\rbr{f'\rbr{x} }p_0\rbr{x}dx,
\end{equation}
where $f'\rbr{x} = w^\top \phi_{\varpi}\rbr{x}$. The sufficient statistic $\phi_{\varpi}\rbr{\cdot}:\Omega\rightarrow \RR^k$ can be any general parametric model, \eg, a neural network. The $\rbr{w, \varpi}$ are the parameters to be learned from observed data. The exponential family definition~\eq{eq:exp_family} includes the energy-based model~\citep{LecChoHadRanetal06} as a special case, by setting $f'\rbr{x} = \phi_{\varpi}\rbr{x}$ with $k=1$, which has been generalized to the infinite dimensional case~\citep{SriFukGreHyvetal17}. The $p_0\rbr{x}$ is fixed and covers the support $\Omega$, which is usually unknown in practical high-dimensional problems. Therefore, we focus on learning $f\rbr{x} = f'\rbr{x} - \log p_0\rbr{x}$ jointly with $p_0\rbr{x}$, which is more difficult: in particular, the doubly dual embedding approach~\citep{DaiDaiGreSon18} is no longer applicable.

Given a sample $\Dcal = \sbr{x_i}_{i=1}^N$ and denoting $f\in\Fcal$ as the valid parametrization family, an exponential family model can be estimated by maximum log-likelihood, \ie,
\begin{equation}\label{eq:mle}
\textstyle
\max_{f\in\Fcal}\,\, L\rbr{f} \defeq \widehat\EE_{\Dcal}\sbr{f\rbr{x}} - A\rbr{f}, \,\, A\rbr{f} =  \log\int_\Omega \exp\rbr{f\rbr{x} } dx,
\end{equation}
with gradient $\nabla_f L\rbr{f} = \widehat\EE_{\Dcal}\sbr{\nabla_f f\rbr{x}}- \EE_{p_f\rbr{x}}\sbr{\nabla_f f\rbr{x}}$. Since $A\rbr{f}$ and $\EE_{p_f\rbr{x}}\sbr{\nabla_f f\rbr{x}}$ are both intractable, solving the MLE for a general exponential family model is very difficult.

\subsection{Dynamics-based MCMC}

Dynamics-based MCMC is a general and effective tool for sampling. The idea is to represent the target distribution as the solution to a set of (stochastic) differential equations, which allows samples from the target distribution to be obtained by simulating along the dynamics defined by the differential equations. 

Hamiltonian Monte-Carlo~(HMC)~\citep{Neal11} is a representative algorithm in this category, which exploits the well-known Hamiltonian dynamics. Specifically, given a target distribution $p_f\rbr{x} \propto \exp\rbr{f\rbr{x}}$, the Hamiltonian is defined as
$\Hcal\rbr{x, v} = -f\rbr{x} + k\rbr{v}$, 
where $k\rbr{v} = \frac{1}{2}v^\top v$ is the kinetic energy. The Hamiltonian dynamics generate $\rbr{x, v}$ over time $t$ by following
\begin{equation}\label{eq:ham_dynamics}
\sbr{\frac{dx}{dt}, \frac{dv}{dt}} = \sbr{\partial_v \Hcal\rbr{x, v}, -\partial_x \Hcal\rbr{x, v}} = \sbr{v, \nabla_x f\rbr{x}}.
\end{equation}
Asymptotically as $t\rightarrow \infty$, $x$ visits the underlying space according to the target distribution. In practice, to reduce discretization error, an acceptance-rejection step is introduced. 
The finite-step dynamics-based MCMC sampler can be used for approximating $\EE_{p_f\rbr{x}}\sbr{\nabla_f f\rbr{x}}$ in $\nabla_f L\rbr{f}$, which leads to the CD algorithm~\citep{Hinton02,ZhuMum98}. 

\subsection{The Primal-Dual View of MLE}\label{subsec:primal_dual}

The Fenchel duality of $A\rbr{f}$ has been exploited~\citep{Rockafellar70,WaiJor08,DaiDaiGreSon18} as another way to address the intractability of the $\log$-partition function.
\begin{theorem}[Fenchel dual of log-partition~\citep{WaiJor08}]
\label{thm:fenchel_log_partition}
Denote the entropy $H\rbr{q}$ as $-\int_\Omega q\rbr{x}\log{q\rbr{x}}dx$, then:
\begin{eqnarray}\label{eq:opt_dual}
A\rbr{f} &=& \max_{q\in \Pcal}\,\, \inner{q(x)}{f\rbr{x}} + H\rbr{q},\\
p_f\rbr{x} &=& \argmax_{q\in \Pcal}\,\, \inner{q(x)}{f(x)} + H\rbr{q},
\end{eqnarray}
where $\Pcal$ denotes the space of distributions and $\inner{f}{g} = \int_\Omega f\rbr{x}g\rbr{x}dx$. 
\end{theorem}
Plugging the Fenchel dual of $A\rbr{f}$ into the MLE~\eqref{eq:mle}, we arrive at a $\max$-$\min$ reformulation
\begin{eqnarray}\label{eq:primal_CD}
\max_{f\in\Fcal}\min_{q\in\Pcal}\,\, \widehat\EE_{\Dcal}\sbr{f(x)} - \EE_{q(x)}\sbr{f(x)} - H\rbr{q},
\end{eqnarray}
which bypasses the explicit computation of the partition function. Another byproduct of the primal-dual view is that the dual distribution can be used for inference, however in vanila estimators this usually requires expensive sampling algorithms.

The dual sampler $q\rbr{\cdot}$ plays a vital role in the primal-dual formulation of the MLE in~\eqref{eq:primal_CD}. To achieve better performance, we have several principal requirements in parameterizing the dual distribution:
\begin{itemize}
  \item[{\bf i)}] the parametrization family needs to be \emph{flexible} enough to achieve small error in solving the inner minimization problem; 
  \item[{\bf ii)}] the entropy of the parametrized dual distribution should be \emph{tractable}. 
\end{itemize}
Moreover, as shown in~\eqref{eq:opt_dual} in~\thmref{thm:fenchel_log_partition}, the optimal dual sampler $q\rbr{\cdot}$ is determined by primal potential function $f\rbr{\cdot}$. This leads to the third requirement: 
\begin{itemize}
  \item[{\bf iii)}] the parametrized dual sampler should \emph{explicitly incorporate} the primal model $f$. 
\end{itemize}
Such a dependence can potentially reduce both the memory and learning sample complexity.  

A variety of techniques have been developed for distribution parameterization, such as reparametrized latent variable models~\citep{KinWel13,RezMohWie14},
transport mapping~\citep{GooPouMirXuetal14}, and normalizing flow~\citep{RezMoh15,DinSohBen16,KinSalJozCheetal16}. However, none of these satisfies the requirements of flexibility and a tractable density simultaneously, nor do they offer a principled way to couple the parameters of the dual sampler with the primal model.

\section{\AlgName}\label{sec:mle}

By augmenting the original exponential family with kinetic variables, we can parametrize the dual sampler with a \emph{dynamics embedding} that satisfies all three requirements without effecting the MLE, allowing the primal potential function and dual sampler to both be trained adversarially. We start with the embedding of classical Hamiltonian dynamics~\citep{Neal11,CatDouSej18} for the dual sampler parametrization, as a concrete example, then discuss its generalization in latent space and the stochastic Langevin dynamics embedding. This technique is extended to other dynamics, with their own advantages, in \appref{appendix:variants}.

\subsection{Primal-Dual View of Augmented MLE}\label{subsec:argumented_mle}

As noted, it is difficult to find a parametrization of $q\rbr{x}$ in~\eqref{eq:primal_CD} that simultaneously satisfies all three requirements. Therefore, instead of directly tackling~\eqref{eq:primal_CD} in the original model, and inspired by HMC, we consider the augmented exponential family $p\rbr{x, v}$ with an auxiliary momentum variable, \ie,
\begin{eqnarray}\label{eq:joint_hmc}
p\rbr{x, v} = \frac{\exp\rbr{f\rbr{x} - \frac{\lambda}{2} v^\top v}}{Z\rbr{f}}, \quad Z\rbr{f} = \int \exp\rbr{f\rbr{x} - \frac{\lambda}{2} v^\top v} dx dv. 
\end{eqnarray}
The MLE of such a model can be formulated as
\begin{equation}\label{eq:joint_MLE}
\max_{f}L\rbr{f}\defeq\widehat\EE_{x\sim\Dcal}\sbr{\log\int p\rbr{x, v} dv} 
=\widehat\EE_{x\sim\Dcal}\EE_{p\rbr{v|x}}\sbr{f\rbr{x} - \frac{\lambda}{2}{v}^\top v - \log p\rbr{v|x}} - \log Z\rbr{f} 
\end{equation}
where the last equation comes from true posterior $p\rbr{v|x} = \Ncal\big(0,\lambda^{-\frac{1}{2}}I\big)$ due to the independence of $x$ and $v$. This independence also induces the equivalent MLE as proved in~\appref{appendix:derivation_sec_mle}.
\begin{theorem}[Equivalent MLE]\label{thm:equivalent_mle}
The MLE of the augmented model is the same as the original MLE. 
\end{theorem}
Applying the Fenchel dual to $Z\rbr{f}$ of the augmented model~\eqref{eq:joint_hmc}, we derive a primal-dual formulation of~\eqref{eq:joint_MLE}, leading to the objective,
\begin{eqnarray}\label{eq:primal_dual_joint_MLE}
L\rbr{f}\propto \min_{q\rbr{x, v}\in\Pcal}\,\,\widehat\EE_{x\sim\Dcal}\sbr{f\rbr{x}} - \EE_{q\rbr{x, v}}\sbr{f\rbr{x} - \frac{\lambda}{2}{v}^\top v - \log q\rbr{x, v}}.
\end{eqnarray}
The $q\rbr{x, v}$ in~\eqref{eq:primal_dual_joint_MLE} contains momentum $v$ as the latent variable. One can also exploit the latent variable model for $q\rbr{x} = \int q\rbr{x|v}q\rbr{v}dv$ in~\eqref{eq:primal_CD}. However, the $H\rbr{q}$ in~\eqref{eq:primal_CD} requires marginalization, which is intractable in general, and usually estimated through variational inference with the introduction of an extra posterior model $q\rbr{v|x}$. Instead, by considering the specifically designed augmented model, ~\eqref{eq:primal_dual_joint_MLE} eliminates these extra variational steps. 

Similarly, one can consider the latent variable augmented model with multiple momenta, \ie,
$
p\rbr{x, \cbr{v^i}_{i=1}^T} = \frac{\exp\rbr{f\rbr{x} - \sum_{i=1}^T \frac{\lambda_i}{2} \nbr{v^i}_2^2 }}{Z\rbr{f}},
$
leading to the optimization
\begin{equation}\label{eq:primal_dual_multi_MLE}
\scalebox{0.86}
{
$L\rbr{f} \propto \min_{q\rbr{x, \cbr{v^i}_{i=1}^T}\in\Pcal}\widehat\EE_{x\sim\Dcal}\sbr{f\rbr{x}} 
- \EE_{q\rbr{x, \cbr{v^i}_{i=1}^T}}\sbr{f\rbr{x} - \sum_{i=1}^T \frac{\lambda_i}{2} \nbr{v^i}_2^2 - \log q\rbr{x, \cbr{v^i}_{i=1}^T}}.$
}
\end{equation}

\subsection{Representing Dual Sampler via Primal Model}\label{subsec:dyna_embed}

We now introduce the Hamiltonian dynamics embedding to represent the dual sampler $q\rbr{\cdot}$, as well as its generalization and special instantiation that satisfy all three of the principal requirements.

The vanilla HMC is derived by discretizing the Hamiltonian dynamics~\eqref{eq:ham_dynamics} with a leapfrog integrator. Specifically, in a single time step, the sample $\rbr{x, v}$ moves towards $\rbr{x', v'}$ according to
\begin{equation}\label{eq:leapfrog}
\textstyle
\rbr{x', v'} = \Lb_{f, \eta}\rbr{x,v} \defeq
\rbr{\begin{array}{c}
v^{\frac{1}{2}} = v+ \frac{\eta}{2}\nabla_x f\rbr{x} \\
\quad x' = x + \eta v^{\frac{1}{2}}\\
\quad v' = v^{\frac{1}{2}} + \frac{\eta}{2}\nabla_x f\rbr{x'}
\end{array}},
\end{equation}
where $\eta$ is defined as the leapfrog stepsize. Let's denote the one-step leapfrog as $\rbr{x',v'} = \Lb_{f, \eta}\rbr{x, v}$ and assume the $\rbr{x^0, v^0}\sim q_{\theta}^0\rbr{x, v}$. After $T$ iterations, we obtain
\begin{equation}\label{eq:hmc_nn}
\rbr{x^T, v^T} = \Lb_{f, \eta}\circ\Lb_{f, \eta}\circ\ldots\circ\Lb_{f, \eta}\rbr{x^0, v^0}
.
\end{equation} 
Note that this can be viewed as a neural network with a special architecture, which we term \emph{Hamiltonian~(HMC) dynamics embedding}. Such a representation explicitly characterizes the dual sampler by the primal model, \ie, the potential function $f$, meeting the  dependence requirement.

The flexibility of the distributions  HMC embedding actually is ensured by the nature of the dynamics-based samplers. In the limiting case, the proposed neural network~\eqref{eq:hmc_nn} reduces to a gradient flow, whose stationary distribution is exactly the model distribution:
$$
p\rbr{x,v} = \argmax_{q\rbr{x, v}\in\Pcal}\, \EE_{q\rbr{x, v}}\sbr{f\rbr{x} - \frac{\lambda}{2}{v}^\top v - \log q\rbr{x, v}}.
$$
The approximation strength of the HMC embedding is formally justified as follows: 
\begin{theorem}[HMC embeddings as gradient flow]\label{thm:grad_flow}
In continuous time, \ie\ with infinitesimal stepsize $\eta\rightarrow 0$, the density of particles $\rbr{x^t, v^t}$, denoted $q^t\rbr{x, v}$, follows the Fokker-Planck equation
\begin{eqnarray}
\textstyle
\frac{\partial q^t\rbr{x, v}}{\partial t} = \nabla\cdot\rbr{q^t\rbr{x, v}G\nabla \Hcal\rbr{x, v}}, \\[-5mm]
\nonumber
\end{eqnarray}
with $\textstyle G = \begin{bmatrix} 0 & \Ib\\
-\Ib & 0\end{bmatrix}$, which has a stationary distribution $p\rbr{x, v}\propto \exp\rbr{-\Hcal\rbr{x, v}}$ with the marginal distribution $p(x)\propto\exp\rbr{f(x)}$.
\end{theorem}
Details of the proofs are given in~\appref{appendix:derivation_sec_mle}. Note that this stationary distribution result is an instance of the more general dynamics described in~\citet{MaCheFox15}, showing the flexility of the induced distributions. As demonstrated in~\thmref{thm:grad_flow}, the neural parametrization formed by the HMC embedding is able to well approximate an exponential family distribution on continuous variables.

\paragraph{Remark (Generalized HMC dynamics in latent space)} 
The leapfrog operation in vanilla HMC works directly in the original observation space, which could be high-dimensional and noisy. We generalize the leapfrog update rule to the latent space and form a new dynamics as follows, 
\begin{equation}\label{eq:generalized_leapfrog}
\textstyle
\resizebox{0.8\textwidth}{!}
{$
\rbr{x',v'} = \Lb_{f, \eta, S, g}\rbr{x, v} \defeq 
\rbr{\begin{array}{c}
v^{\frac{1}{2}} = v\odot \exp\rbr{S_v\rbr{\nabla_x f\rbr{x}, x}} + \frac{\eta}{2}g_v\rbr{\nabla_x f\rbr{x}, x} \\
x' = x\odot \exp\rbr{S_x\rbr{v^{\frac{1}{2}}}} +\eta g_x\rbr{v^{\frac{1}{2}}}\\
v' = v^{\frac{1}{2}}\odot \exp\rbr{S_v\rbr{\nabla_x f\rbr{x'}, x'}} + \frac{\eta}{2}g_v\rbr{\nabla_x f\rbr{x'}, x'}
\end{array}},
$}
\end{equation}
where $v\in \RR^l$ denote the momentum evolving space and $\odot$ denotes element-wise product. Specifically, the terms $S_v\rbr{\nabla_x f\rbr{x}, x}$ and $S_x\rbr{v^{\frac{1}{2}}}$ rescale $v$ and $x$ coordinatewise. The term $g_v\rbr{\nabla_x f\rbr{x}, x}\mapsto \RR^l$ can be understood as projecting the gradient information to the essential latent space where the momentum is evolving. Then, for updating $x$, the latent momentum is projected back to original space via $g_x\rbr{v^{\frac{1}{2}}}\mapsto\Omega$. With these generalized leapfrog updates, the dynamical system avoids operating in the high-dimensional noisy input space, and becomes more computationally efficient. We emphasize that the proposed generalized leapfrog parametrization~\eqref{eq:generalized_leapfrog} is different from the one used in~\citet{LevHofSoh17}, which is inspired from the real-NVP flow~\citep{DinSohBen16}. 

By the generalized HMC embedding~\eqref{eq:generalized_leapfrog}, we have a flexible layer $\rbr{x',v'} = \Lb_{f, \eta, S, g}\rbr{x, v}$, where $\rbr{S_v, S_x, g_v, g_x}$ will be learned in addition to the stepsize. Obviously, the classic HMC layer $\Lb_{f, \eta, M}\rbr{x, v}$ is a special case of $\Lb_{f, \eta, S, g}\rbr{x, v}$ by setting $\rbr{S_v, S_x}$ to zero and $\rbr{g_v, g_f}$ to  identity functions.

\paragraph{Remark (Stochastic Langevin dynamics)} The stochastic Langevin dynamics can also be recovered from the leapfrog step by resampling momentum in every step. Specifically, the sample $\rbr{x, \xi}$ moves according to 
\begin{equation}\label{eq:langevin}
\rbr{x', v'} = \Lb^{\xi}_{f, \eta}\rbr{x} \defeq
\rbr{\begin{array}{c}
v' = \xi+ \frac{\eta}{2}\nabla_x f\rbr{x} \\
\quad x' = x + v' \\
\end{array}},\text{~with~~} \xi\sim q_\theta\rbr{\xi}.
\end{equation}
Hence, stochastic Langevin dynamics resample $\xi$ to replace the momentum in leapfrog~\eqref{eq:leapfrog}, ignoring the accumulated gradients. By unfolding $T$ updates, we obtain
\begin{equation}\label{eq:langevin_nn}
\rbr{x^T, \cbr{v^i}_{i=1}^T} = \Lb_{f, \eta}^{\xi^{T-1}}\circ\Lb_{f, \eta}^{\xi^{T-2}}\circ\ldots\circ\Lb_{f, \eta}^{\xi^0}\rbr{x^0}
\end{equation}
as the derived neural network. Similarly, we can also generalize the stochastic Langevin updates $\Lb_{f, \eta}^{\xi}$ to a low-dimension latent space by introducing $g_v\rbr{\nabla_x f\rbr{x}, x}$ and $g_x\rbr{v'}$ correspondingly.

One of the major advantages of the proposed distribution parametrization is its density value is also tractable, leading to tractable entropy estimation in~\eqref{eq:primal_dual_joint_MLE} and~\eqref{eq:primal_dual_multi_MLE}. In particular, we have the following,
\begin{theorem}[Density value evaluation]\label{thm:hmc_density}
If $\rbr{x^0, v^0}\!\sim\! q^0_{\theta}\rbr{x, v}$, after $T$ vanilla HMC steps~\eqref{eq:leapfrog}, then 
\begin{equation}\label{eq:leap_frog_density}
\textstyle
q^T\rbr{x^T, v^T} = q^0_{\theta}\rbr{x^0, v^0}.
\end{equation}
For $\rbr{x^T, v^T}$ from the generalized leapfrog steps~\eqref{eq:generalized_leapfrog}, we have 
\begin{equation}\label{eq:general_leap_frog_density}
\textstyle
q^T\rbr{x^T, v^T} =q^0_{\theta}\rbr{x^0, v^0}\prod_{t=1}^T\rbr{\Delta_x\rbr{x^t}\Delta_v\rbr{v^t}},
\end{equation}
where $\Delta_x\rbr{x^t}$ and $\Delta_v\rbr{v^t}$ denote
\begin{equation}
\textstyle
\Delta_x\rbr{x^t} = \abr{\det\rbr{\diag\rbr{\exp\rbr{2S_v\rbr{\nabla_x f\rbr{x^t}, x^t}}}}}, 
\Delta_v\rbr{v^t} = \abr{\det\rbr{\diag\rbr{\exp\rbr{ S_x\rbr{v^{\frac{1}{2}}}}}}}.  
\end{equation}
For $\rbr{x^T, \cbr{v^i}_{i=1}^T}$ from the Langevin dynamics~\eqref{eq:langevin} with $\rbr{x^0, \cbr{\xi^i}_{i=0}^{T-1}}\sim q_\theta^0\rbr{x, \xi}\prod_{i=i}^{T-1} q_{\theta_i}\rbr{\xi}$, we have
\begin{equation}\label{eq:langevin_density}
\textstyle
q^T\rbr{x^T, \cbr{v^i}_{i=1}^T} = q_\theta^0\rbr{x^0, \xi^0}\prod_{i=1}^{T-1} q_{\theta_i}\rbr{\xi^i}.
\end{equation}
\end{theorem}
The proof of~\thmref{thm:hmc_density} can be found in~\appref{appendix:derivation_sec_mle}.

The proposed dynamics embedding satisfies all three requirements: it defines a flexible family of distributions with computable entropy; and couples the learning of the dual sampler with the primal model, leading to memory and sample efficient learning algorithms, as we introduce in next section.

\subsection{Coupled Model and Sampler Learning}\label{subsec:coupled_learning} 

\begin{algorithm}[t] 
\caption{MLE via~\AlgName~(\algshort)} \label{alg:mle_de}
  \begin{algorithmic}[1]
    \STATE Initialize $\Theta_1$ randomly, set length of steps $T$. 
    \FOR{iteration $k=1, \ldots, K$}
        \STATE Sample mini-batch $\cbr{x_i}_{i=1}^m$ from dataset $\Dcal$ and $\cbr{x^0_i, v^0_i}_{i=1}^m$ from $q_\theta^0\rbr{x, v}$.
        \FOR{iteration $t=1, \ldots, T$}
          \STATE Compute $\rbr{x^t, v^t} = \Lb\rbr{x^{t-1}, v^{t-1}}$ for each pair of $\cbr{x^0_i, v^0_i}_{i=1}^m$.
        \ENDFOR
      \STATE {\color{blue}\bf[Learning the sampler]} $\Theta_{k+1}=\Theta_k-\gamma_k\hat{\nabla}_{\Theta}\ell\rbr{f_k; \Theta_k}$
      \STATE {\color{blue}\bf [Estimating the exponential family]} $f_{k+1}=f_k+\gamma_k \hat{\nabla}_f \ell\rbr{f_k; \Theta_k}$. 
    \ENDFOR
  \end{algorithmic}
\end{algorithm}

By plugging the $T$-step Hamiltonian dynamics embedding~\eqref{eq:leapfrog} into the primal-dual MLE of the augmented model~\eqref{eq:primal_dual_joint_MLE} and applying the density value evaluation~\eqref{eq:leap_frog_density}, we obtain the proposed optimization, which learns primal potential $f$ and the dual sampler adversarially,
\begin{equation}\label{eq:vanilla_hmc_param}
\max_{f\in\Fcal}\min_\Theta \,\,\ell\rbr{f, \Theta}\defeq 
\widehat\EE_\Dcal\sbr{f}-\EE_{\rbr{x^0, v^0}\sim q_\theta^0\rbr{x, v}}\sbr{f\rbr{x^T} -\frac{\lambda}{2}\nbr{v^T}_2^2} - H\rbr{q_\theta^0}.
\end{equation}
Here $\Theta$ denotes the learnable components in the dynamics embedding, \eg, initialization $q_\theta^0$, the stepsize $\rbr{\eta}$ in the HMC/Langevin updates, and the adaptive part $\rbr{S_v, S_x, g_v, g_x}$ in the generalized HMC. The parametrization of the initial distribution is discussed in~\appref{appendix:practical}. Compared to the optimization in GANs~\citep{GooPouMirXuetal14,ArjChiBot17,DaiAlmBacHovetal17}, beside the reversal of $\min$-$\max$ in~\eqref{eq:vanilla_hmc_param}, the major difference is that our ``generator'' (the dual sampler) shares parameters with the ``discriminator'' (the primal potential function). In our formulation, the updates of the potential function automatically push the generator toward the target distribution, thus accelerating learning efficiency. Meanwhile, the tunable parameters in the dynamics embedding are learned adversarially, further promoting the efficiency of the dual sampler. These benefits will be empirically demonstrated in~\secref{sec:experiments}.

Similar optimization can be derived for generalized HMC~\eqref{eq:generalized_leapfrog} with density~\eqref{eq:general_leap_frog_density}. For the $T$-step stochastic Langevin dynamics embedding~\eqref{eq:langevin}, we apply the density value~\eqref{eq:langevin_density} to~\eqref{eq:primal_dual_multi_MLE}, which also leads to a $\max$-$\min$ optimization with multiple momenta.

We use stochastic gradient descent to estimate $f$ for the exponential families as well as the parameters of the dynamics embedding $\Theta$ adversarially. Note that since the generated sample $\rbr{x_f^T, v_f^T}$ depends on $f$, the gradient w.r.t.\ $f$ should also take these variables into account as back-propagation through time~(BPTT), \ie,
\begin{eqnarray}\label{eq:hmc_grad_f}
\nabla_f \ell\rbr{f; \Theta}= \widehat\EE_\Dcal\sbr{\nabla_f f\rbr{x}} - \EE_{q^0}\sbr{\nabla_f f\rbr{x^T}}- \EE_{q^0}\sbr{\nabla_x f\rbr{x^T}\nabla_f x^T + \lambda v^T\nabla_f v^T}.
\end{eqnarray}
We illustrate the MLE via HMC~\algname~in~\algref{alg:mle_de}. The same technique can be applied to alternative dynamics embeddings parametrized dual sampler as in~\appref{appendix:variants}. Considering the dynamics embedding as an \emph{adaptive} sampler that automatically learns w.r.t. different models and datasets, the updates for $\Theta$ can be understood as \emph{learning to sample}.

\section{Connections to Other Estimators}\label{sec:connection}

\begin{table}
\caption{(Fix) dual samplers used in alternative estimators. We denote $p_\Dcal$ as the empirical data distribution, $x_{-i}$ as $x$ without $i$-th coordinate, $p_n$ as the prefixed noise distribution, $\Tcal_f\rbr{x'|x}$ as the HMC/Langevin transition kernel, $T_{\Dcal, f}\rbr{x}$ as the Stein variational gradient descent, and $A\rbr{x, x'}$ as the acceptance ratio.}
\label{table:connections}
\begin{center}
\begin{tabular}{c|c}
\hline
Estimators  &Dual Sampler $q(x)$ \\
\hline
CD & $\int\prod_{i=1}^T\Tcal_f\rbr{x^i|x^{i-1}}A(x^i, x^{i-1})p_\Dcal\rbr{x_0}dx_0^{T-1}$\\
SM & $\int \Tcal_f\rbr{x'|x}p_\Dcal\rbr{x}dx$ with Taylar expansion\\
DSKD & $x' = T_{\Dcal, f}\rbr{x}$\\
PL & $q(x) = \frac{1}{d}\sum_{i=1}^d p_f(x_i|x_{-i})p_\Dcal(x_{-i})$\\
CL & $q(x) = \frac{1}{m}\sum_{i=1}^m p_f(x_{A_i}|x_{-A_i})p_\Dcal(x_{-A_i})$\\
& $\cbr{A_i}_{i=1}^m = d$ and $A_i\cap A_j = \emptyset$\\
NLCO & $\sum_{i=1}^m\int p_{\rbr{f, i}}\rbr{x}p\rbr{S_i|x'}p_\Dcal\rbr{x'}dx$\\
& $p_{\rbr{f, i}}\rbr{x} = \frac{\exp\rbr{f\rbr{x}}}{Z_i\rbr{f}}$, $x\in S_i$\\
MPF & $\int\Tcal_f\rbr{x'|x}\exp\rbr{\frac{1}{2}\rbr{f\rbr{x'} - f\rbr{x}}}p_\Dcal\rbr{x}dx$\\
NCE & $\rbr{\frac{1}{2}p_\Dcal + \frac{1}{2}p_n} \frac{\exp\rbr{f\rbr{x}}}{\exp\rbr{f\rbr{x}} + {p_n\rbr{x}}}$ \\
\hline
\end{tabular}

\end{center}
\end{table}
The primal-dual view of the MLE also allows us to establish connections between the proposed estimator, \emph{adversarial dynamics embedding} (\algshort), and existing approaches, including contrastive divergence~\citep{Hinton02}, pseudo-likelihood~(PL)~\citep{Besag75}, conditional composite likelihood~(CL)~\citep{Lindsay88}, score matching~(SM)~\citep{Hyvarinen05}, minimum (diffusion) Stein kernel discrepancy estimator~(DSKD)~\citep{BarBriDunGiretal19}, non-local contrastive objectives~(NLCO)~\citep{VicLinKol10}, minimum probability flow~(MPF)~\citep{SohBatDew11}, and noise-contrastive estimation~(NCE)~\citep{GutHyv10}. As summarized in~\tabref{table:connections}, these existing estimators can be recast as the special cases of~\algshort, by replacing the adaptive dual sampler with hand-designed samplers, which can lead to extra error and inferior solutions. We provide detailed derivations of the connections below.

\subsection{Connection to Contrastive Divergence}\label{subsec:connection_CD} 

The CD algorithm~\citep{Hinton02} is a special case of the proposed algorithm. By~\thmref{thm:fenchel_log_partition}, the optimal solution to the inner optimization is $p\rbr{x, v}\propto\exp\rbr{-\Hcal\rbr{x, v}}$. Applying Danskin's theorem~\citep{Bertsekas95}, the gradient of $L\rbr{f}$ w.r.t. $f$ is 
\begin{equation}\label{eq:true_grad}
\nabla_f L\rbr{f} = \widehat\EE_{\Dcal}\sbr{\nabla_f f(x)} - \EE_{p_f(x)}\sbr{\nabla_f f(x)}.
\end{equation}
To estimate the integral $\EE_{p_f}\sbr{\nabla_f f\rbr{x}}$, the CD algorithm approximates the negative term in~\eqref{eq:true_grad} stochastically with a finite MCMC step away from empirical data. 
 
In the proposed dual sampler, by setting $p_\theta^0\rbr{x}$ to be the empirical distribution and eliminating the sampling learning, the dynamic embedding will collapse to CD with $T$-HMC steps if we remove gradient through the sampler, \ie, ignoring the third term in~\eqref{eq:hmc_grad_f}. Similarly, the persistent CD~(PCD)~\citep{Tieleman08} and recent ensemble CD~\citep{DuMor19} can also be recast as special cases by setting the negative sampler to be MCMC with initial samples from previous model and ensemble of MCMC samplers, respectively.

From this perspective, the CD and PCD algorithms induce errors not only from the sampler, but also from the gradient back-propagation truncation. The proposed algorithm escapes these sources of bias by learning to sample, and by adopting true gradients, respectively. Therefore, the proposed estimator is expected to achieve better performance than CD as demonstrated in the empirical experiments~\secref{sec:real_exp}.

\subsection{Connection to Score Matching}\label{subsec:connection_sm} 
The score matching~\citep{Hyvarinen05} estimates the exponential family by minimizing the Fisher divergence, \ie, 
\begin{equation}\label{eq:score_matching}
L_{SM}\rbr{f}\defeq -\EE_{\Dcal}\sbr{\sum_{i=1}^d\rbr{\frac{1}{2}\rbr{\partial_i f\rbr{x}}^2} + \partial^2_i f\rbr{x}}.
\end{equation}
As explained in~\citet{Hyvarinen07}, the objective~\eqref{eq:score_matching} can be derived as the $2$nd-order Taylor approximation of the MLE with $1$-step Langevin Monte Carlo as the dual sampler. Specifically, the Langevin Monte Carlo generates samples via
\begin{eqnarray*}
x' = x + \frac{\eta}{2}\nabla_x f\rbr{x} + \sqrt{\eta}\xi,\quad \xi\sim\Ncal\rbr{0, I},
\end{eqnarray*}
then, a simple Taylor expansion gives
\begin{eqnarray*}
\log p_f\rbr{x'} = \log p_f\rbr{x} +\sum_{i=1}^d \partial_i f\rbr{x}\rbr{ \frac{\eta}{2}\partial_i f\rbr{x} + \sqrt{\eta}\xi_i}+ \eta\sum_{i, j=1}^d\xi_i\xi_j\partial_{ij}^2 f\rbr{x} + o\rbr{\eta}.
\end{eqnarray*}
Plug such into the negative expectation in $L\rbr{f}$, leading to 
\begin{eqnarray*}
L\rbr{f}\approx \widehat \EE_\Dcal\sbr{\log p_f\rbr{x} - \EE_{x'|x}\sbr{\log p_f\rbr{x'}}}\approx - \eta\EE_{\Dcal}\sbr{\sum_{i=1}^d\rbr{\frac{1}{2}\rbr{\partial_i f\rbr{x}}^2} + \partial^2_i f\rbr{x}}, 
\end{eqnarray*}
which is exactly the scaled $L_{SM}\rbr{f}$ defined in~\eqref{eq:score_matching}. 

Therefore, the score matching can be viewed as applying Taylor expansion approximation with fixed $1$-step Langevin sampler in our framework, which is compared in~\secref{sec:synthetic_exp}.

\subsection{Connection to Minimum Stein Discrepancy Estimator}\label{subsec:connection_MSDE} 
The minimum Stein discrepancy estimator~\citep{BarBriDunGiretal19} is obtained by minimizing the Stein discrepancy, including the diffusion kernel Stein discrepancy~(DKSD) and diffusion score matching. Without loss of the generality, for simplicity, we recast the DKSD with an identity diffusion matrix as a special approximation to the MLE.

The identity DKSD maximizes the following objective,
\begin{equation}\label{eq:kernel_stein}
L_{DKSD}\rbr{f}\defeq -\sup_{h\in \Hcal_k, \nbr{h}_{\Hcal_k}\le 1} \widehat\EE_\Dcal\sbr{\Scal_{f}h\rbr{x}} = - \widehat\EE_{x, x'\sim\Dcal}\sbr{\Scal_f\rbr{x, \cdot}\otimes_k\Scal_f\rbr{x', \cdot}}
\end{equation}
where $\Scal_fh\rbr{x}\defeq \inner{\Scal_f\rbr{x, \cdot}}{h} = \inner{\nabla_x f\rbr{x}^\top k\rbr{x, \cdot} + \nabla k\rbr{x, \cdot}}{h}$. 

In fact, the objective~\eqref{eq:kernel_stein} can be derived as the Taylor approximation of the MLE with Stein variational gradient descent~(SVGD) as the dual sampler. Specifically, the SVGD generates samples via
$$
x' = T_{\Dcal, f}\rbr{x}\defeq x+ \eta h^*_{\Dcal, f}\rbr{x},\quad x\sim p_\Dcal\rbr{x},
$$
where $h^*_{\Dcal, f}\rbr{\cdot} \propto \EE_{y\sim\Dcal}\sbr{\Scal_f\rbr{y, \cdot}}$. Then, by Taylor-expansion, we have
\begin{eqnarray*}
f\rbr{x'} = f\rbr{x} + \eta \nabla_x f^\top\rbr{x} h^*_{\Dcal, f}\rbr{x}+ o\rbr{\eta}.
\end{eqnarray*}

We apply the change-of-variable rule, leading to $q\rbr{x'} = p_\Dcal\rbr{x}\det\abr{\frac{\partial x}{\partial x'}}$, therefore,
\begin{eqnarray*}
\log q(x') &=& \log p_\Dcal\rbr{x} + \log\det\abr{\frac{\partial x}{\partial x'}}\\
&=& \log p_\Dcal\rbr{x} - \log\det\abr{\frac{\partial x'}{\partial x}}\\
&=& \log p_\Dcal\rbr{x} - \log\det\abr{ I + \eta \nabla_x h^*_{\Dcal}\rbr{x}}\\
&\approx& \log p_\Dcal\rbr{x} - \eta\tr\rbr{ \nabla_x h^*_{\Dcal}\rbr{x}},
\end{eqnarray*}
where the last equation comes from Taylor expansion.

Plug these into the primal-dual view of MLE~\eqref{eq:primal_CD} with the fixed SVGD dual sampler, we have
\begin{eqnarray*}
L\rbr{f}&\approx& \widehat\EE_{x\sim\Dcal}\sbr{f\rbr{x} - f\rbr{x'} + \log q\rbr{x'}}\\
& =& \widehat\EE_{x\sim\Dcal}\sbr{ - \eta \nabla_x f^\top\rbr{x} h^*_{\Dcal, f}\rbr{x}  - \eta\tr\rbr{ \nabla_x h^*_{\Dcal}\rbr{x}}} + \widehat\EE_{x\sim\Dcal}\sbr{\log p_\Dcal\rbr{x}}  + o\rbr{\eta}\\
& = & -\eta \underbrace{\widehat\EE_{x, x'\sim\Dcal}\sbr{\Scal_f\rbr{x, \cdot}\otimes_k\Scal_f\rbr{x', \cdot}}}_{L_{DSKD}\rbr{f}} + \texttt{const} + o\rbr{\eta},
\end{eqnarray*}
which is the scaled $L_{DSKD}\rbr{f}$ defined in~\eqref{eq:kernel_stein}.

Therefore, the (diffusion) Stein kernel estimator can be viewed as Taylor expansion with fixed $1$-step Stein variational gradient descent dual sampler in our framework.

\subsection{Connection to Pseudo-Likelihood and Conditional Composite Likelihood}\label{subsec:connection_pl} 
The pseudo-likelihood estimation~\citep{Besag75} is a special case of the proposed algorithm by restricting the parametrization of the dual distribution. Specifically, denote the $p_f\rbr{x_i|x_{-i}}=\frac{\exp\rbr{f\rbr{x_i, x_{-i}}}}{Z\rbr{x_{-i}}}$ with $Z\rbr{x_{-i}}\defeq \int \exp\rbr{f\rbr{x_i, x_{-i}}} d x_i$, instead of directly maximizing likelihood, the pseudo-likelihood estimator is maximizing
\begin{equation}\label{eq:psuedo_likelihood}
L_{PL}\rbr{f}\defeq \widehat\EE_{\Dcal}\sbr{\sum_{i=1}^d \log p_f\rbr{x_i|x_{-i}}}.
\end{equation}
Then, the $f$ is updated by the following the gradient of $L_{pl}\rbr{f}$, \ie, 
\begin{eqnarray}\label{eq:pl_grad}
\nabla_f L_{PL}\rbr{f} \propto\widehat\EE_{\Dcal}\sbr{\nabla_{f}f\rbr{x}} - \EE_{i\sim \Ucal\rbr{d}}\widehat\EE_{x_{-i}}\EE_{p_f\rbr{x_i|x_{-i}}}\sbr{\nabla_f f\rbr{x_i, x_{-i}}}.\nonumber
\end{eqnarray}
The pseudo-likelihood estimator can be recast as a special case of the proposed framework if we fix the dual sampler as {\bf i)}, sample $i\in \cbr{1, \ldots, d}$ uniformly; {\bf ii)}, sample $x\sim\Dcal$ and mask $x_i$; {\bf iii)}, sample $x_i\sim p_f\rbr{x_i|x_{-1}}$ and compose $\rbr{x_i, x_{-i}}$.

The conditional composite likelihood~\citep{Lindsay88} is a generalization of pseudo-likelihood by maximizing 
\begin{equation}\label{eq:composite_likelihood}
L_{CL}\rbr{f}\defeq \widehat\EE_{\Dcal}\sbr{\sum_{A_i=1}^m \log p_f\rbr{x_{A_i}|x_{-A_i}}},
\end{equation}
where $\cbr{A_i}_{i=1}^m = d$ and $A_i\cap A_j = \emptyset$. Similarly, the composite likelihood is updating with prefixed conditional block sampler for negative sampling. 

Same as CD, the prefixed sampler and the biased gradient in pseudo-likelihood and composite likelihood estimator will induce extra errors and lead to inferior solution. Moreover, the pseudo-likelihood may not applicable to the general exponential family with continuous variables, whose conditional distribution is also intractable.

\subsection{Connection to Non-local Contrastive Objectives}\label{subsec:connection_NLCO} 

The non-local contrastive estimator~\citep{VicLinKol10} is obtained by maximizing
\begin{equation}\label{eq:nonlocal}
L_{NCO}\rbr{f} \defeq \widehat\EE_{\Dcal}\sbr{\sum_{i=1}^m w\rbr{x, S_i}\rbr{f\rbr{x} - \log Z_i\rbr{f}}},
\end{equation}
where $\sbr{S_i}_{i=1}^m$ denotes some prefixed partition of $\Omega$, $Z_i\rbr{f} = \int_{x\in S_i} \exp\rbr{f\rbr{x}}dx$, and $w\rbr{x, S_i} = P\rbr{x\in S_i | x}$ with $\sum_{i=1}^m w\rbr{x, S_i} = 1$. The objective~\eqref{eq:nonlocal} leads to the update direction as
\begin{equation}\label{eq:nonlocal_grad}
\nabla_f L_{NCO}\rbr{f} =\widehat\EE_{\Dcal}\sbr{\nabla_f f\rbr{x}} - \EE_{q_f\rbr{x}}\sbr{\nabla_f f},
\end{equation}
where $q_f\rbr{x} = \sum_{i=1}^m\int p_{\rbr{f, i}}\rbr{x}w\rbr{x', S_i}p_\Dcal\rbr{x'}dx'$ with $p_\Dcal$ as the empirical distribution and $p_{\rbr{f, i}}\rbr{x} = \frac{\exp\rbr{f\rbr{x}}}{Z_i\rbr{f}}$, $x\in S_i$. Therefore, the non-local contrastive objective is a special case of the proposed framework with the dual sampler as {\bf i)}, sample $x'$ uniformly from $\Dcal$; {\bf ii)}, sample $S_i$ conditional on $x'$ according to $w\rbr{x, S_i}$; {\bf iii)}, sample $x_i\sim p_{\rbr{f, i}}\rbr{x}$ within $S_i$. Such negative sampling method is also not applicable to the general exponential family with continuous variables.

\subsection{Connection to Minimum Probability Flow}\label{subsec:connection_MPF} 

In the continuous state model, the minimum probability flow~\citep{SohBatDew11} estimates the exponential family by maximizing
\begin{eqnarray*}\label{eq:mpf_obj}
L_{MPF}\rbr{f}\defeq - \widehat\EE_{x\sim\Dcal}\EE_{x'\sim \Tcal_f\rbr{x'|x}}\sbr{\exp\rbr{\frac{1}{2}\rbr{f\rbr{x'} - f\rbr{x}}}},
\end{eqnarray*}
where $\Tcal_f$ is a \emph{hand-designed} {symmetric} transition kernel based on the potential function $f\rbr{x}$, \eg, Hamiltonian or Langevin simulation. Then, the MPF update direction can be rewritten as 
\begin{equation}\label{eq:mpf_grad}
\widehat\EE_{x\sim\Dcal}\EE_{x'\sim \Gamma\rbr{x'|x}}\sbr{{\nabla_f f\rbr{x} - \nabla_f f\rbr{x'} - \nabla_x f\rbr{x'} \nabla_f x'}}.
\end{equation}
where $\Gamma\rbr{x'|x} \defeq \Tcal_f{\rbr{x'|x}}\exp\rbr{\frac{1}{2}\rbr{f\rbr{x'} - f\rbr{x}}}$. The probability flow operator $\Gamma\rbr{x'|x}$ actually defines a Markov chain sampler that achieves the following balance equation, 
$$
\Gamma\rbr{x'|x}p_f\rbr{x} = \Gamma\rbr{x|x'}p_f\rbr{x'}.
$$
Similar to CD and score matching, the MPF exploits the $1$-step MCMC. Moreover, the gradient in MPF also considers the effects in sampler as the third term in~\eqref{eq:mpf_grad}. Therefore, the MPF can be recast as a special case of our algorithm with the prefixed dual sampler as $x\sim\Dcal$ and $x'\sim\Gamma\rbr{x'|x}$.

\subsection{Connection to Noise-Contrastive Estimator}\label{subsec:connection_nce} 

Instead of directly estimating the $f$ in the exponential family,~\citet{GutHyv10} propose the noise-contrastive estimation~(NCE) for the density ratio between the exponential family and some user defined reference distribution $p_n\rbr{x}$, from which the parameter $f$ can be reconstructed. Specifically, the NCE considers an alternative representation of exponential family distribution as $p_f\rbr{x} = \exp\rbr{f\rbr{x}}$, which explicitly enforces $\int \exp\rbr{f\rbr{x}}dx = 1$. The NCE is obtained by maximizing
\begin{equation}\label{eq:nce_obj}
L_{NCE}\rbr{f}\defeq \widehat\EE_{\Dcal}\sbr{\log h\rbr{x}} + \EE_{p_n\rbr{x}}\sbr{\log\rbr{1-h\rbr{x}}},
\end{equation}
where $h\rbr{x} = \frac{\exp\rbr{f\rbr{x}}}{\exp\rbr{f\rbr{x}} + {p_n\rbr{x}}}$. Then, we have the gradient of $L_{NCE}\rbr{f}$ as
\begin{equation}\label{eq:nce_grad}
\nabla_f L_{NCE}\rbr{f} = \widehat\EE_{\Dcal}\sbr{\nabla_f f\rbr{x}} - \EE_{\frac{1}{2}p_\Dcal + \frac{1}{2}p_n}\sbr{h\rbr{x}\nabla_f f\rbr{x}}.
\end{equation}
The negative sampler in the~\eq{eq:nce_grad} can be understood as an approximate importance sampling algorithm where the proposal is $\frac{1}{2}p_\Dcal + \frac{1}{2}p_n$ and the reweighting part is $h\rbr{x}$. As the $\exp\rbr{f}$ approaching $p_\Dcal$, the $h\rbr{x}$ will approach the true ratio $\frac{\exp\rbr{f(x)}}{p_\Dcal + p_{n}\rbr{x}}$, and thus, the negative samples will converge to true model samples.

The NCE can be understood as learning an important sampler. However, the performance of NCE highly relies on the quality $h\rbr{x}$, \ie, the choice of $p_n\rbr{x}$. It is required to cover the support of $p_\Dcal\rbr{x}$, which is non-trivial in practical high-dimensional applications.

\section{Related Work}\label{sec:related_work}

Exploiting deep models for energy-based model estimation has been investigated in~\citet{KimBengio16, DaiAlmBacHovetal17,LiuWan17,DaiDaiGreSon18}. However, the parametrization of the dual sampler should both be flexible and tractable to achieve better performance. Existing work is limited in one aspect or another.~\citet{KimBengio16} parameterized the sampler via a deep directed graphical model, whose approximation ability is restrictive and the entropy is intractable.~\citet{DaiAlmBacHovetal17} proposed algorithms relying either on a heuristic approximation or a lower bound of the entropy, and requiring learning an extra auxiliary component besides the dual sampler.~\citet{DaiDaiGreSon18} applied the Fenchel dual representation twice to reformulate the entropy term, but the algorithm requires knowing a proposal distribution with the same support, which is impractical for high-dimensional data. By contrast, \algshort~achieves both sufficient flexibility and tractability by exploiting the augmented model and a novel parametrization within the primal-dual view.

One of our major contributions is learning a sampling strategy for the exponential family estimation through the primal-dual view of MLE. \algshort~also shares some similarity with meta learning for sampling~\citep{LevHofSoh17,FenWanLiu17,SonZhaErm17,GonLiHer18}, where the sampler is parametrized via a neural network and learned through certain objectives. The most significant difference lies in the ultimate goal: we focus on exponential family \emph{model estimation}, where the learned sampler \emph{assists} with this objective. By contrast, learning to sample techniques target on a sampler for a \emph{fixed} model. This fundamentally distinguishes \algshort~from methods that only learn samplers. Moreover,~\algshort~exploits an augmented model that yields tractable entropy estimation, which has not  been fully investigated in previous literature.

\section{Experiments}\label{sec:experiments}
In this section, we test ~\algshort~on several synthetic datasets in~\secref{sec:synthetic_exp} and real-world image datasets in~\secref{sec:real_exp}. The details of each experiment setting can be found in~\appref{sec:exp_details}.

\subsection{Synthetic experiments}\label{sec:synthetic_exp}

\begin{wraptable}{r}{0.5\textwidth}
\vspace{-4mm}
\centering
\caption{Comparison on synthetic data using maximum mean discrepancy (MMD $\times 1e^{-3}$).} \label{tab:synthetic_mmd}
\vspace{-2mm}
\resizebox{0.48\textwidth}{!}{%
\begin{tabular}{ccccc}
  \toprule
  Dataset & SM & NF & CD-$15$ & \algshort \\
  \midrule
  \texttt{2spirals} & 5.09 & 0.69 & -0.45 & {\bf -0.61} \\
  \texttt{Banana} & 8.10 & 0.88 & -0.31 &  {\bf -0.99} \\
  \texttt{circles} & 4.90 & 0.76 & -0.83 &  {\bf -1.13} \\
  \texttt{cos} & 10.36 & 0.91 & 7.15 & {\bf -0.55} \\
  \texttt{Cosine} & 8.34 & 2.15 & 0.78 & {\bf -1.09} \\
  \texttt{Funnel} & 13.07 & {\bf -0.92}  & -0.38 & -0.75 \\
  \texttt{swissroll} & 19.93 & 1.97 & 0.20 & {\bf -0.36} \\
  \texttt{line} & 10.28 & 0.39 & 10.5 & {\bf -1.30} \\
  \texttt{moons} & 41.34 & 0.80 & 2.21 & {\bf -1.10} \\
  \texttt{Multiring} & 2.01 & 0.30 & -0.38 & {\bf -1.02} \\
  \texttt{pinwheel} & 18.41 & 3.01 & {\bf -1.03} & -0.95 \\
  \texttt{Ring} & 9.22 & 161.89 & 0.12 & {\bf -0.91} \\
  \texttt{Spiral} & 9.48 & 5.96 & -0.41 & {\bf -0.81} \\
  \texttt{Uniform} & 5.88 & 0.00 & {\bf -1.17} & -0.94 \\
  \bottomrule
\end{tabular}
}
\vspace{-4mm}
\end{wraptable}
We compare~\algshort~with SM, CD, and primal-dual MLE with the normalizing planar flow~\citep{RezMoh15} sampler~(NF) to investigate the claimed benefits. SM, CD and primal-dual with NF can be viewed as special cases of our method, with either a fixed sampler or restricted parametrized $q_{\theta}$. Thus, this also serves as an ablation study of~\algshort~to verify the significance of its different subcomponents. We keep the model sizes the same in NF and ADE ($10$ planar layers). Then we perform $5$-steps stochastic Langevin steps to obtain the final  samples $x^T$ with standard Gaussian noise in each step, and without incurring extra memory cost. For fairness, we conduct CD with $15$ steps. This setup is preferable to CD with an extra acceptance-rejection step. We emphasize that, by comparison to SM and CD, ~\algshort~learns the sampler and exploits the gradients through the sampler. In comparison to primal-dual with NF, dynamics embedding achieves more flexibility without introducing extra parameters. Complete experiment details are given in~\appref{subsec:synthetic_exp_details}.

\begin{figure*}[t]
\hspace{-5mm}
  \begin{tabular}{ccccccc}
    \includegraphics[width=0.16\textwidth]{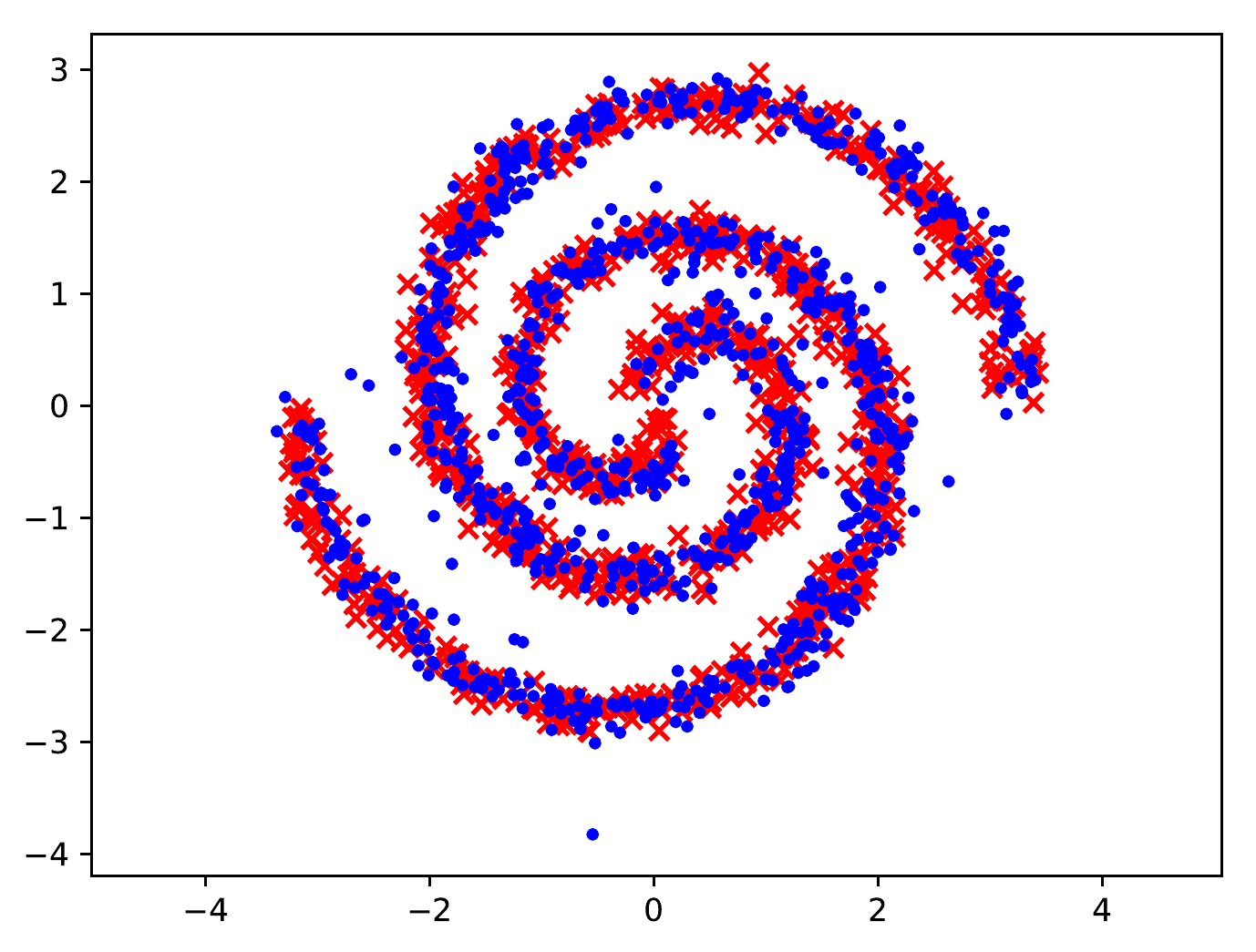} &\hspace{-3mm} 
    \includegraphics[width=0.16\textwidth]{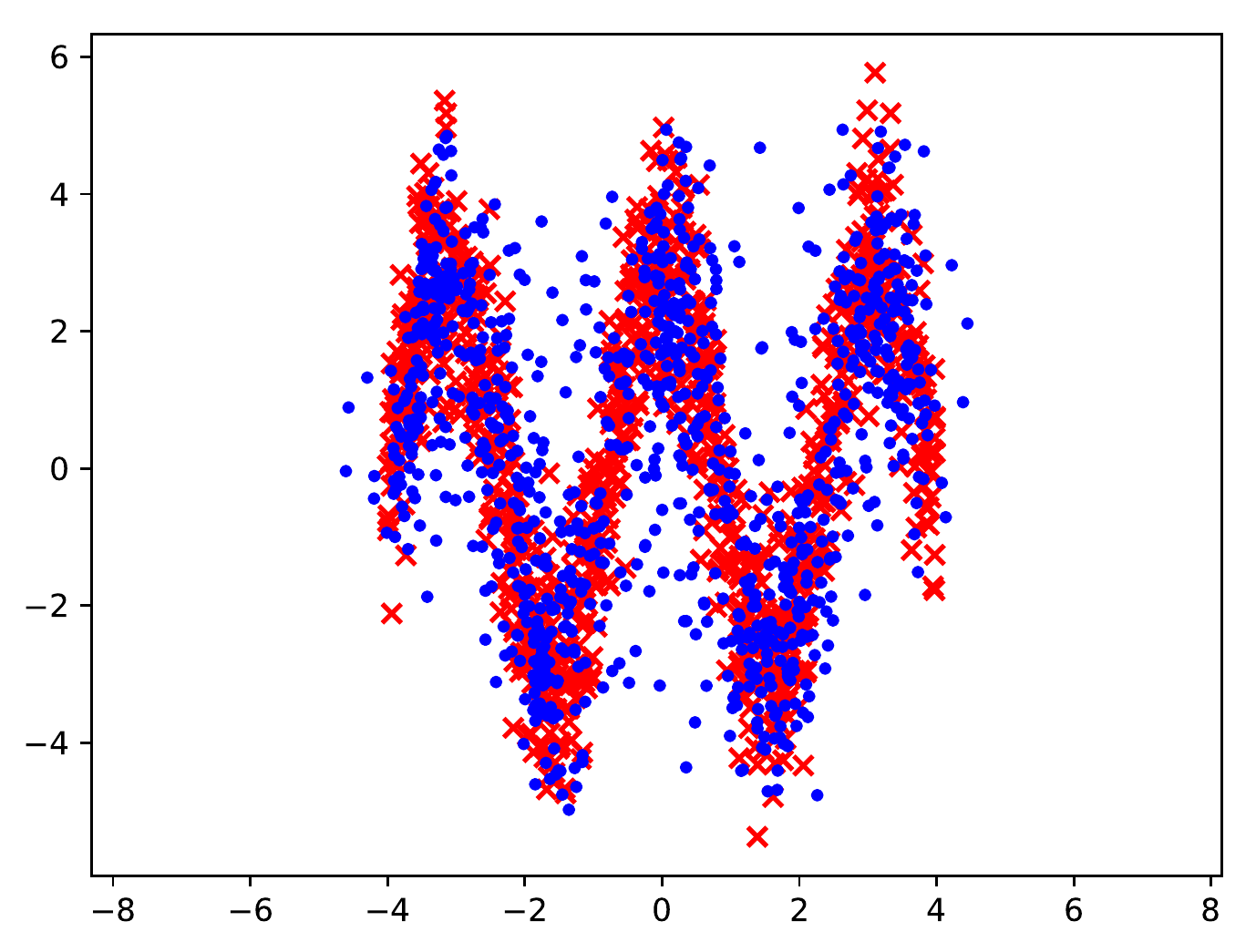} &\hspace{-3mm}
    \includegraphics[width=0.16\textwidth]{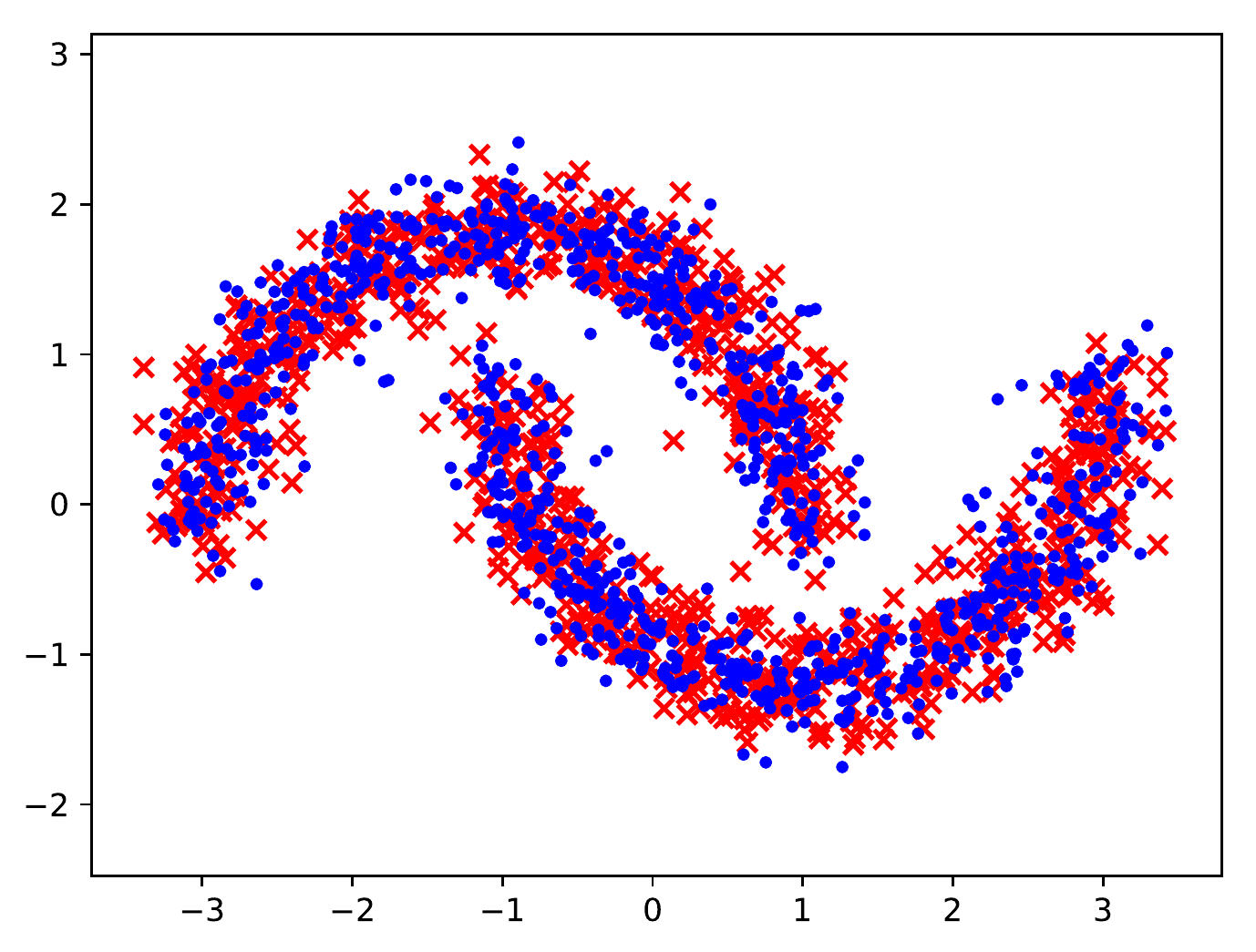} &\hspace{-3mm}
    \includegraphics[width=0.16\textwidth]{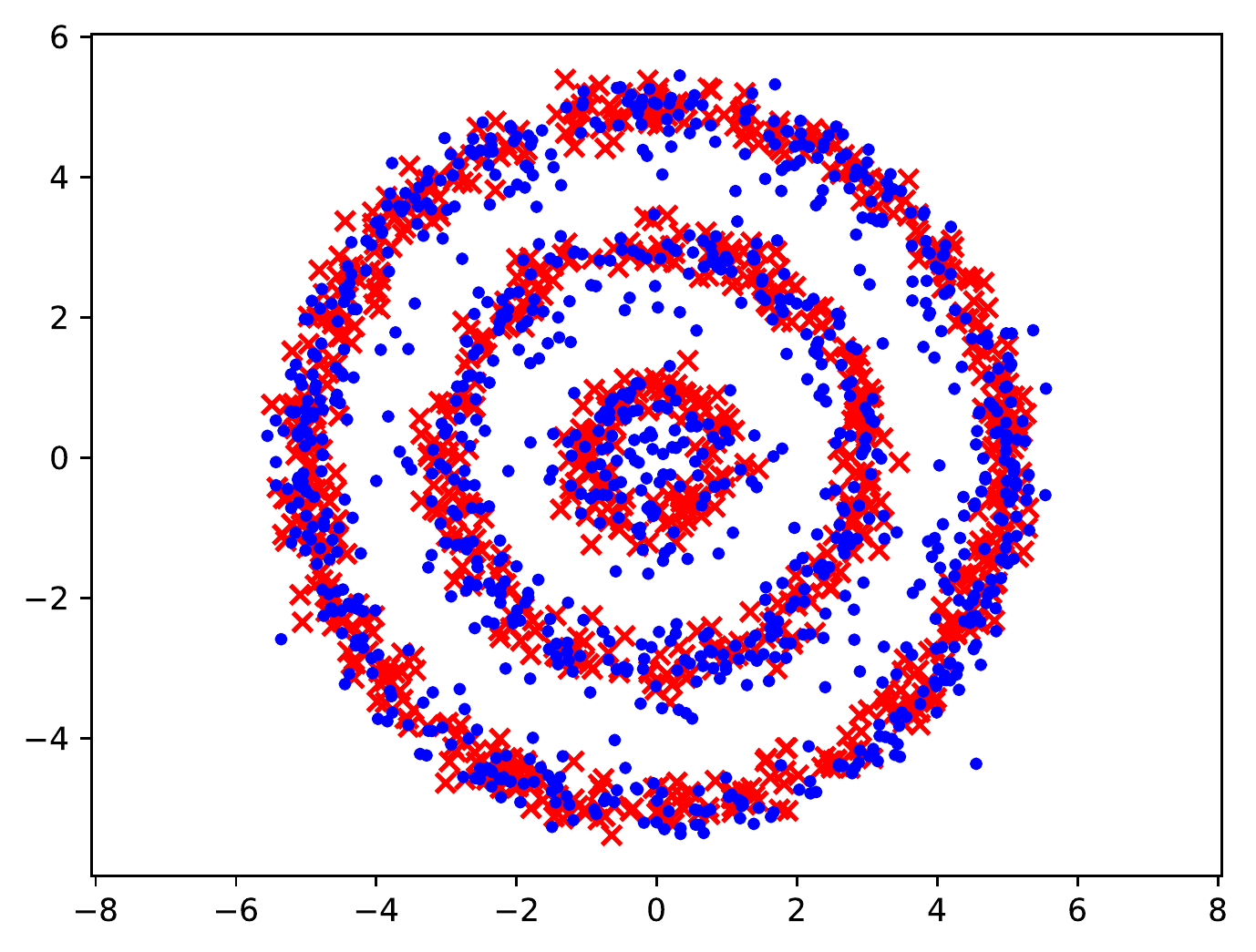} & \hspace{-3mm}
    \includegraphics[width=0.16\textwidth]{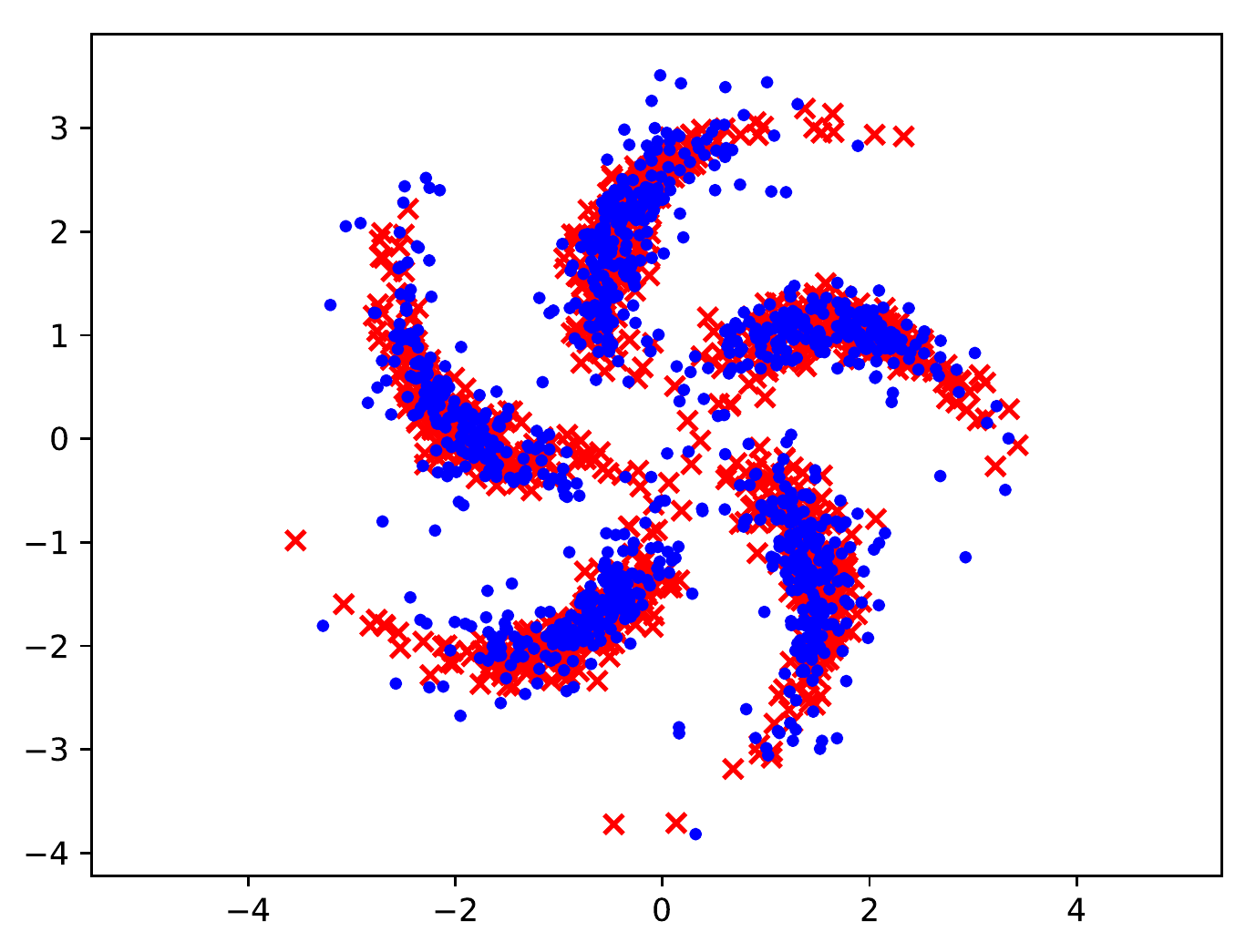} & \hspace{-3mm}
    \includegraphics[width=0.16\textwidth]{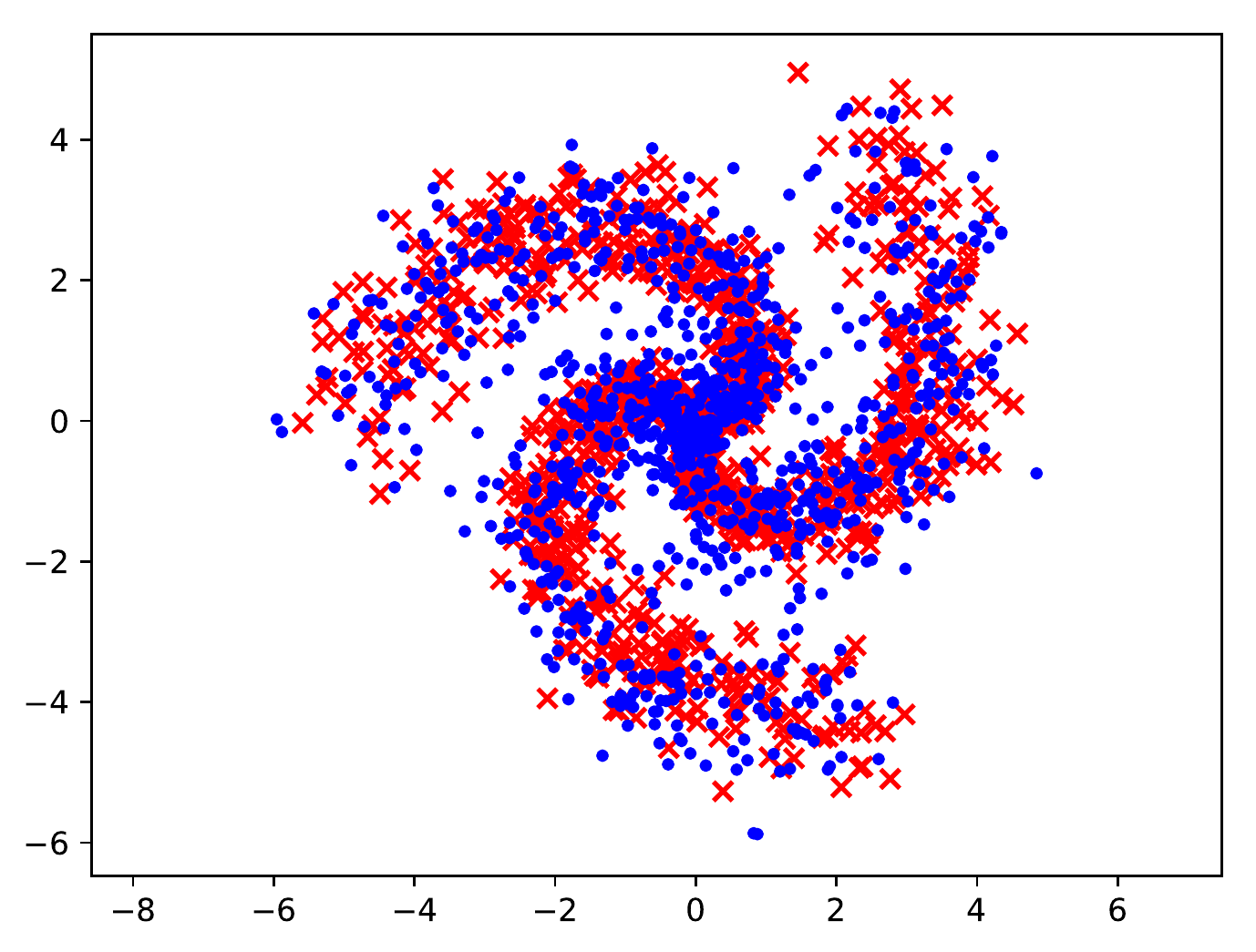} \\[-2mm]

    \tikz[baseline=(a.north)]\node[yscale=-1,inner sep=0,outer sep=0](a){\includegraphics[width=0.16\textwidth, trim={2cm 0cm 2cm 1cm},clip]{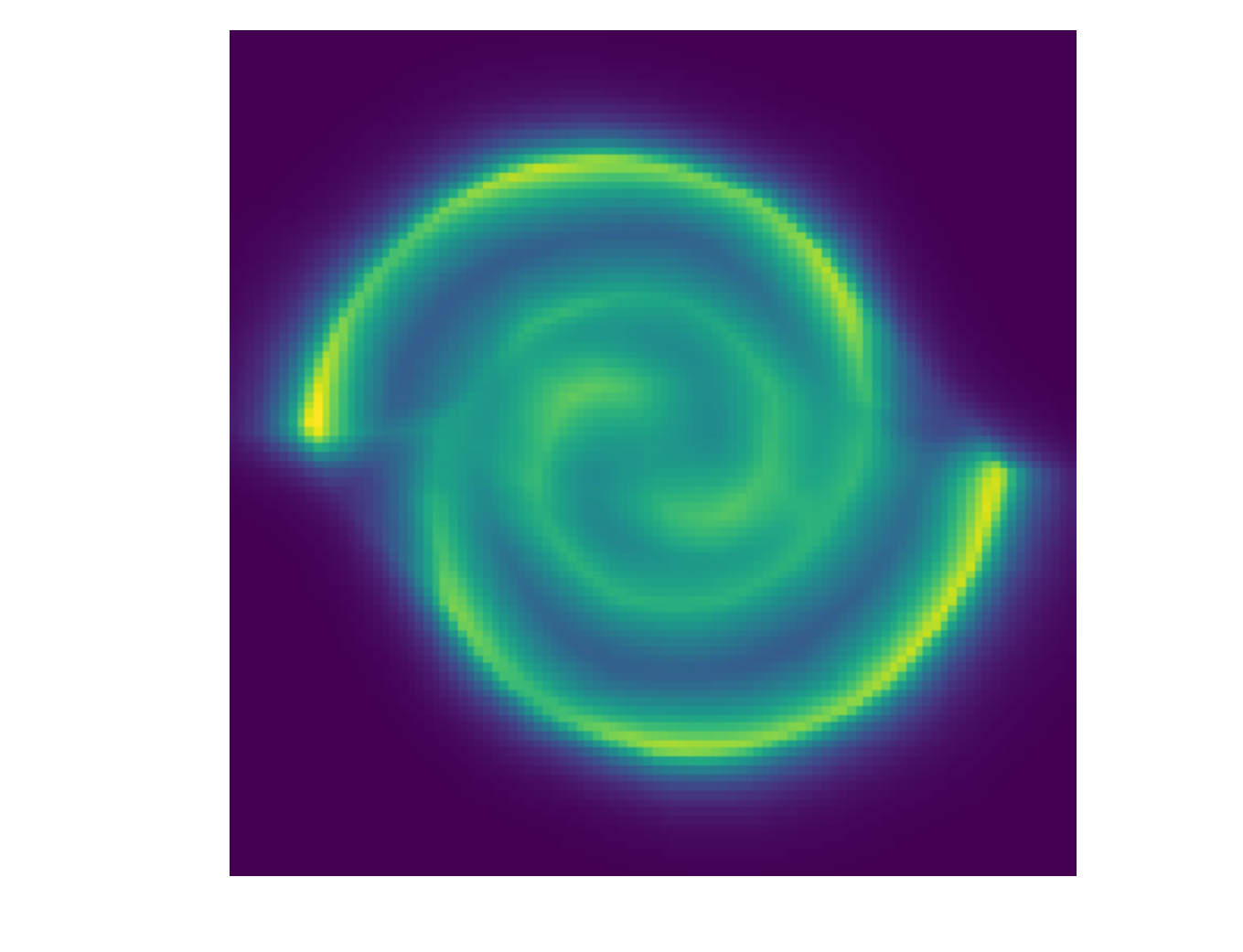}}; &\hspace{-3mm}
    \tikz[baseline=(a.north)]\node[yscale=-1,inner sep=0,outer sep=0](a){\includegraphics[width=0.16\textwidth, trim={2cm 0cm 2cm 1cm},clip]{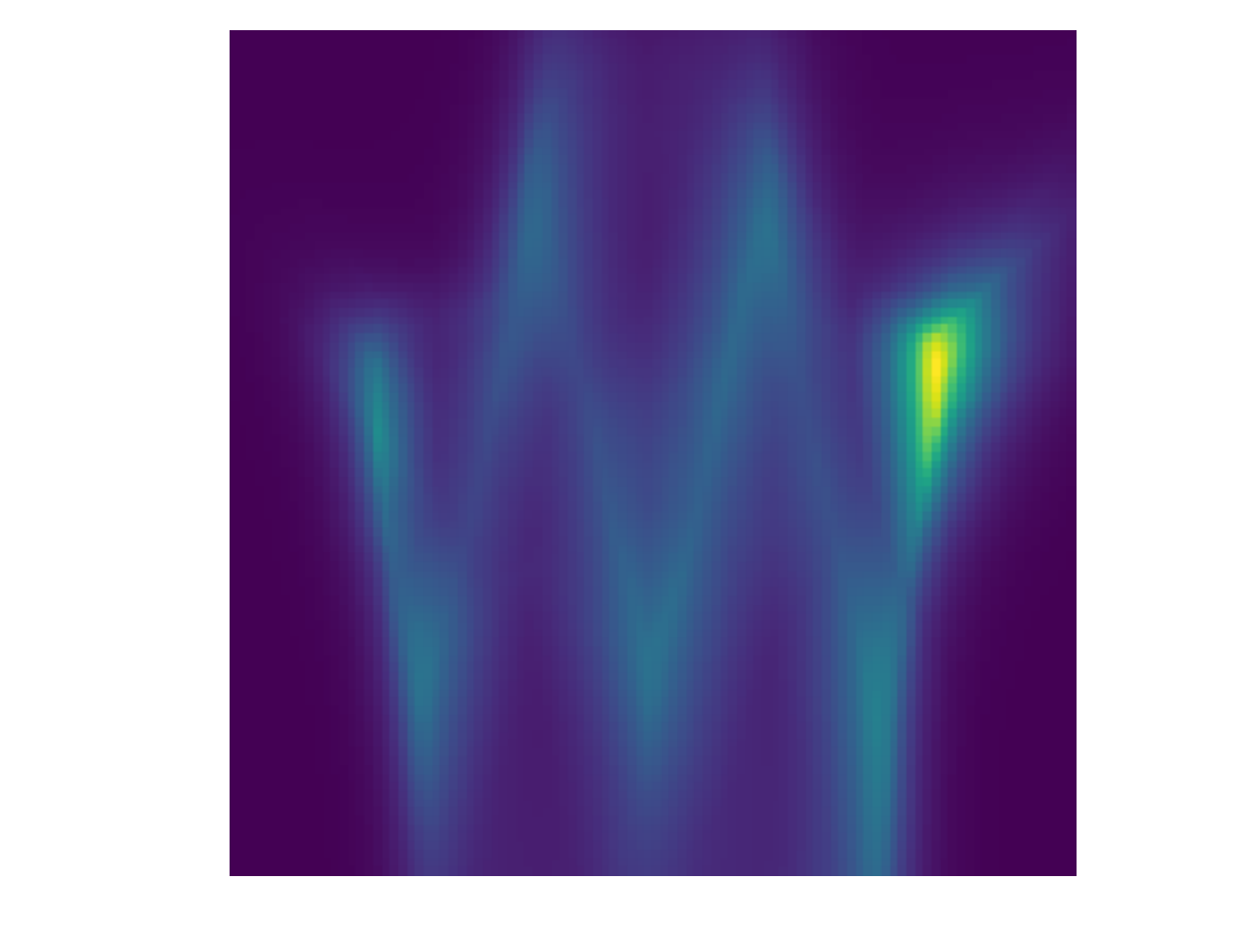}}; & \hspace{-3mm}
    \tikz[baseline=(a.north)]\node[yscale=-1,inner sep=0,outer sep=0](a){\includegraphics[width=0.16\textwidth, trim={2cm 0cm 2cm 1cm},clip]{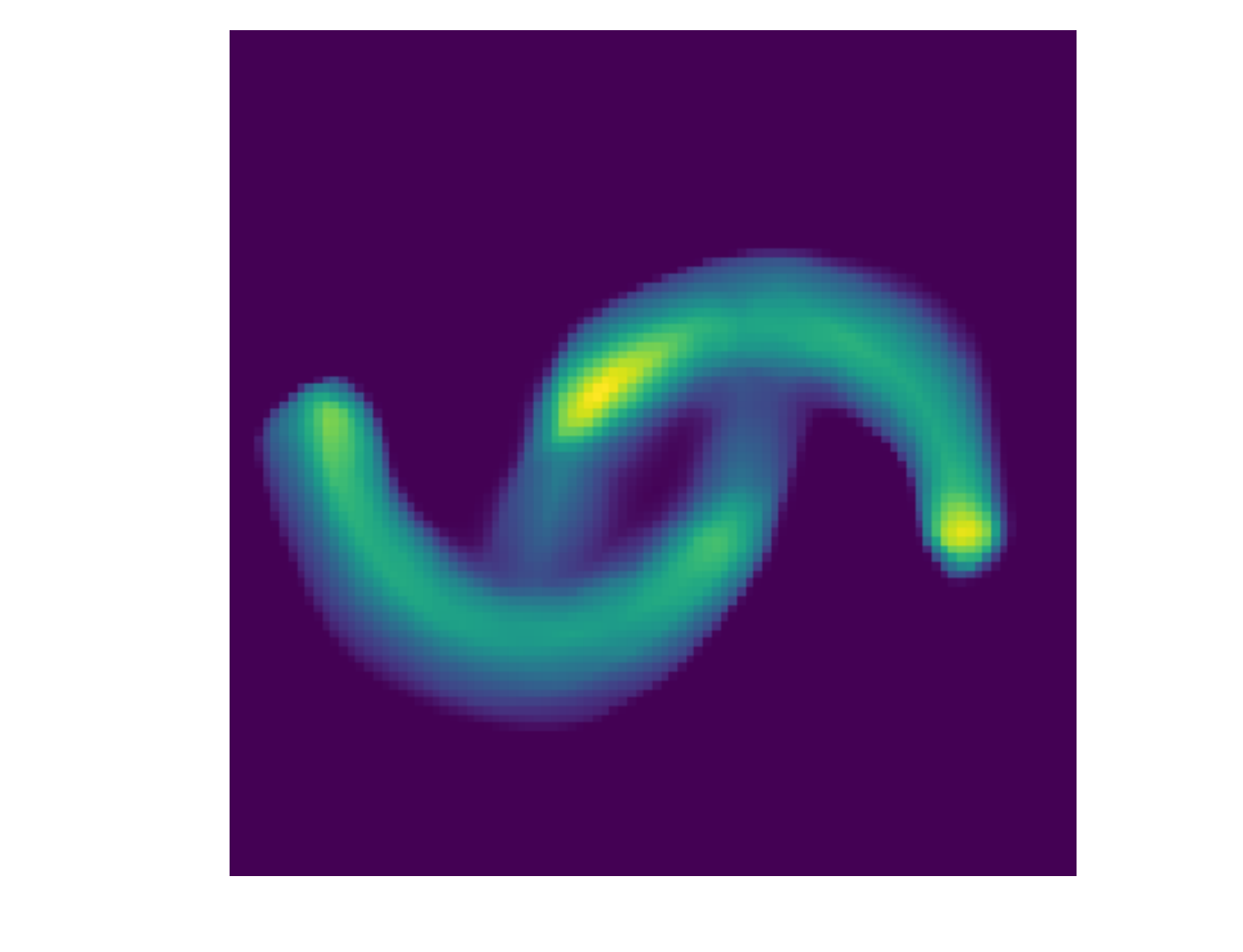}}; &\hspace{-3mm}
    \tikz[baseline=(a.north)]\node[yscale=-1,inner sep=0,outer sep=0](a){\includegraphics[width=0.16\textwidth, trim={2cm 0cm 2cm 1cm},clip]{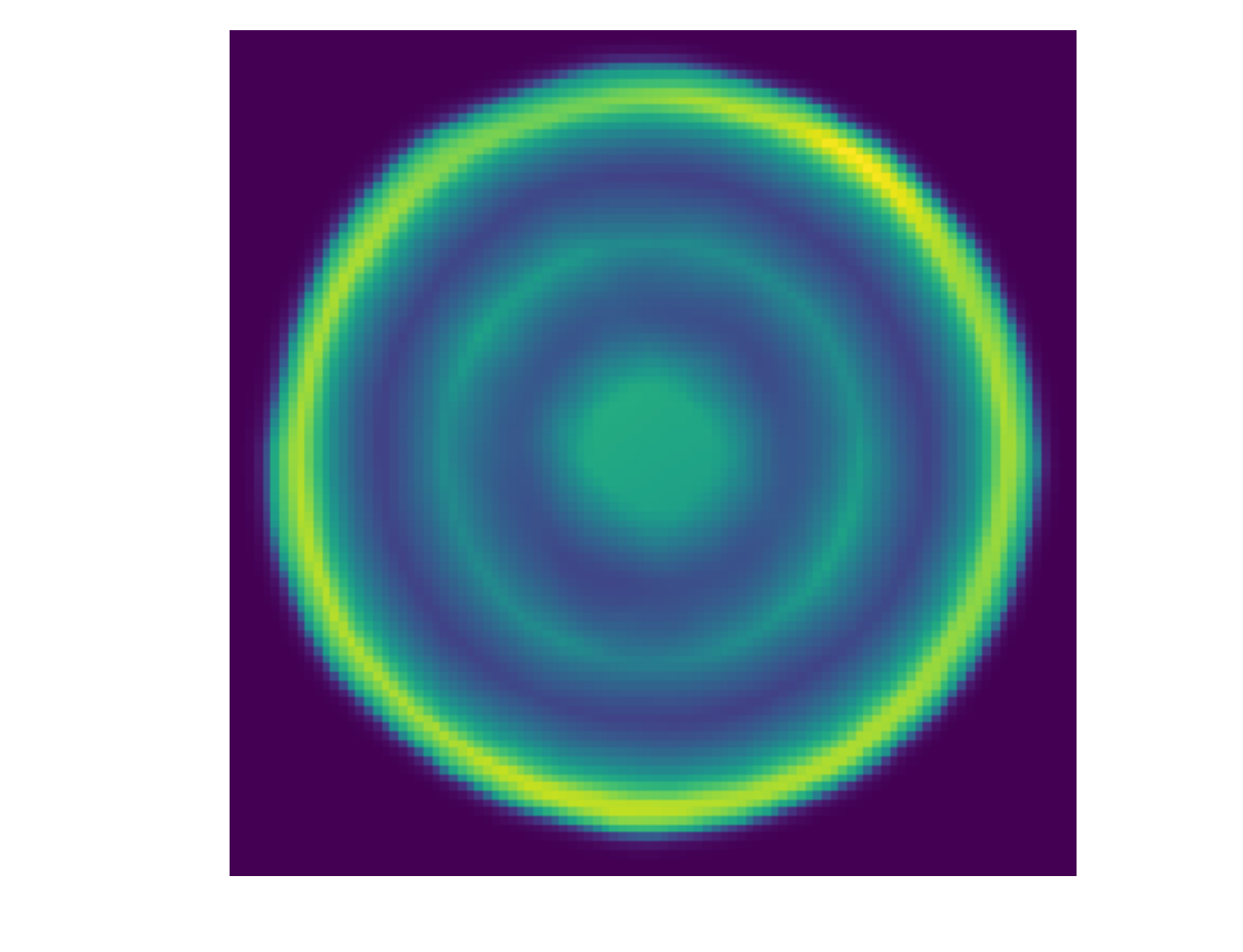}}; & \hspace{-3mm}
    \tikz[baseline=(a.north)]\node[yscale=-1,inner sep=0,outer sep=0](a){\includegraphics[width=0.16\textwidth, trim={2cm 0cm 2cm 1cm},clip]{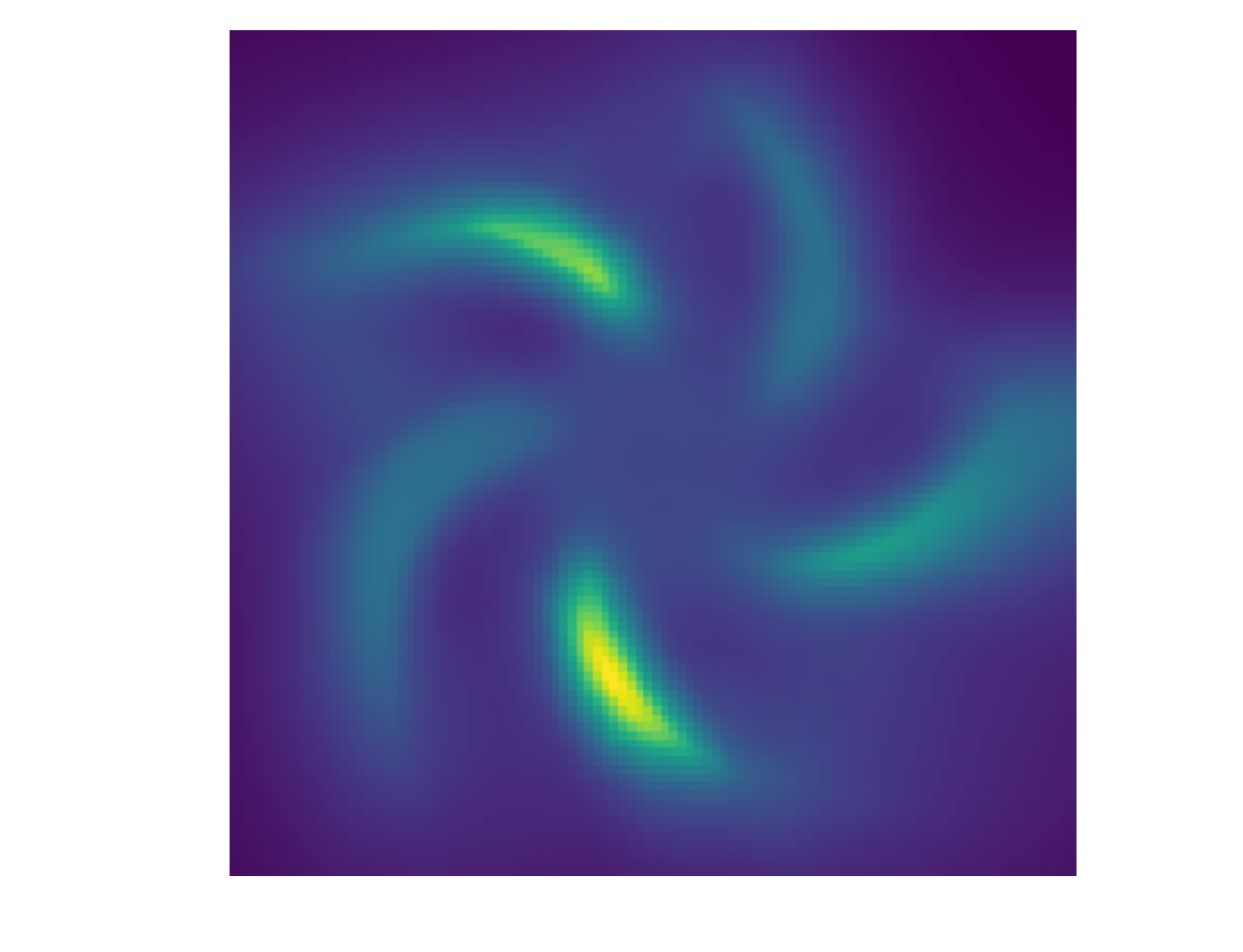}}; & \hspace{-3mm}
    \tikz[baseline=(a.north)]\node[yscale=-1,inner sep=0,outer sep=0](a){\includegraphics[width=0.16\textwidth, trim={2cm 0cm 2cm 1cm},clip]{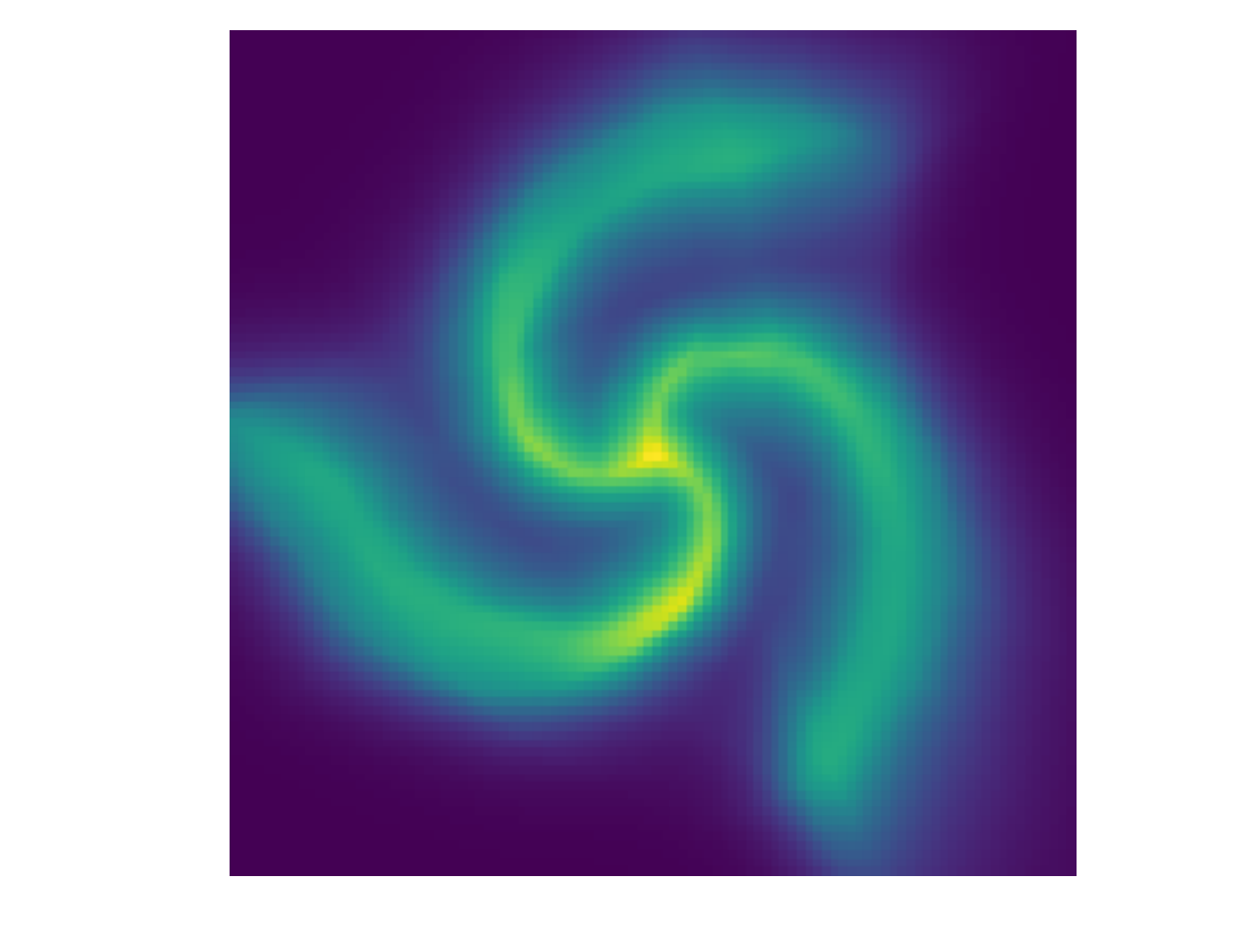}}; \\
    (a) \texttt{2spirals} &\hspace{-3mm}(b) \texttt{Cosine} &\hspace{-3mm}(c) \texttt{moons} &\hspace{-3mm}(d) \texttt{Multiring} &\hspace{-3mm}(e) \texttt{pinwheel} &\hspace{-3mm}(f) \texttt{Spiral} 

  \end{tabular}
  \vspace{-1mm}
  \caption{We illustrated the learned samplers from different synthetic datasets in the first row. The {\color{red}$\times$} denotes training data and {\color{blue}$\bullet$} denotes the ADE samplers. The learned potential functions $f$ are illustrated in the second row.}
  \label{fig:synthetic}
\end{figure*}

\begin{figure*}[t]
\resizebox{1.0\textwidth}{!}{%
  \begin{tabular}{cccccccc}
  \includegraphics{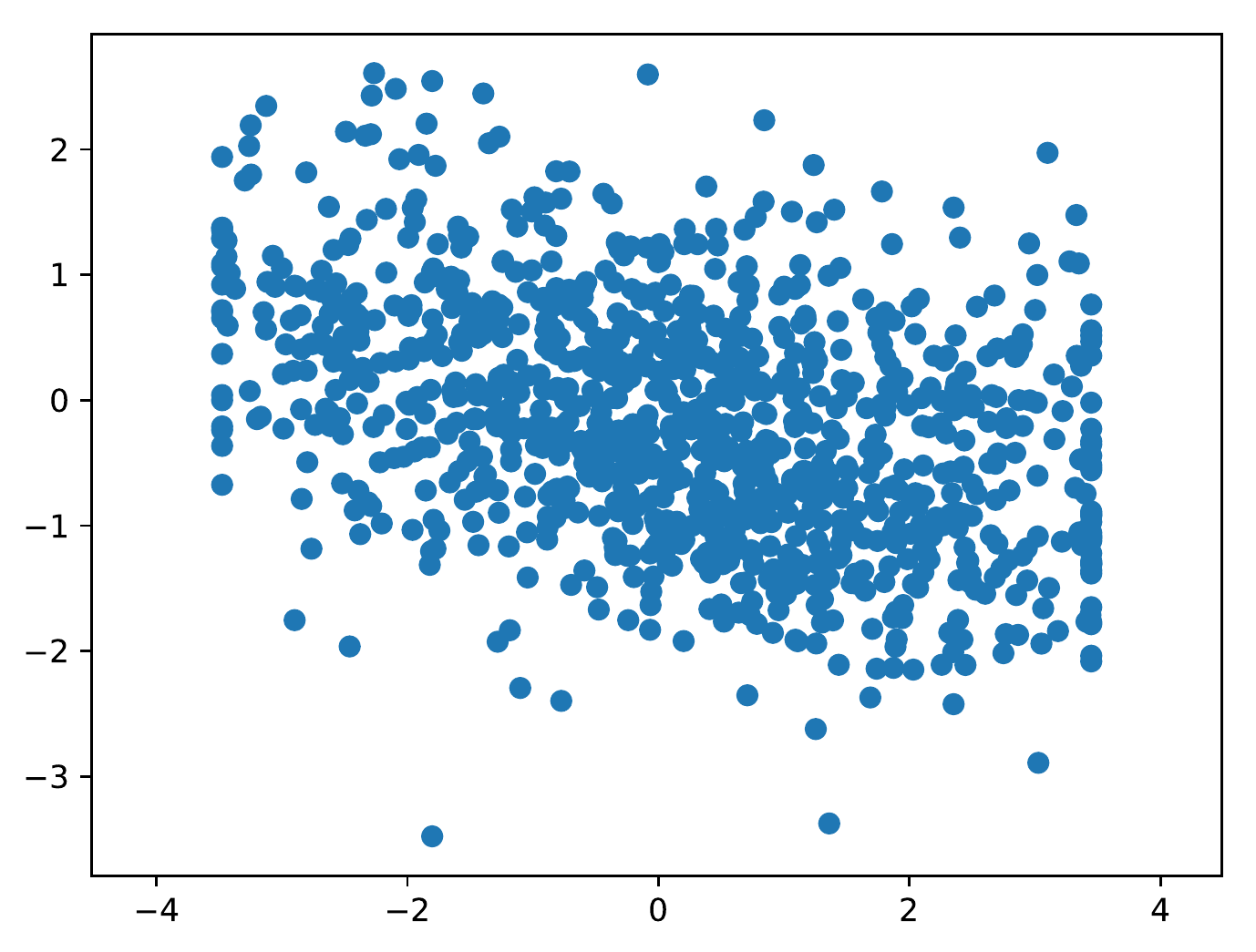} &
  \includegraphics{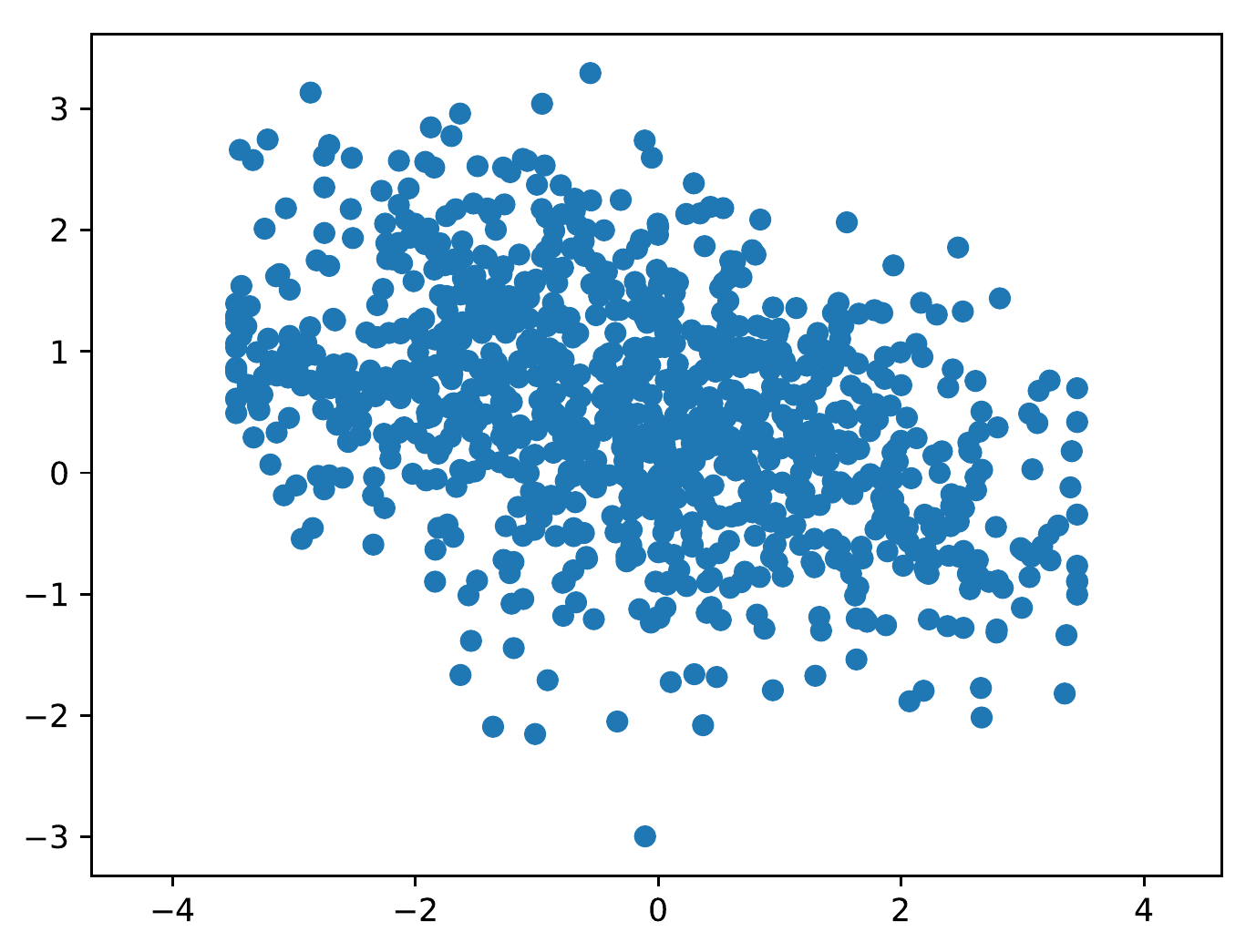} &
  \includegraphics{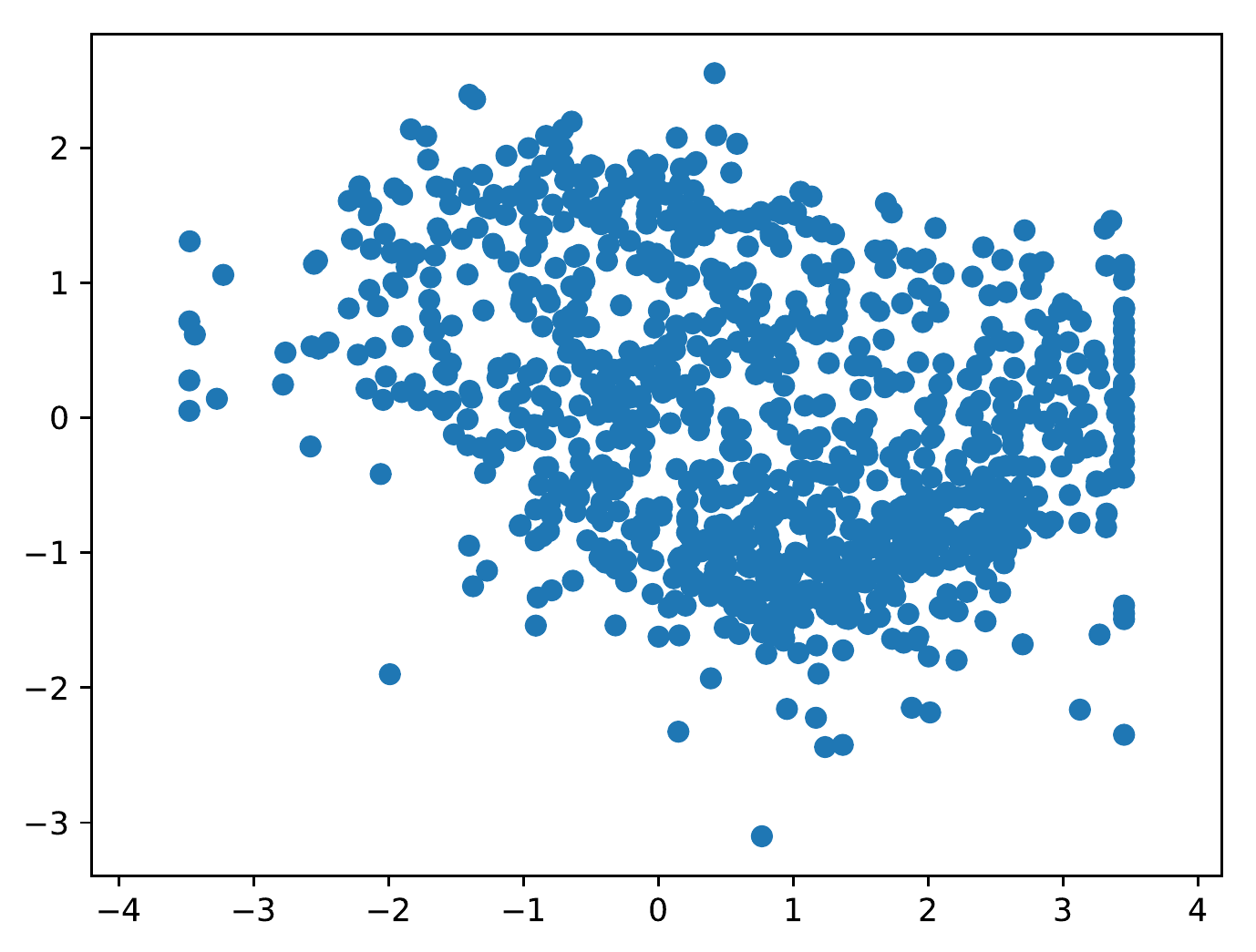} &
  \includegraphics{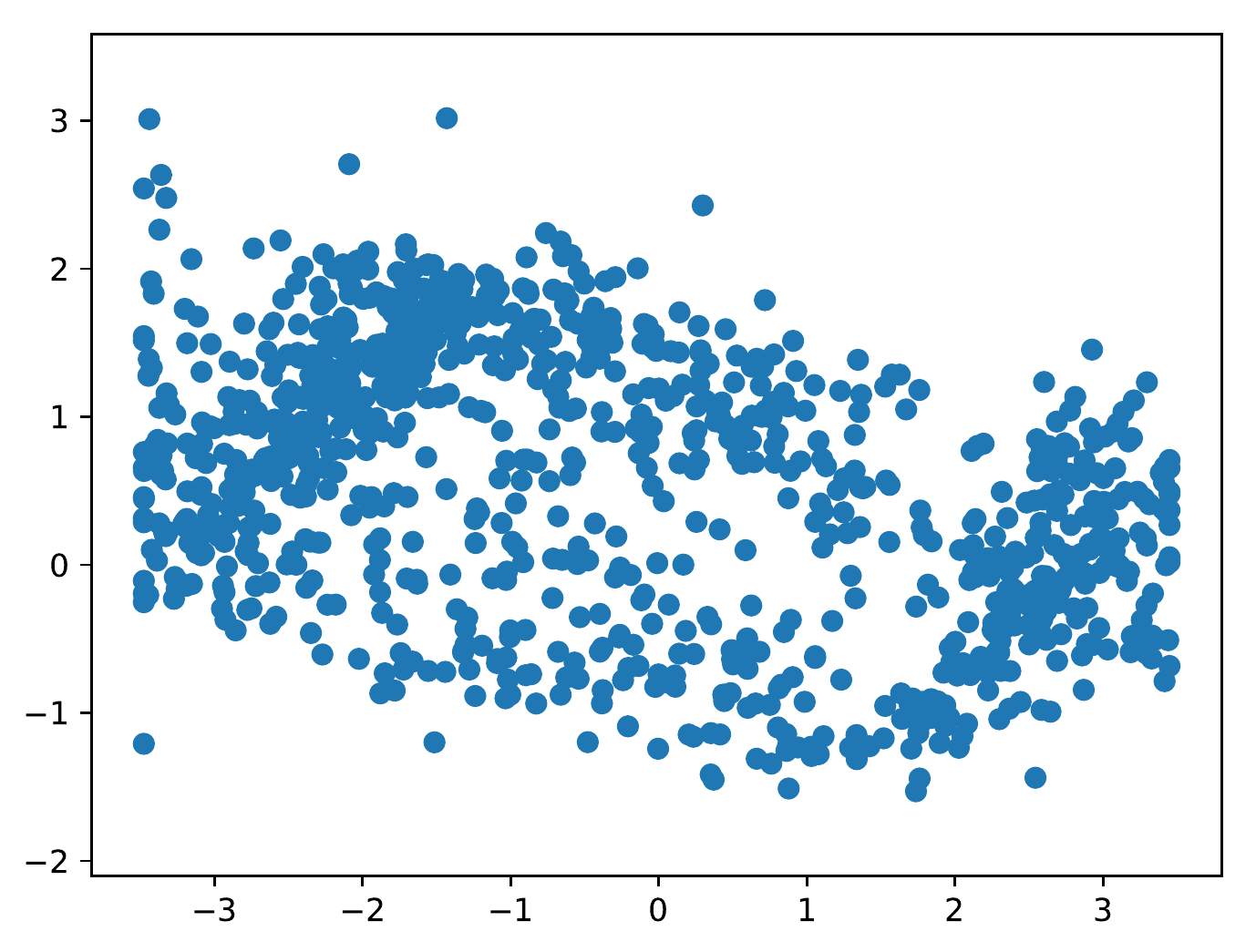} &
  \includegraphics{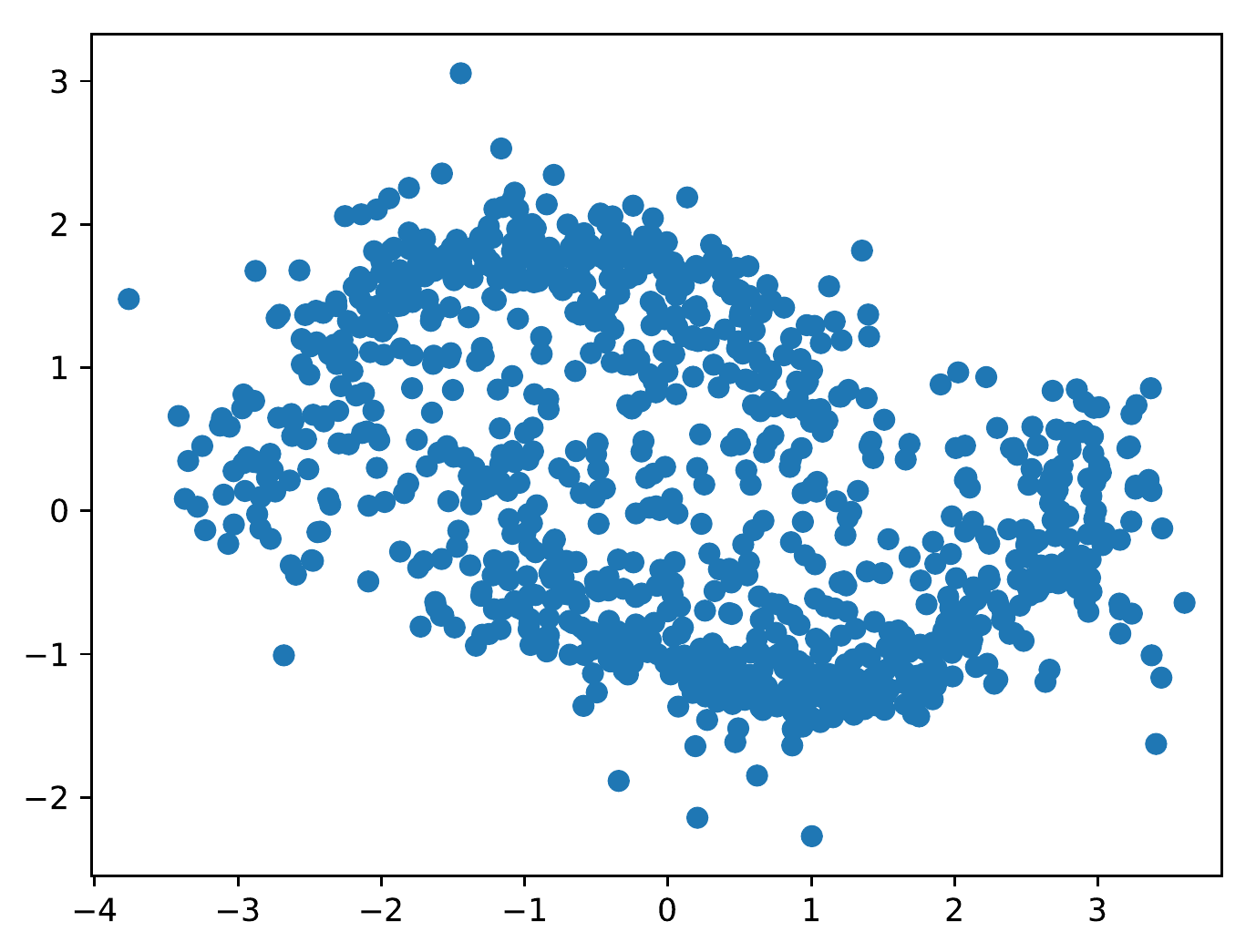} & 
  \includegraphics{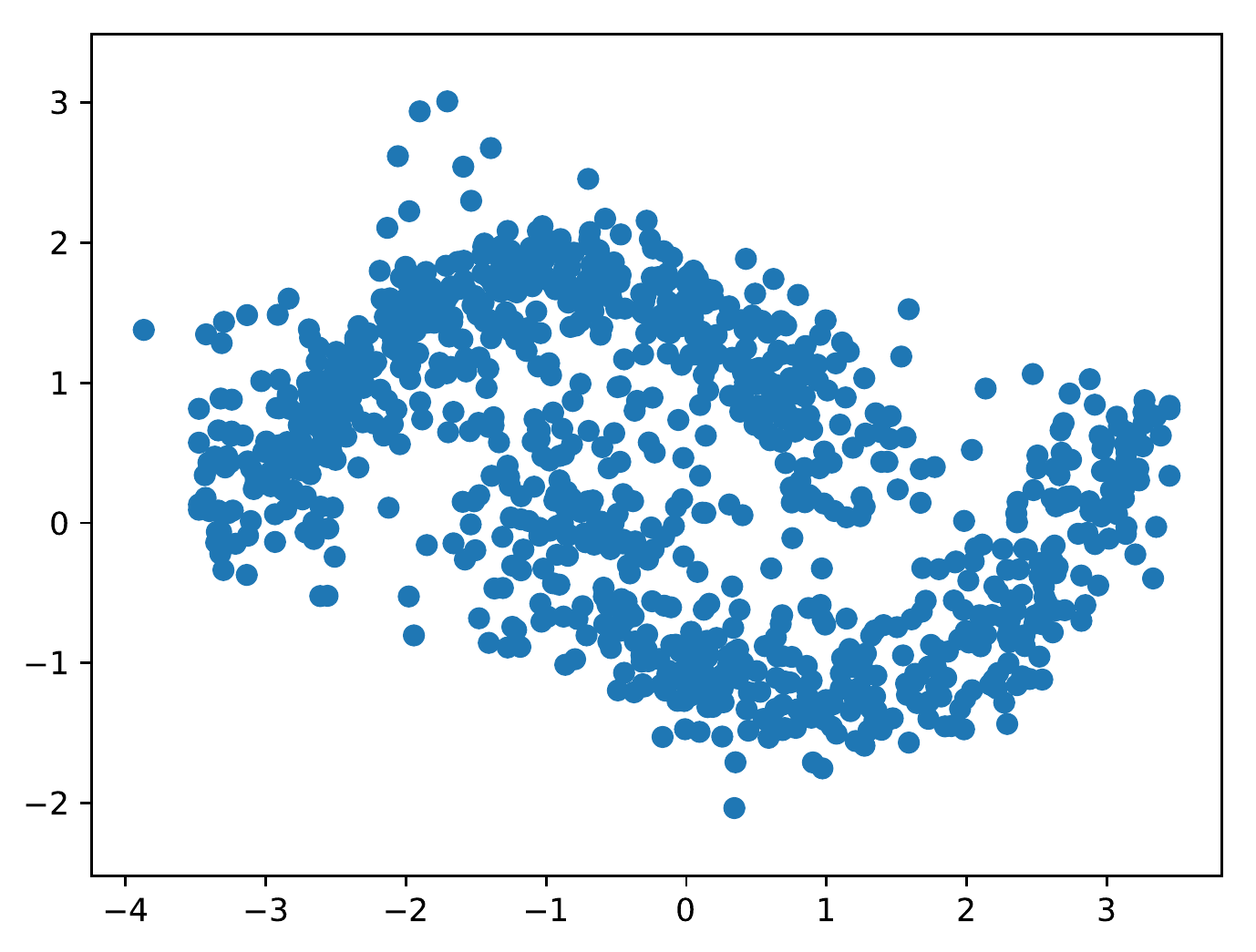} & 
  \includegraphics{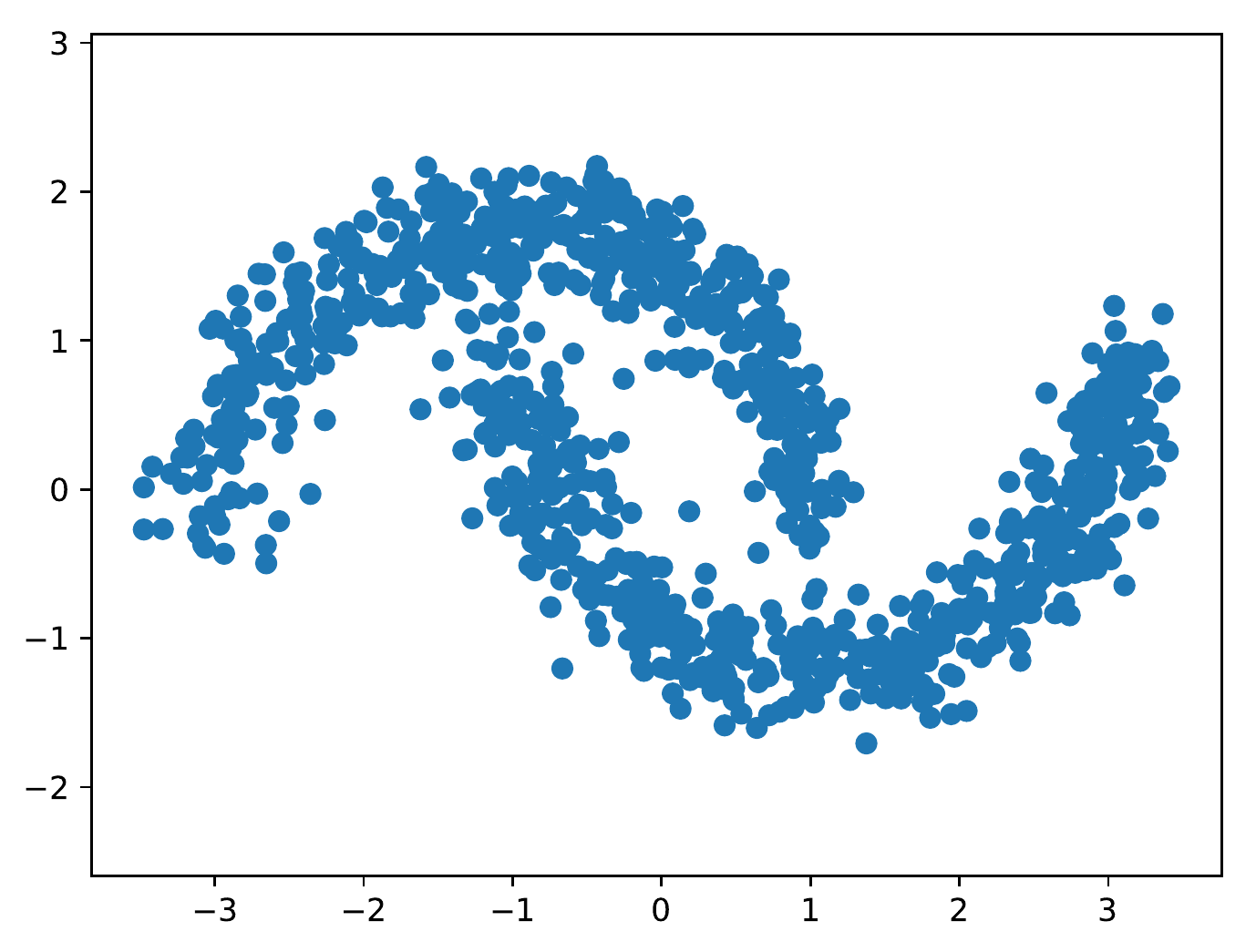} & 
  \includegraphics{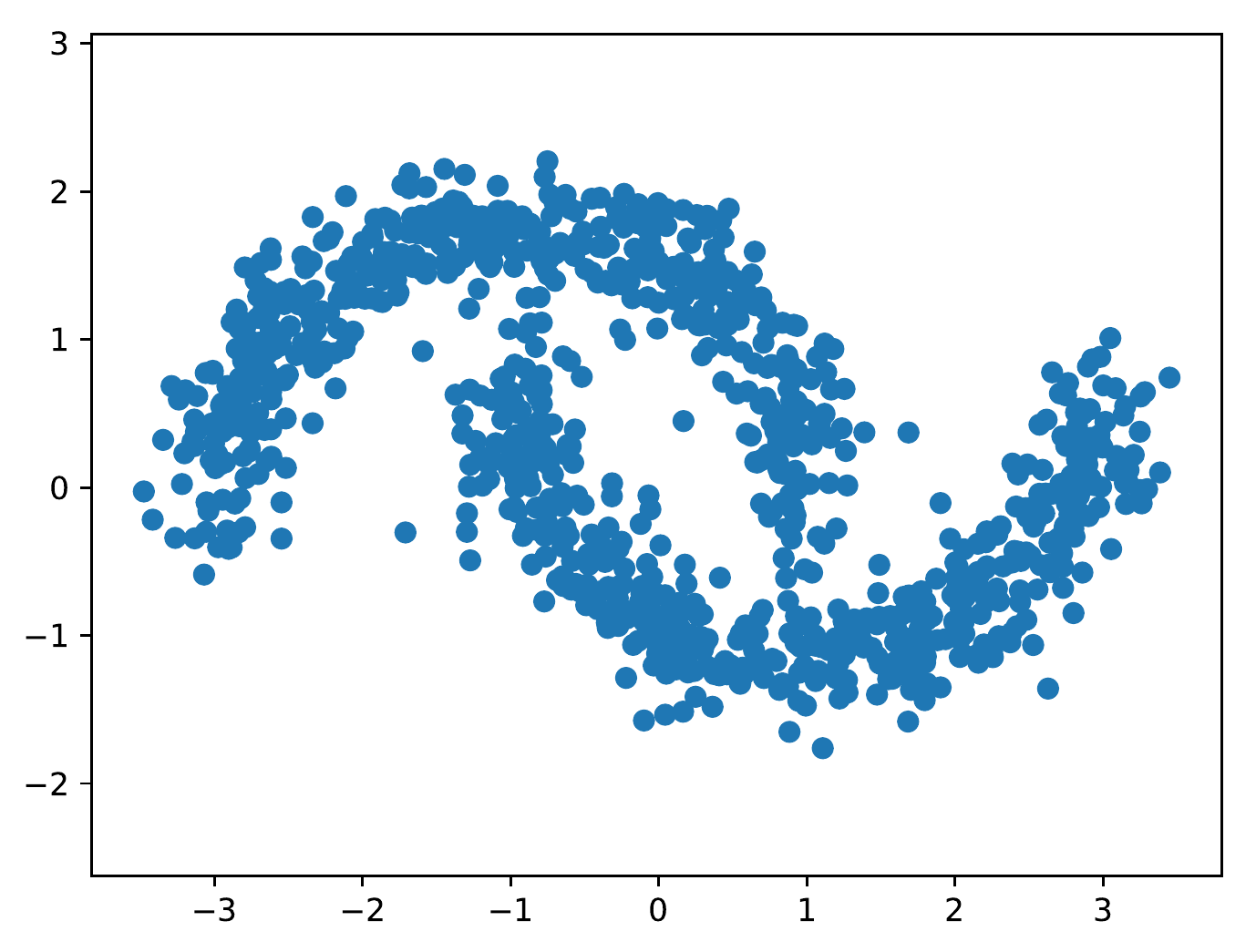} \\
  \includegraphics{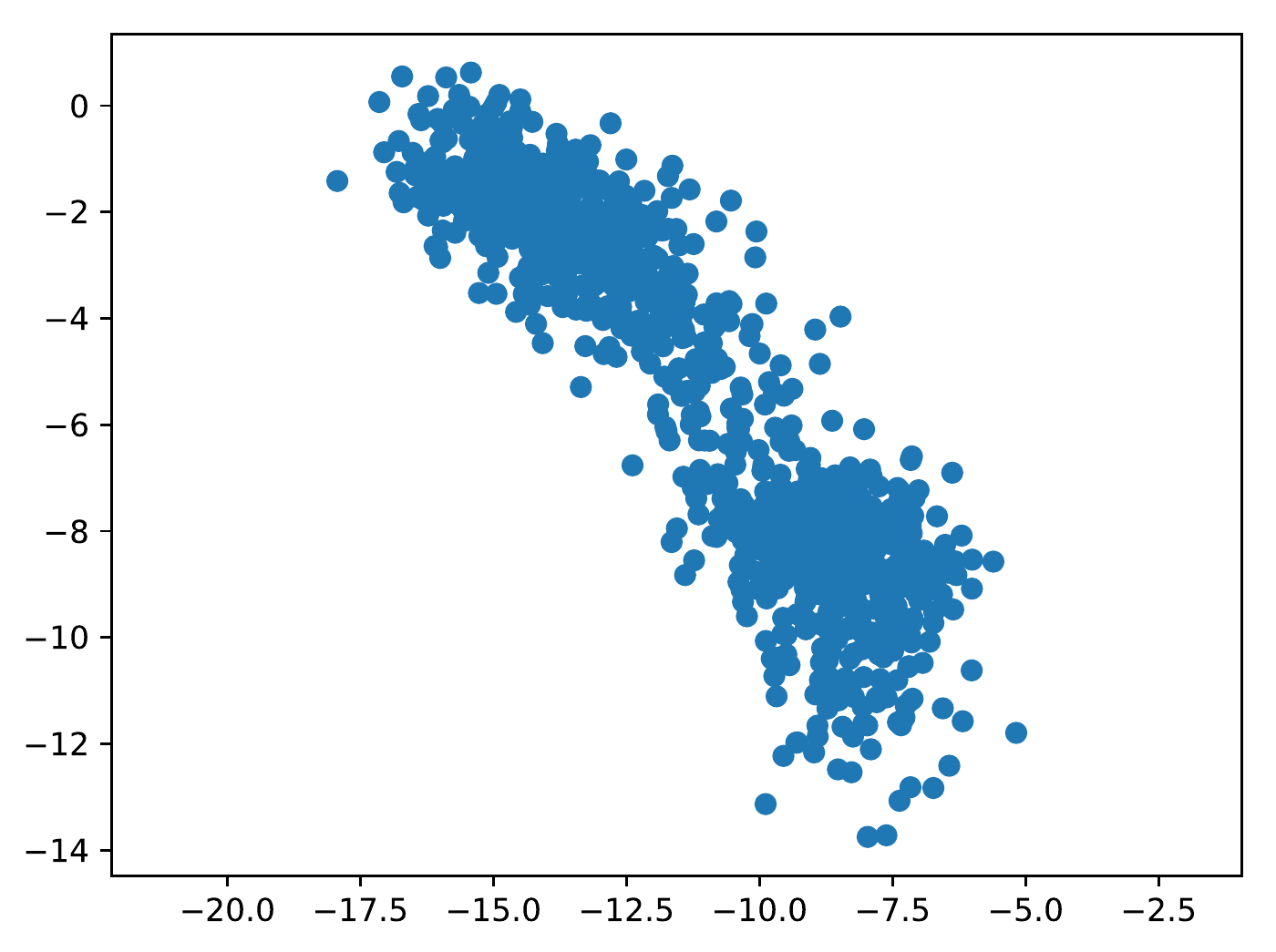} & 
  \includegraphics{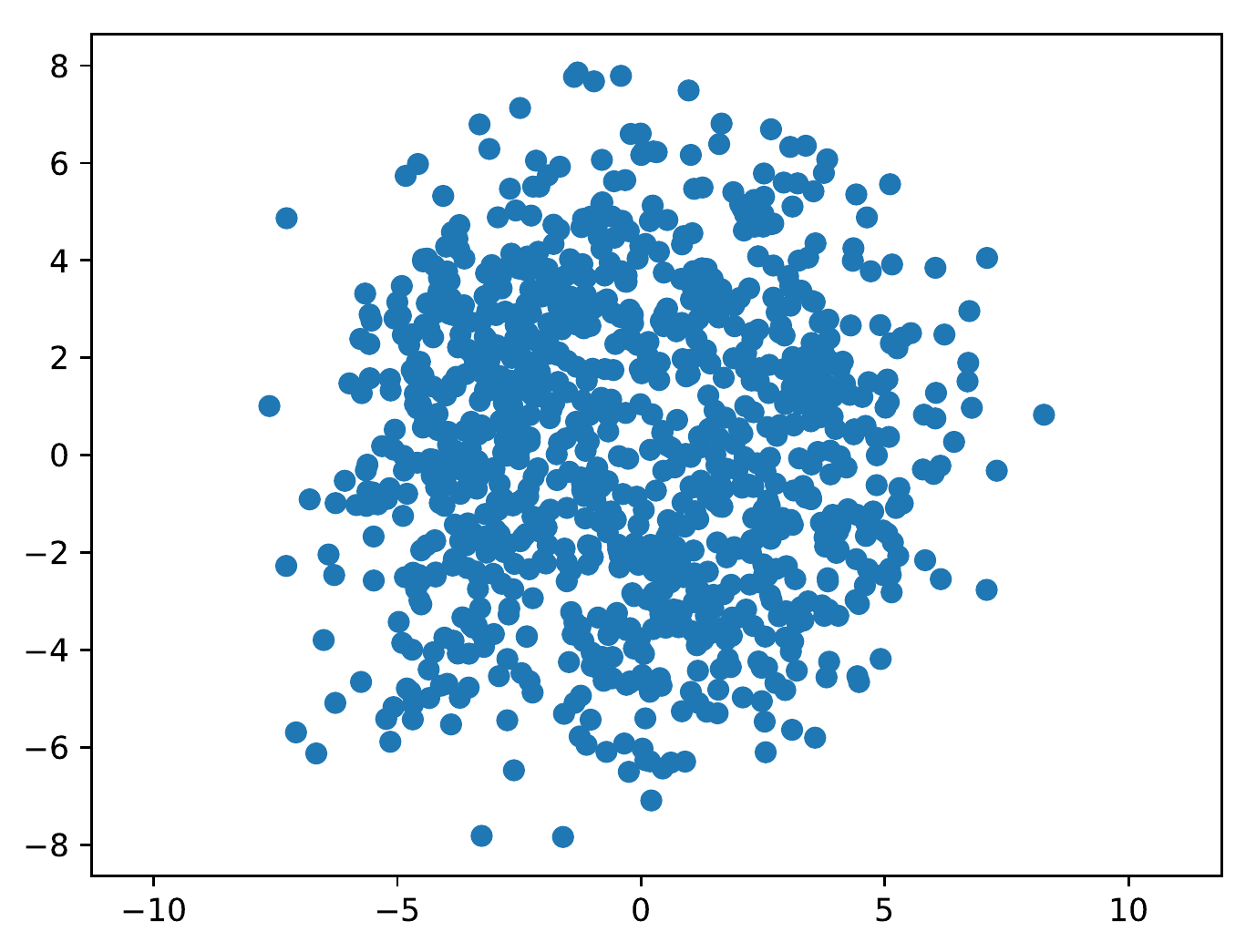} &
  \includegraphics{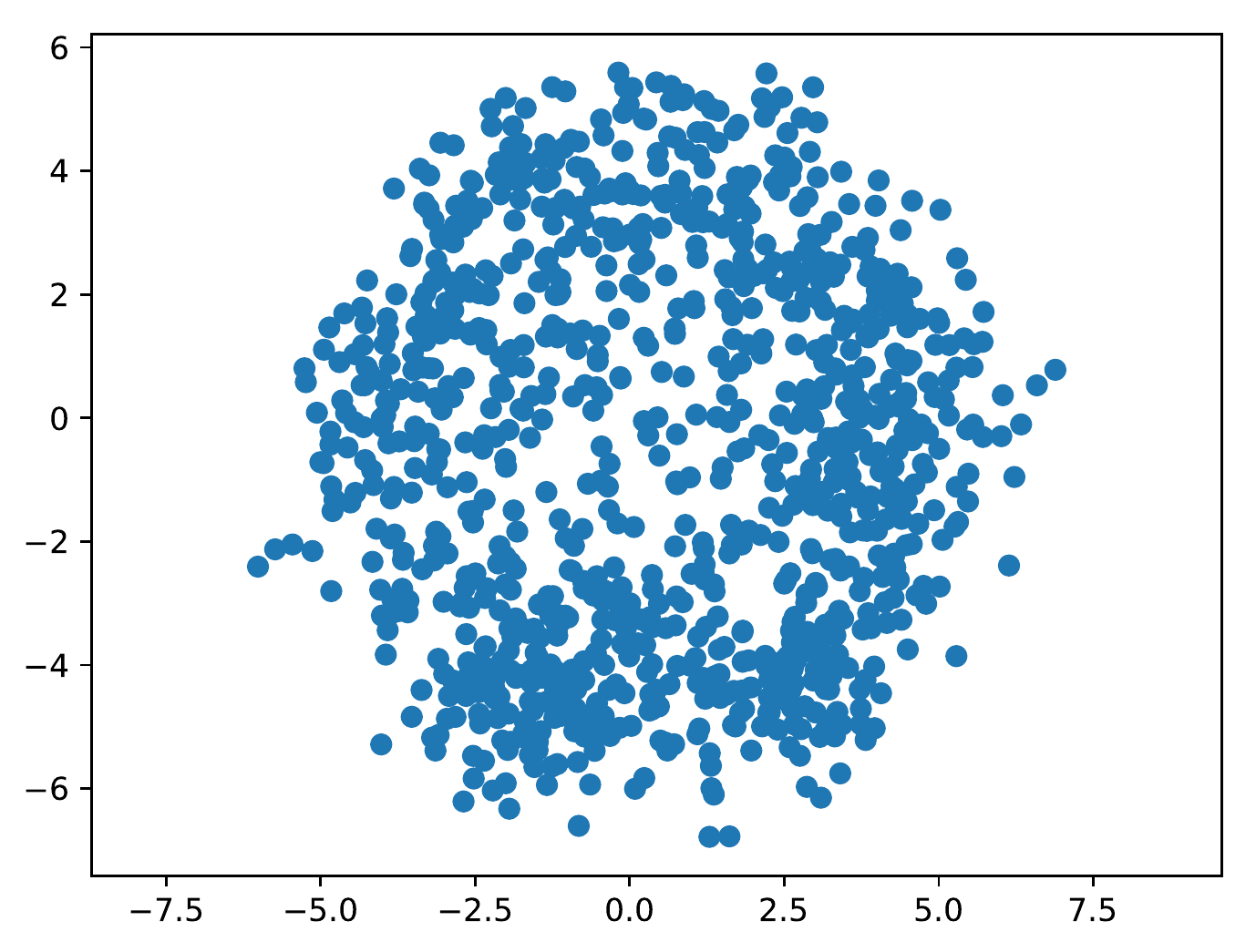} &
  \includegraphics{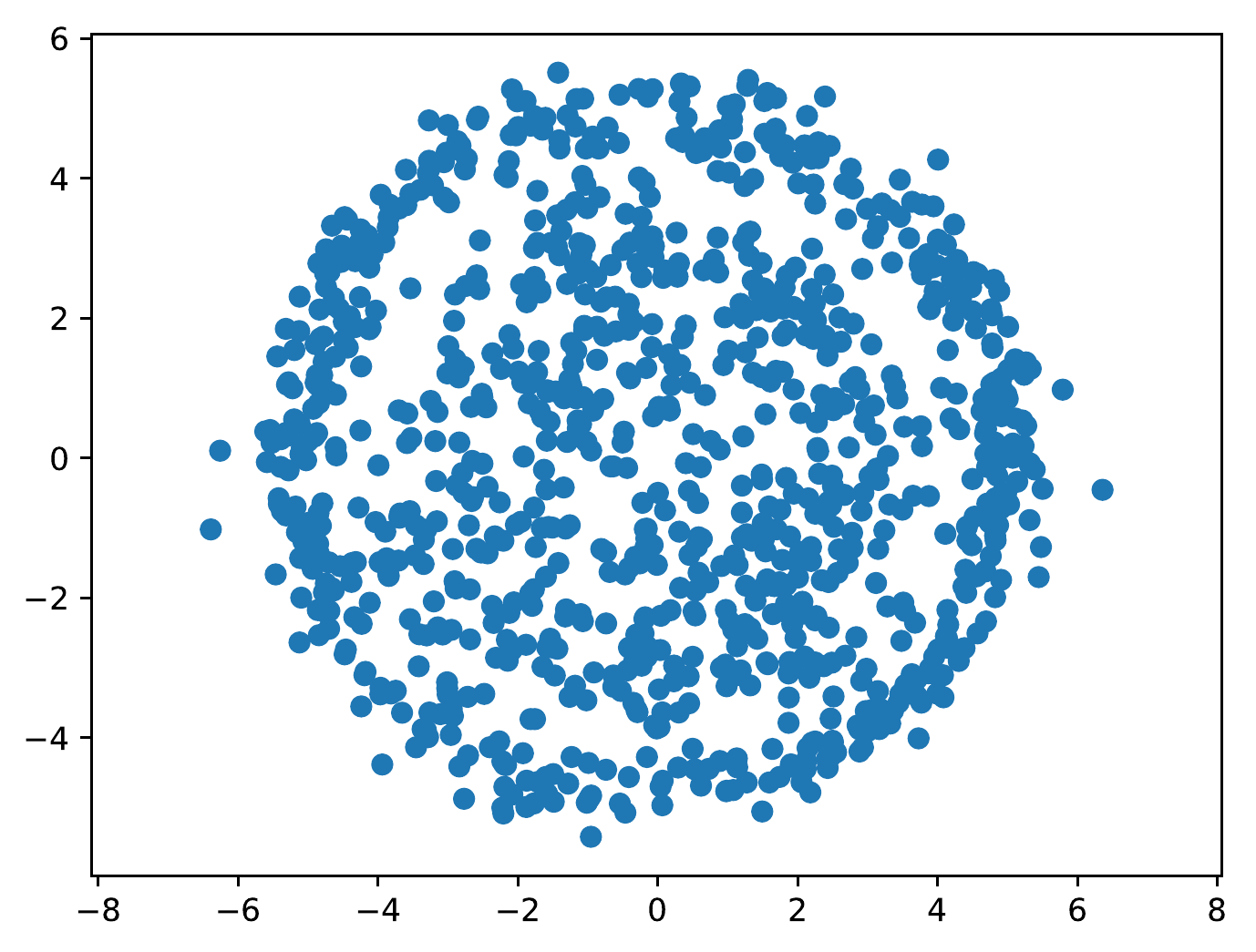} &
  \includegraphics{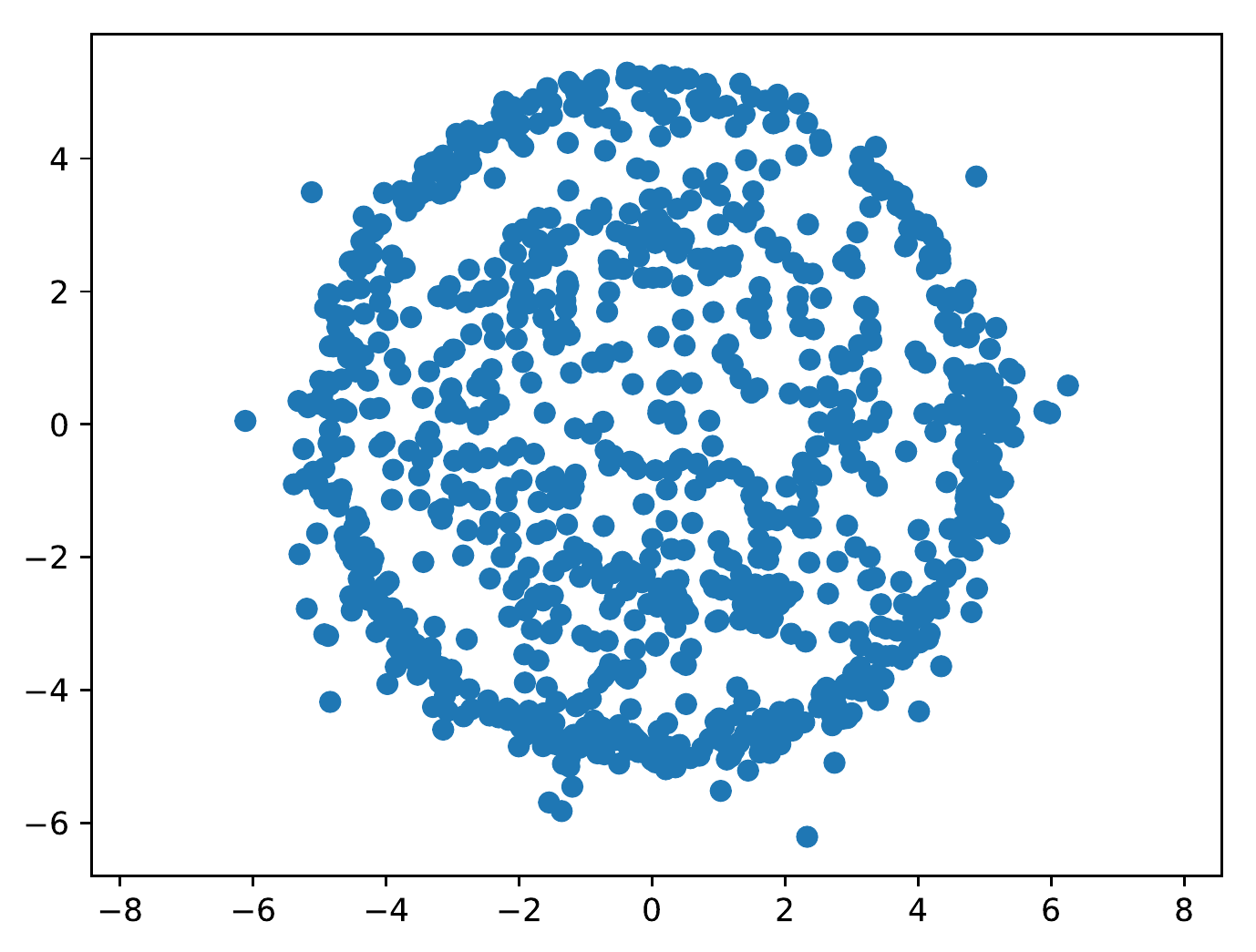} &
  \includegraphics{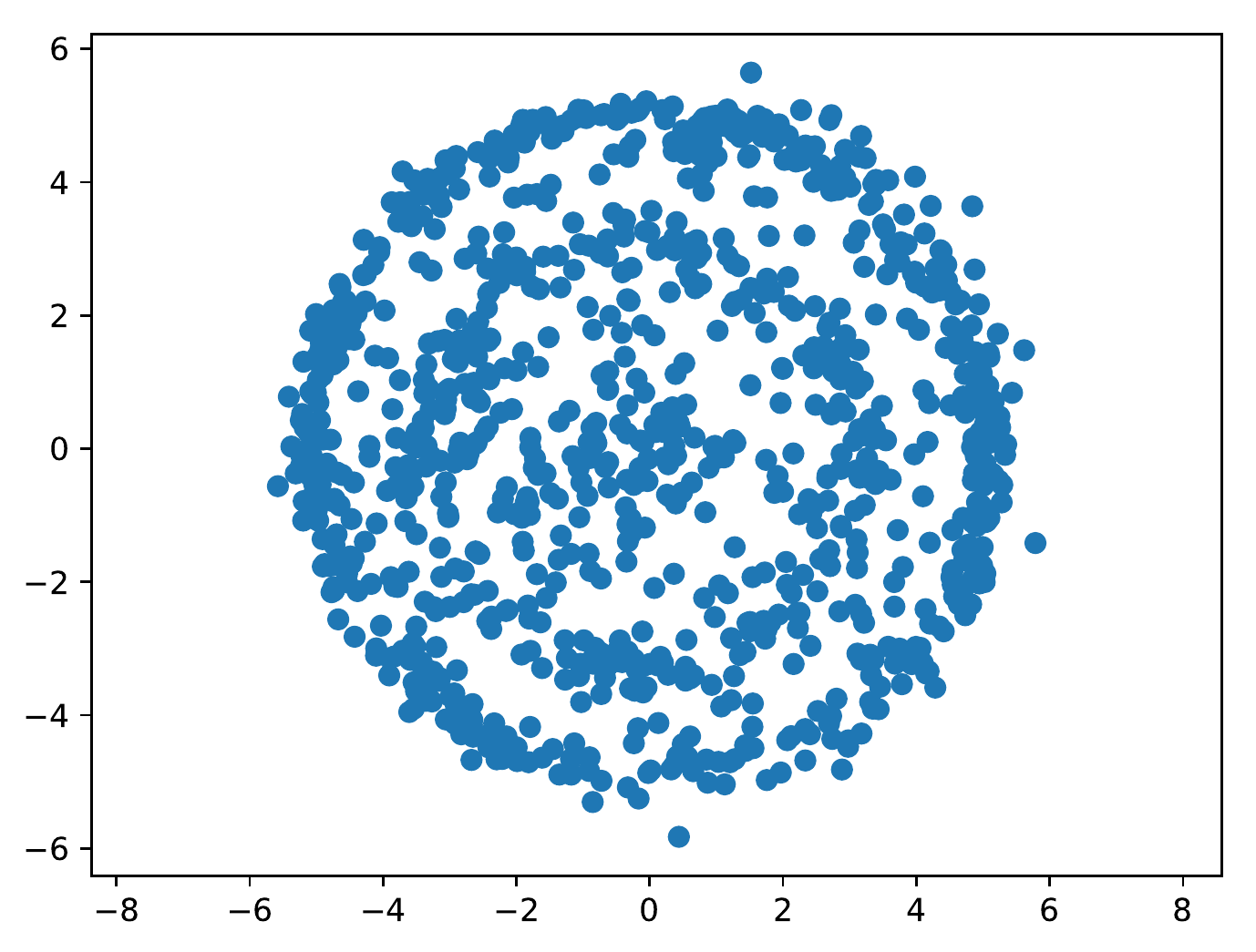} &
  \includegraphics{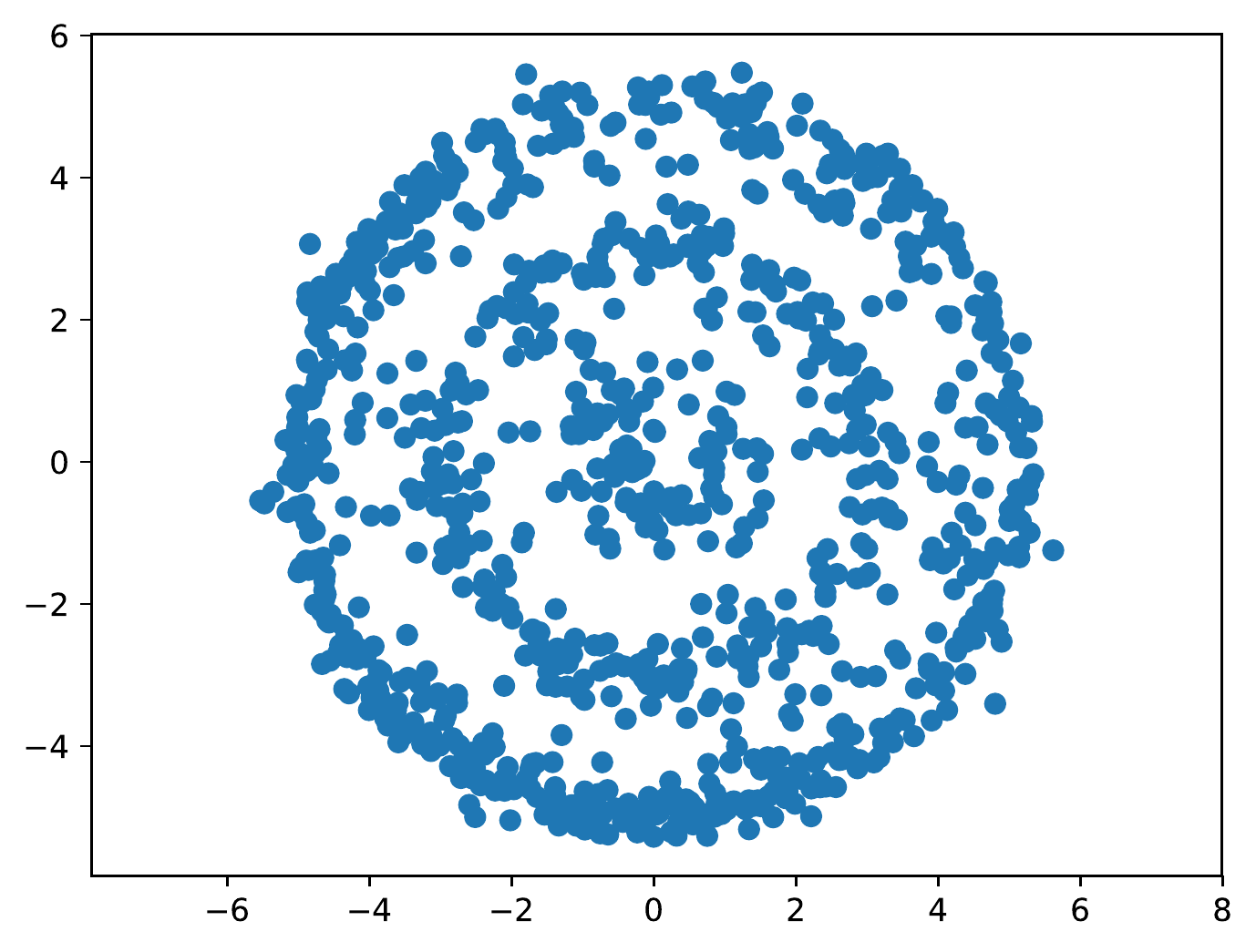} &
  \includegraphics{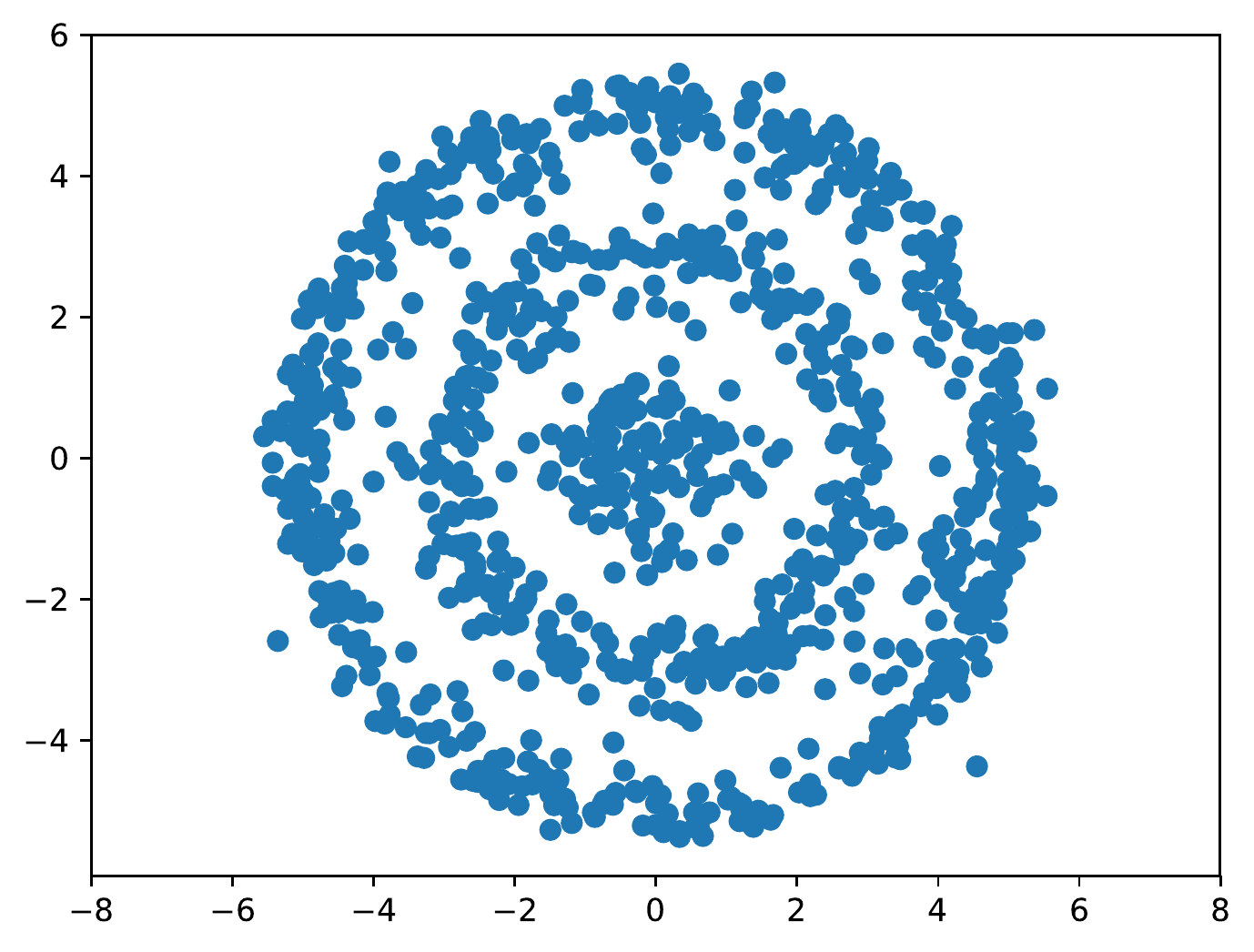} \\
  \includegraphics{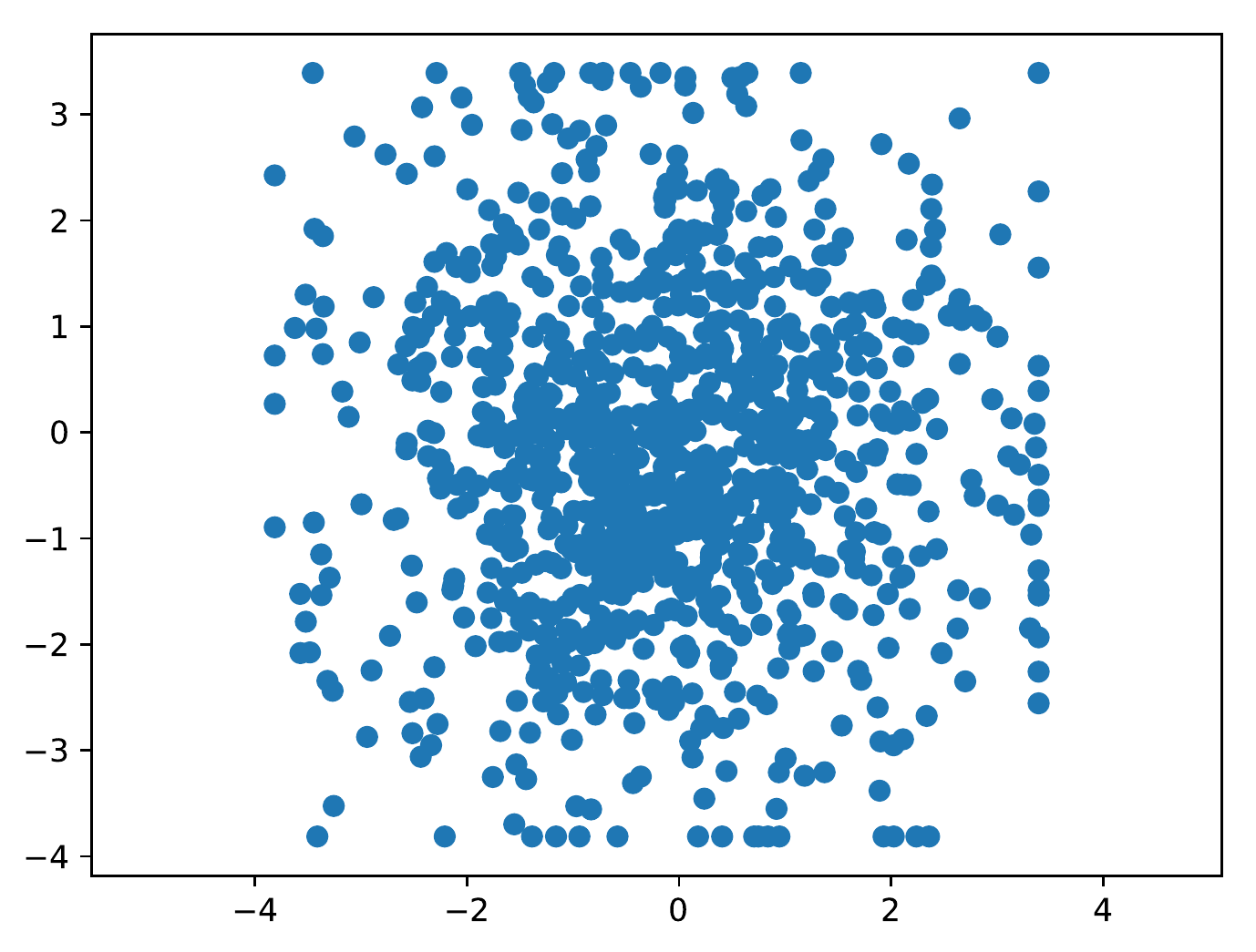} & 
  \includegraphics{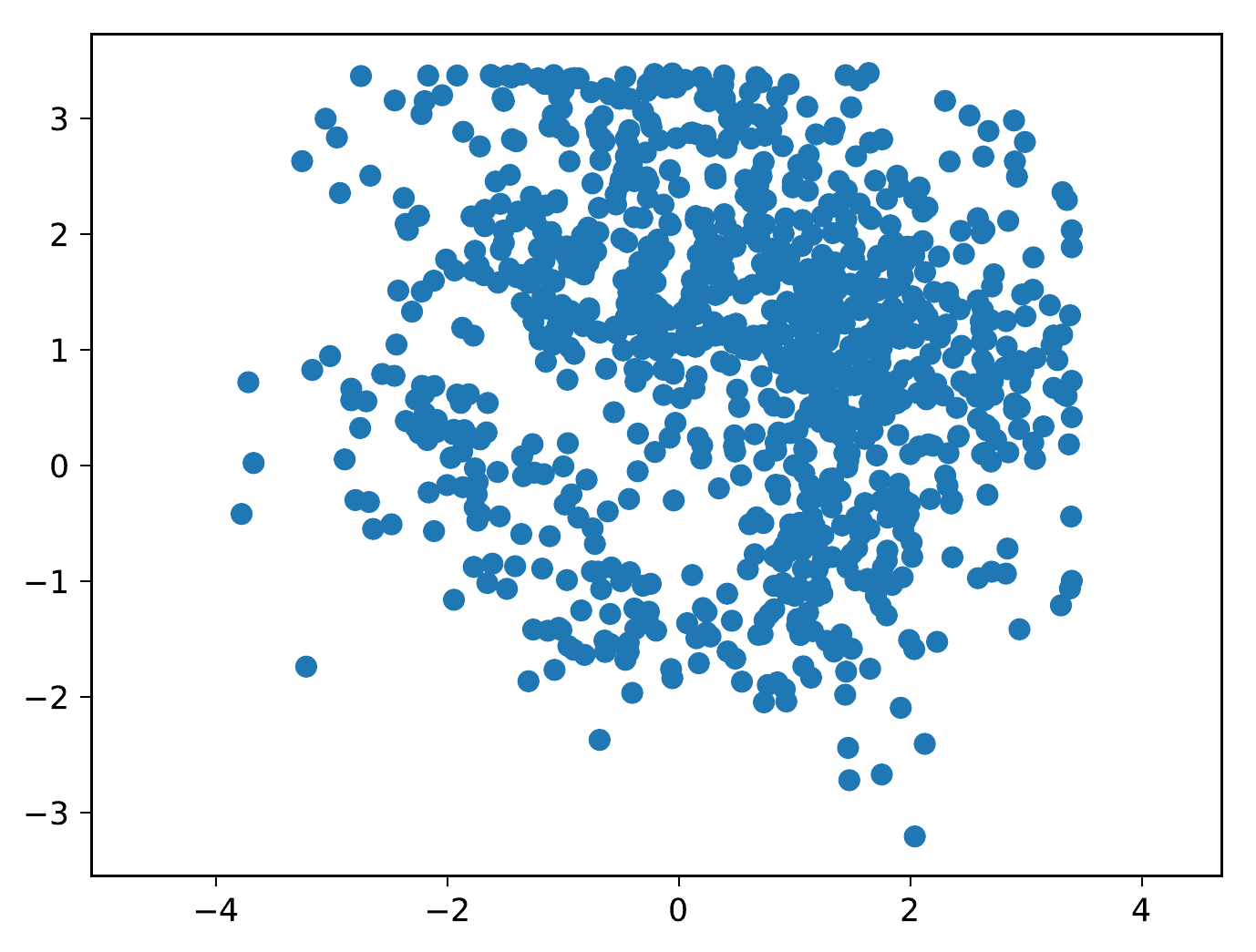} & 
  \includegraphics{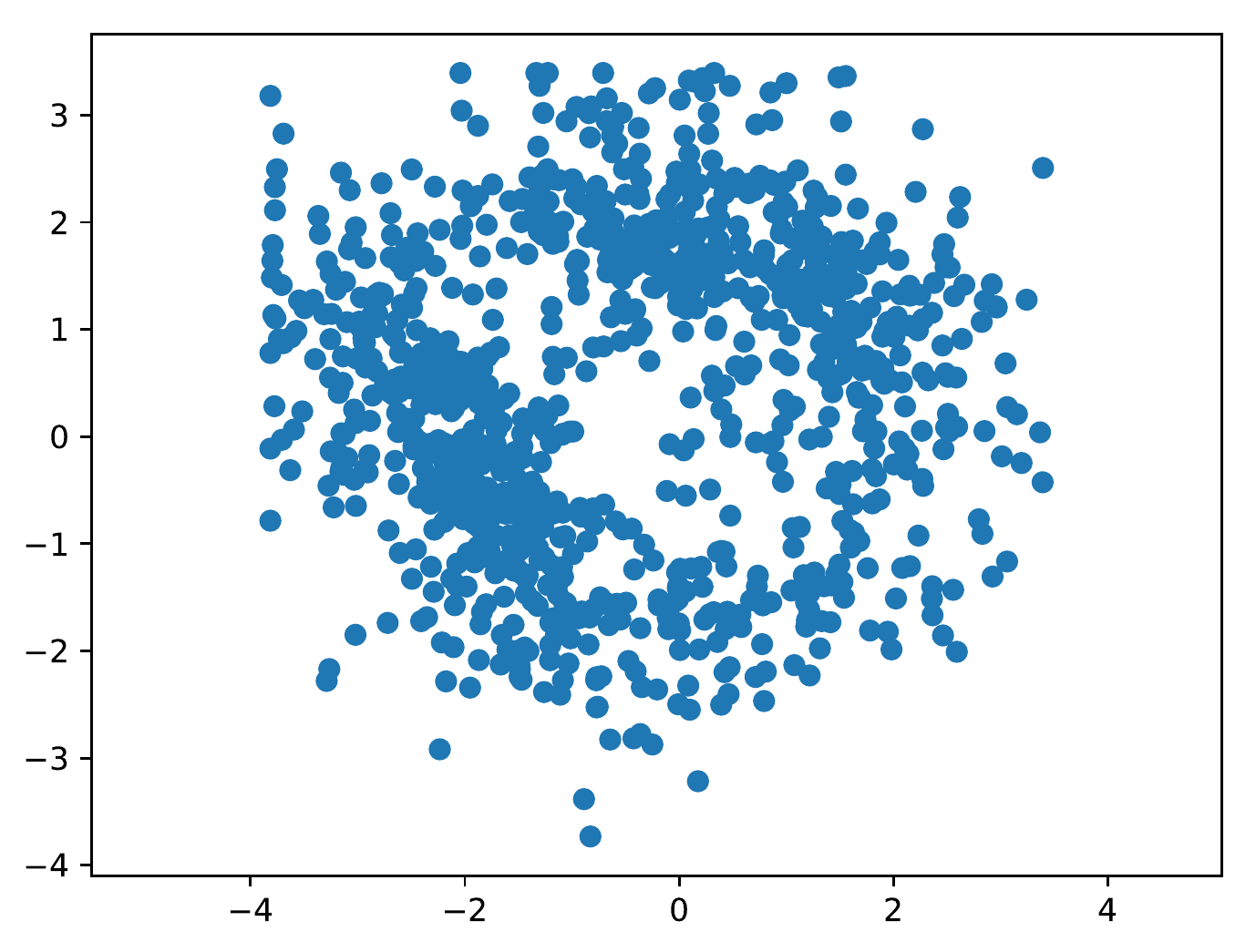} & 
  \includegraphics{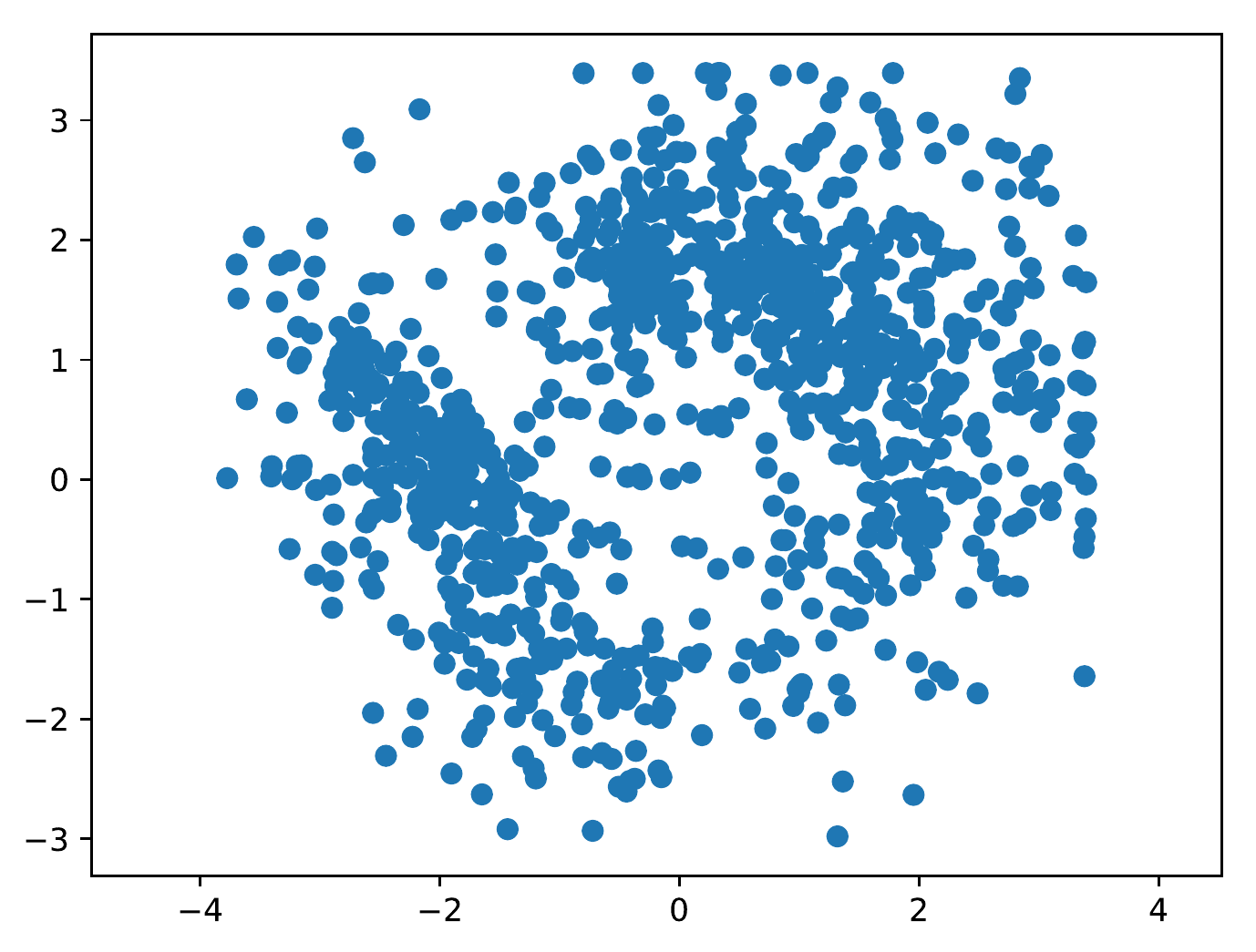} & 
  \includegraphics{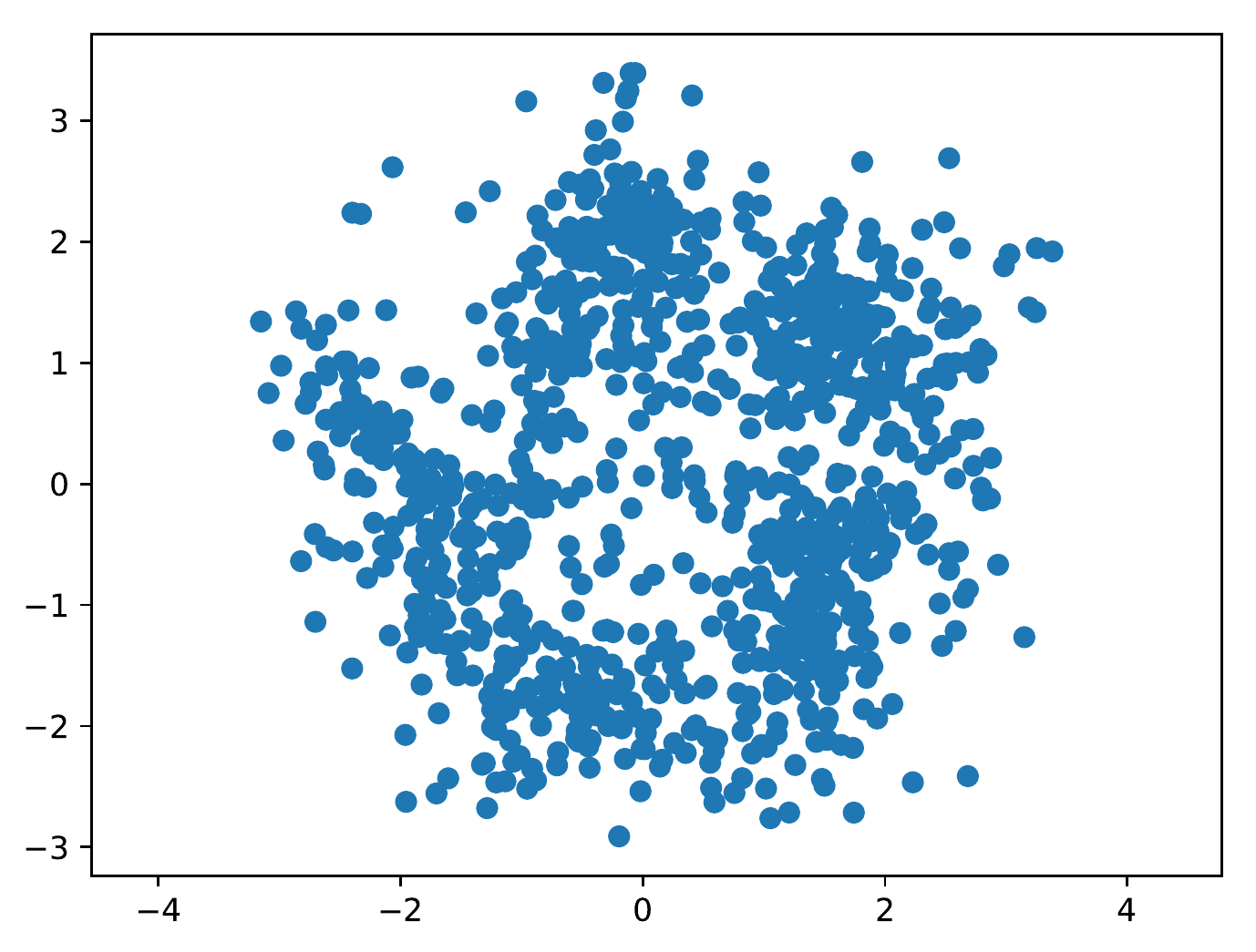} & 
  \includegraphics{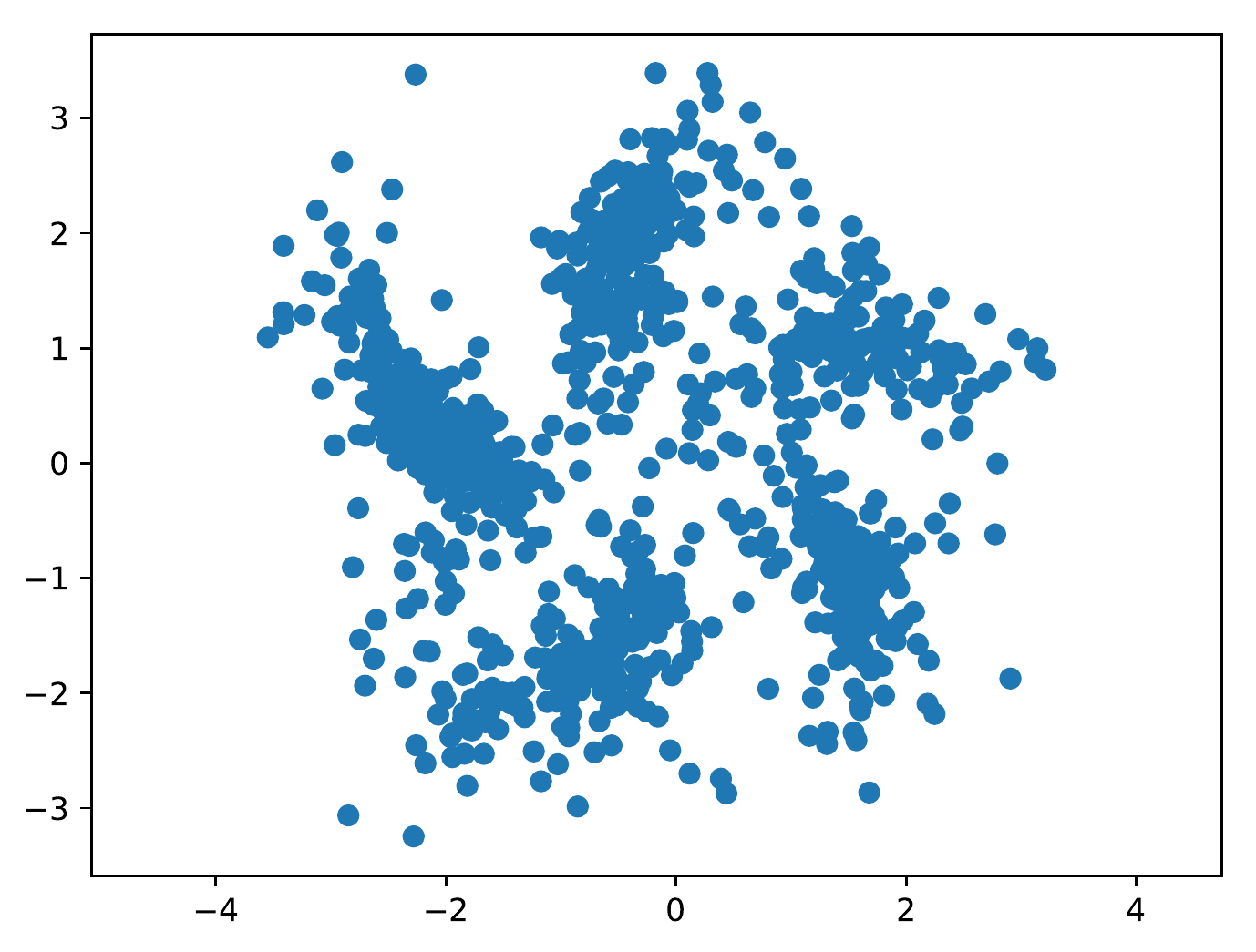} & 
  \includegraphics{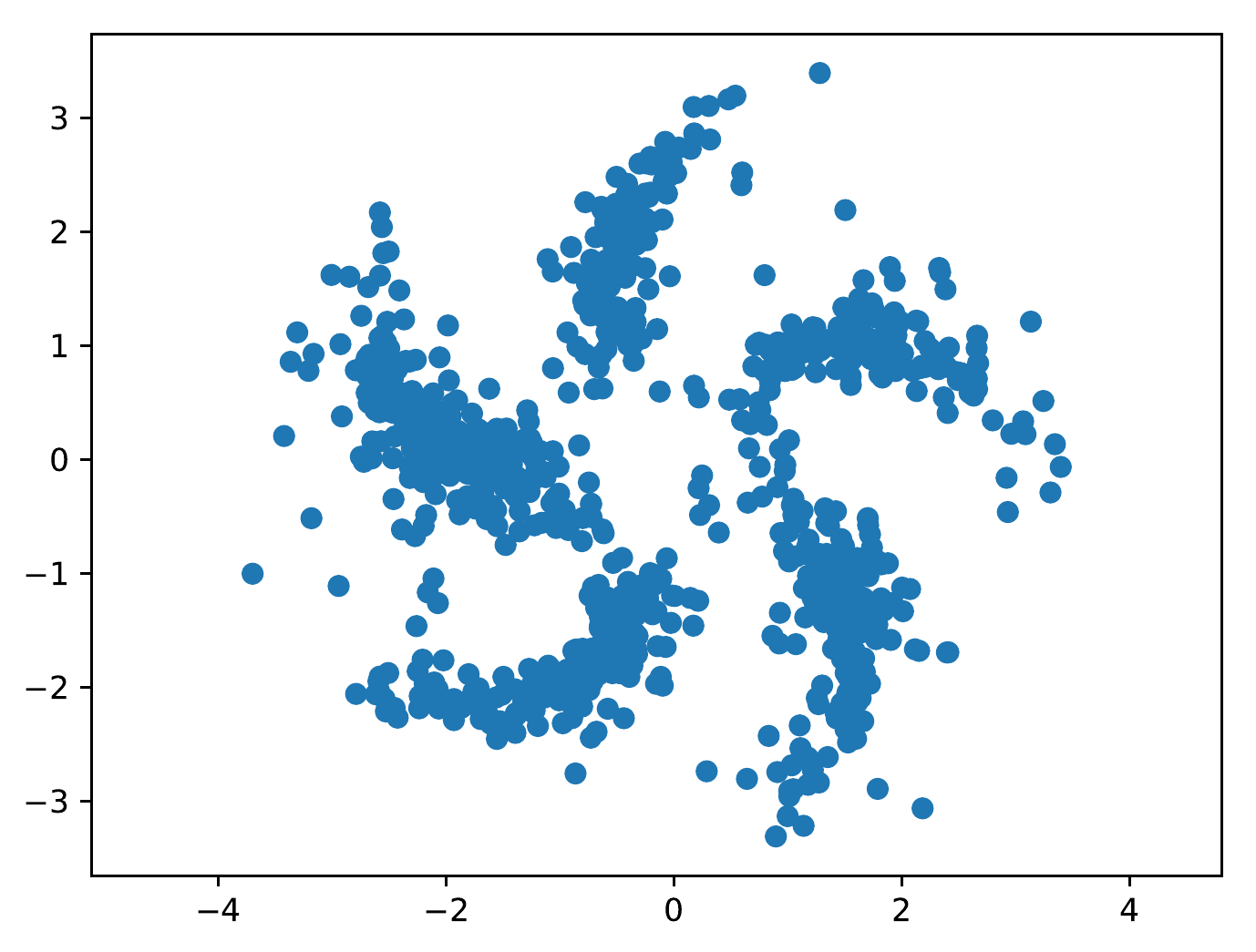} & 
  \includegraphics{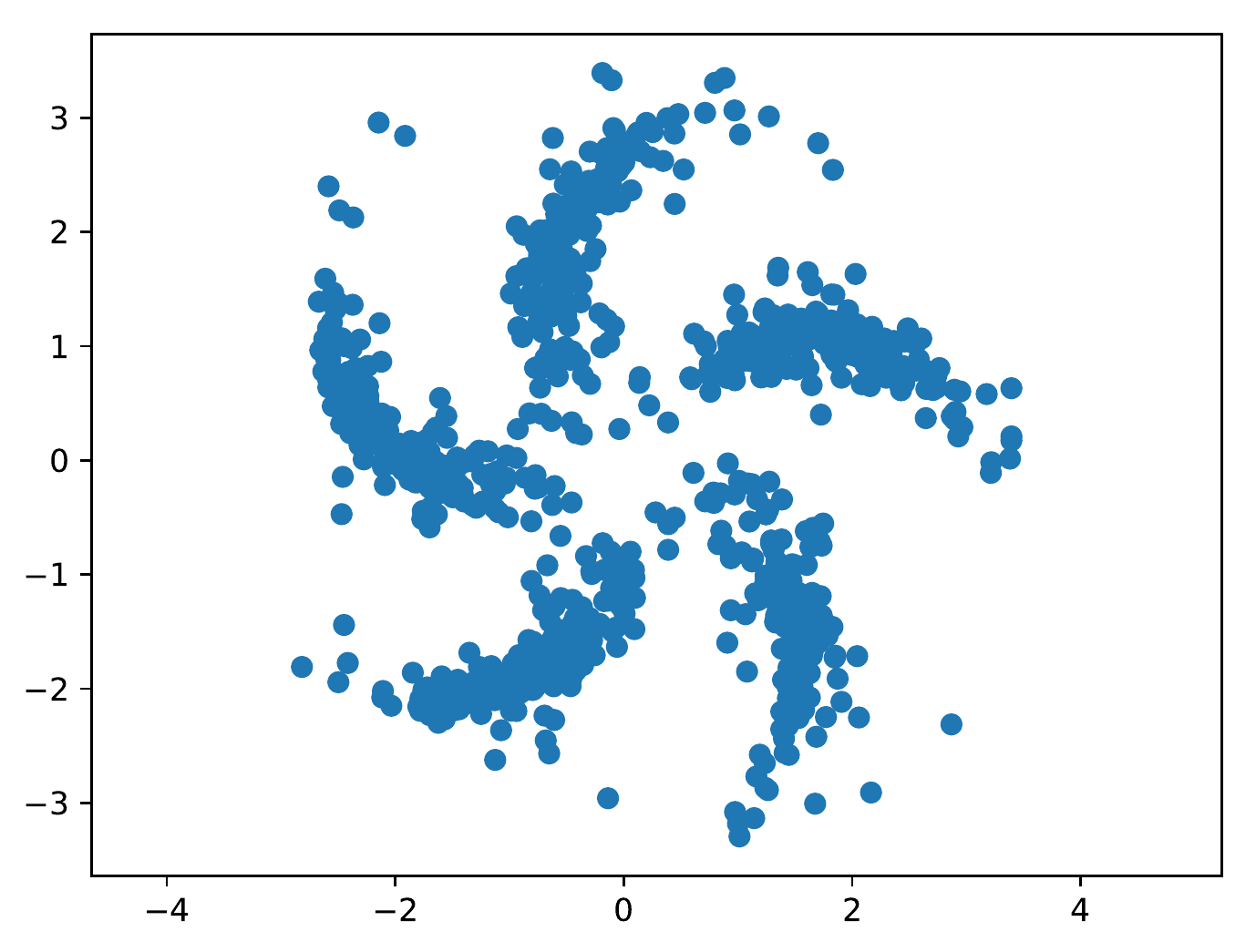} \\
  \end{tabular}
}
  \caption{Convergence behavior of sampler on \texttt{moons, Multiring, pinwheel} synthetic datasets.  \label{fig:synthetic_sampler}}
\end{figure*}

In~\figref{fig:synthetic}, we visualize the learned distribution using both the learned dual sampler and the unnormalized exponential model on several synthetic datasets. Overall, the sampler almost perfectly recovers the distribution, and the learned $f$ captures the landscape of the distribution. We also plot the convergence behavior in~\figref{fig:synthetic_sampler}. We observe that the samples are smoothly converging to the true data distribution. As the learned sampler depends on $f$, this figure also indirectly suggests good convergence behavior for $f$. More results for the learned models can be found in~\figref{fig:more_synthetic} in~\appref{sec:more_exp_results}.

A quantitative comparison in terms of the MMD~\citep{GreBorRasSchetal12} of the samplers is in~\tabref{tab:synthetic_mmd}. To compute the MMD, for NF and~\algshort, we use 1,000 samples from their sampler with Gaussian kernel. The kernel bandwidth is chosen using median trick~\citep{DaiHeDaiSon16}. For SM, since there is no such sampler available, we use vanilla HMC to get samples from the learned model $f$, and use them to estimate MMD~\citep{DaiDaiGreSon18}. As we can see from~\tabref{tab:synthetic_mmd},~\algshort~obtains the best MMD in most cases, which demonstrates the flexibility of dynamics embedding compared to normalizing flow, and the effectiveness of adversarial training compared to SM and CD.

\subsection{Real-world Image Datasets}\label{sec:real_exp}

\begin{figure*}[t]
\centering
\resizebox{1.02\textwidth}{!}{%
\centering
  \begin{tabular}{cccc}
  \includegraphics[width=0.48\textwidth]{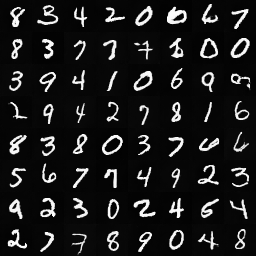} &
  \hspace{-4mm}
  \includegraphics[width=0.50\textwidth, height=0.48\textwidth, trim={0.8cm 0.8cm 1cm 0.8cm}, clip]{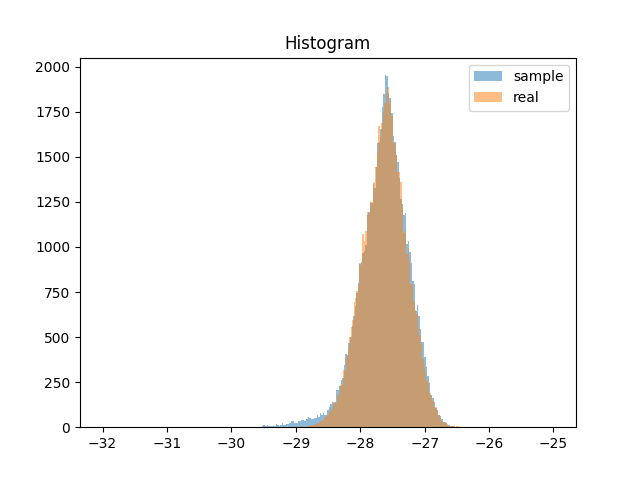} &
  \hspace{-4mm}
  \includegraphics[width=0.48\textwidth]{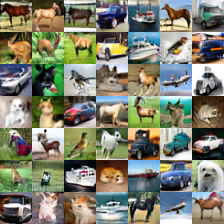} & 
  \hspace{-3mm}
  \includegraphics[width=0.50\textwidth, height=0.48\textwidth, trim={0.8cm 0.8cm 1cm 0.8cm}, clip]{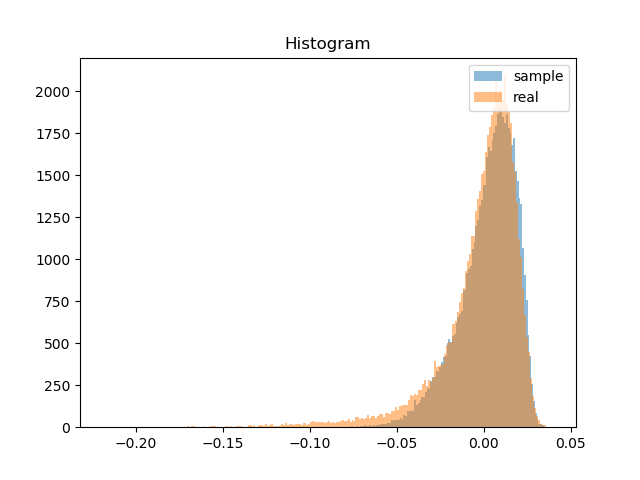} \\
  (a) Samples on \texttt{MNIST} & 
  (b) Histogram on \texttt{MNIST} & 
  (c) Samples on \texttt{CIFAR-10} & 
  (d) Histogram on \texttt{CIFAR-10}
  \end{tabular}
  }
  \caption{{The generated images on \texttt{MNIST} and \texttt{CIFAR-10} and the comparison between energies of generated samples and real images. The blue histogram illustrates the distribution of $f\rbr{x}$ on generated samples, and the orange histogram is generated by $f\rbr{x}$ on testing samples. As we can see, the learned potential function $f\rbr{x}$ matches the empirical dataset well.}\label{fig:real_images}}
\end{figure*}

We apply~\algshort~to \texttt{MNIST} and \texttt{CIFAR-10} data. In both cases, we use a CNN architecture for the discriminator, following~\citet{MiyKatKoyYos18}, with spectral normalization added to the discriminator layers. In particular, for the discriminator in the \texttt{CIFAR-10} experiments, we replace all downsampling operations by average pooling, as in~\citet{DuMor19}. We parametrize the initial distribution $p_0\rbr{x, v}$ with a deep Gaussian latent variable model~(Deep LVM), specified in~\appref{appendix:practical}. The output sample is clipped to $[0,1]$ after each HMC step and the Deep LVM initialization. The detailed architectures and experimental configurations are described in~\appref{subsec:real_archi}. 
\begin{table}[H]
\centering
\caption{Inception scores of different models on \texttt{CIFAR-10} (unconditional).}
\label{tab:inception_score}
\begin{tabular}{cc}
  \toprule
  Model & Inception Score \\
  \midrule
  WGAN-GP~\citep{GulAhmArjDumetal17} & 6.50 \\
  Spectral GAN~\citep{MiyKatKoyYos18} & 7.42 \\
  \hline
  \hline
  Langevin PCD~\citep{DuMor19} & 6.02 \\
  Langevin PCD (10 ensemble)~\citep{DuMor19} & 6.78 \\
  \algshort: Deep LVM init w/o HMC & 7.26 \\
  \algshort: Deep LVM init w/ HMC & {\bf 7.55} \\
  \bottomrule
\end{tabular}
\end{table}
We report the inception scores in~\tabref{tab:inception_score}. For~\algshort, we train with Deep LVM as the initial $q_\theta^0$ with/without HMC steps for an ablation study. The HMC embedding greatly improves the performance of the samples generated by the initial $q_\theta^0$ alone. The proposed~\algshort~not only achieves better performance, compared to the fixed Langevin PCD for energy-based models reported in~\citep{DuMor19}, but also enables the generator to outperform the Spectral GAN.

We show some of the generated images in~\figref{fig:real_images}(a) and (c);
additional sampled images can be found in~\figref{fig:more_mnist} and~\ref{fig:more_cifar} in~\appref{sec:more_exp_results}. We also plot the potential distribution (unnormalized) of the generated samples and that of the real images for \texttt{MNIST} and \texttt{CIFAR-10}~(using 1000 data points for each) in~\figref{fig:real_images}(b) and (d). The energy distributions of both the generated and real images show significant overlap, demonstrating that the obtained energy functions have successfully learned the desired distributions. 

\begin{figure}[t]
\centering
  \includegraphics[width=\linewidth]{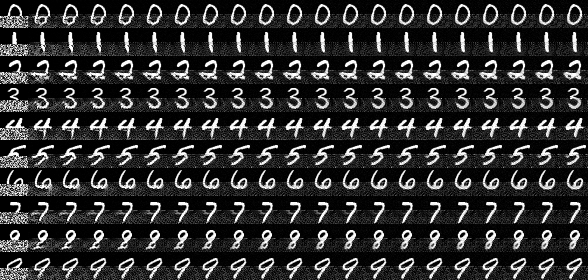} 
  \caption{Image completion with the~\algshort~learned model and sampler on~\texttt{MNIST}.}
  \label{fig:mnist_completion}
\end{figure}
Since \algshort~learns an energy-based model, the learned model and sampler can also be used for image completion. To further illustrate the versatility of~\algshort, we provide several image completions on \texttt{MNIST} in~\figref{fig:mnist_completion}. Specifically, we estimate the model with~\algshort~on fully observed images. For the input images, we mask the lower half with uniform noise. To complete the corrupted images, we perform the {learned dual sampler} steps to update the lower half of images with the upper half images fixed. We visualize the output from each of the $20$ HMC runs in~\figref{fig:mnist_completion}. Further details are given in~\appref{subsec:real_archi}.

\section{Conclusion}\label{sec:conclusion}

We proposed~\AlgName~(\algshort) to efficiently perform MLE with general exponential families. In particular, by utilizing the primal-dual formulation of the MLE for an augmented distribution with auxiliary kinetic variables, we incorporate the parametrization of the dual sampler into the estimation process in a fully differentiable way. This approach allows for shared parameters between the primal and dual, achieving better estimation quality and inference efficiency. We also established the connection between~\algshort~and existing estimators. Our empirical results on both synthetic and real data illustrate the advantages of the proposed approach.

\subsubsection*{Acknowledgments}

We thank Arnaud Doucet, David Duvenaud, George Tucker and the Google Brain team for helpful discussions, as well as the anonymous reviewers of NeurIPS 2019 for their insightful comments and suggestions. NH was supported in part by NSF-CRII-1755829, NSF-CMMI-1761699, and NCSA Faculty Fellowship. L.S. was supported in part by NSF grants CDS\&E-1900017 D3SC, CCF-1836936 FMitF, IIS-1841351, CAREER IIS-1350983 and Google Cloud.


\clearpage
\newpage

\appendix
\onecolumn

\begin{appendix}

\thispagestyle{plain}
\begin{center}
{\huge Appendix}
\end{center}

\section{Proof of Theorems in~\secref{sec:mle}}\label{appendix:derivation_sec_mle}

{\bf \thmref{thm:equivalent_mle}~(Equivalent MLE)}
\emph{The MLE of the augmented model is the same as the original MLE. }
\begin{proof}
The conclusion is straightforward from independence between $x$ and $v$. We rewrite the MLE~\eqref{eq:joint_MLE} in another way as follows
\begin{eqnarray}
\max_{f}L\rbr{f} &=& \widehat\EE_{x\sim\Dcal}\sbr{\log\int p\rbr{x, v} dv}\\
& = &\widehat\EE_{x\sim\Dcal}\sbr{\log \rbr{p\rbr{x} \int p\rbr{v}dv}}\\
& = &\widehat\EE_{x\sim\Dcal}\sbr{\log p\rbr{x}  + \underbrace{\log \int p\rbr{v}dv}_{\log 1= 0}} = \widehat\EE_{x\sim\Dcal}\sbr{\log p\rbr{x}},
\end{eqnarray}
where the second equation comes from the definition of the $p\rbr{x, v}$ in~\eqref{eq:joint_hmc} with independent $x$ and $v$. 
\end{proof}

{\bf \thmref{thm:grad_flow}~(HMC embeddings as gradient flow)}
\emph{
For a continuous time with infinitesimal stepsize $\eta\rightarrow 0$, the density of the particles $\rbr{x^t, v^t}$, denoted as $q_t\rbr{x, v}$, follows Fokker-Planck equation
\begin{equation}
\frac{\partial q^t\rbr{x, v}}{\partial t} = \nabla\cdot\rbr{q^t\rbr{x, v}G\nabla \Hcal\rbr{x, v}},
\end{equation}
with $G = \begin{bmatrix} 0 & \Ib\\
-\Ib & 0\end{bmatrix}$. Then $q^t\rbr{x, v}\rightarrow p\rbr{x, v}\propto \exp\rbr{-\Hcal\rbr{x, v}}$ as $t\rightarrow\infty$. 
}
\begin{proof}
The first part of the theorem is trivial. When $\eta\rightarrow 0$, the HMC follows the dynamical system
$$
\sbr{\frac{dx}{dt}, \frac{dv}{dt}} = \sbr{\partial_v \Hcal\rbr{x, v}, -\partial_x \Hcal\rbr{x, v}} = G\nabla \Hcal\rbr{x, v}. 
$$
By applying the Fokker-Planck equation, we obtain 
\begin{equation}\label{eq:fp_equation}
\frac{\partial q^t\rbr{x, v}}{\partial t} = \nabla\cdot\rbr{q^t\rbr{x, v}G\nabla \Hcal\rbr{x, v}}.
\end{equation}
To show that the stationary distribution of such dynamical system converges to $p\rbr{x, v}\propto \exp\rbr{-\Hcal\rbr{x, v}}$, recall the fact that
\begin{equation}
\nabla\cdot\rbr{G\nabla q^t\rbr{x, v}} = -\partial_x\partial_v q^t\rbr{x, v} + \partial_v\partial_x q^t\rbr{x, v} = 0.
\end{equation}
The Fokker-Planck equation can be rewritten as 
\begin{equation}\label{eq:fp_equation_II}
\frac{\partial q^t\rbr{x, v}}{\partial t} = \nabla\cdot\rbr{q^t\rbr{x, v}G\nabla \Hcal\rbr{x, v} + G\nabla q^t\rbr{x, v}}.
\end{equation}
Substitute $p\rbr{x, v}\propto \exp\rbr{-\Hcal\rbr{x, v}}$ into~\eqref{eq:fp_equation_II} and notice 
$$
\exp\rbr{-\Hcal\rbr{x, v}}\nabla \Hcal\rbr{x, v} + \nabla \exp\rbr{-\Hcal\rbr{x, v}} = 0,
$$
we have $\partial p\rbr{x, v} = 0$, \ie, $p\rbr{x, v}$ is the stationary distribution. 
\end{proof}

{\bf \thmref{thm:hmc_density}~(Density value evaluation)}
{\itshape
If $\rbr{x^0, v^0}\sim q^0_{\theta}\rbr{x, v}$, after $T$ vanilla HMC steps~\eqref{eq:leapfrog}, we have 
$$
q^T\rbr{x^T, v^T} = q^0_{\theta}\rbr{x^0, v^0}.
$$
For the $\rbr{x^T, v^T}$ from the generalized leapfrog steps~\eqref{eq:generalized_leapfrog}, we have 
$$
q^T\rbr{x^T, v^T} =q^0_{\theta}\rbr{x^0, v^0}\prod_{t=1}^T\rbr{\Delta_x\rbr{x^t}\Delta_v\rbr{v^t}},
$$
where $\Delta_x\rbr{x^t}$ and $\Delta_v\rbr{v^t}$ are defined in~\eqref{eq:delta}.

For the $\rbr{x^T, \cbr{v^i}_{i=1}^T}$ from the stochastic Langevin dynamics~\eqref{eq:langevin} with $\rbr{x^0, \cbr{\xi^i}_{i=0}^{T-1}}\sim q_{\theta}^0\rbr{x, \xi}\prod_{i=1}^{T-1} q_{\theta_i}\rbr{\xi^i}$, we have
$$
q^T\rbr{x^T, \cbr{v^i}_{i=1}^T} = q_\theta^0\rbr{x^0, \xi^0}\prod_{i=1}^{T-1} q_{\theta_i}\rbr{\xi^i}.
$$
}

\begin{proof}
The claim can be obtained by simply applying the change-of-variable rule, \ie, 
$$
q^T\rbr{x^T, v^T} = q_\theta^0\rbr{x^0, v^0}\prod_{t=1}^{T}\abr{\det\nabla \Lb_{f, M}\rbr{x^t, v^t}}.
$$
The Jacobian of the transformation from $\rbr{x, v}$ to $\rbr{x, v^{-\frac{1}{2}}}$ is 
$
\begin{bmatrix}
\Ib & \zero\\
\frac{\eta}{2}\nabla_x^2 f\rbr{x} &\Ib
\end{bmatrix},
$
whose determinant is $1$. Similarly, the determinant of the Jacobian of the transform from $\rbr{x, v^{-\frac{1}{2}}}$ to $\rbr{x', v'}$ is also $1$. Therefore, $\abr{\det\rbr{\nabla \Lb_{f, M}\rbr{x^t, v^t}}} = 1$, $\forall i=1,\ldots, T$, and we prove the first claim. 

The second claim can also be obtained in a similar way. By simple algebraic manipulations, we have that the Jacobians of the transformation are all diagonal matrices. Thus,
\begin{eqnarray}\label{eq:delta}
\Delta_x\rbr{x^t} &=& \abr{\det\rbr{\diag\rbr{\exp\rbr{2S_v\rbr{\nabla_x f\rbr{x^t}, x^t}}}}},\nonumber\\
\Delta_v\rbr{v^t} &=& \abr{\det\rbr{\diag\rbr{\exp\rbr{ S_x\rbr{v^{\frac{1}{2}}}}}}}.  
\end{eqnarray}

Similarly, we calculate the Jacobian for the stochastic Langevin update. Specifically, during the $t$-th step, the Jacobian of the transformation from $\rbr{x^{t-1}, \cbr{v^i}_{i=1}^{t-1}, \xi^{t-1}}$ to $\rbr{x^{t-1}, \cbr{v^i}_{i=1}^{t-1}, v^t}$ is
$
\begin{bmatrix}
\Ib & \zero &\zero\\
\zero &\Ib &\zero\\
\frac{\eta}{2}\nabla_x^2 f\rbr{x} &\zero &\Ib
\end{bmatrix},
$
whose determinant is $1$. Similarly, the Jacobian of the transformation from $\rbr{x^{t-1}, \cbr{v^i}_{i=1}^{t-1}, v^t}$ to $\rbr{x^t, \cbr{v^i}_{i=1}^{t-1}, v^t}$ is
$
\begin{bmatrix}
\Ib & \zero &\zero\\
\zero &\Ib &\zero\\
\zero &\zero &\Ib
\end{bmatrix},
$
whose determinant is also $1$. Therefore $\abr{\det \rbr{\nabla \Lb_f\rbr{x^t, \cbr{v^i}_{i=1}^{t}}}} = 1$, which implies
$$
q^t\rbr{x^t, \cbr{v^i}_{i=1}^{t-1}, v^t} = q^{t-1}\rbr{x^{t-1}, \cbr{v^i}_{i=1}^{t-1}, \xi^{t-1}} = q^{t-1}\rbr{x^{t-1}, \cbr{v^i}_{i=1}^{t-1}}q_{\theta^{t-1}}\rbr{\xi^{t-1}}.
$$
Apply the same argument for $\forall t=1, \ldots, T$, we obtain the third claim. 
\end{proof}

\section{Variants of Dynamics Embedding}\label{appendix:variants}

Besides the vanilla Hamiltonian/Langevin embedding and its generalized version we introduced in the main text, we can also embed alternative dynamics, \ie, deterministic Langevin dynamics and its continuous and generalized version.

\subsection{Deterministic Langevin Embedding}\label{subsec:langevin_embedding}

We embed the \emph{deterministic Langevin dynamics} to form $x' = \Lb_{f, M}\rbr{x}$ as $x' = x + \eta \nabla_x f\rbr{x}$ with $x^0\sim q_\theta^0\rbr{x}$.  By the change-of-variable rule, we have 
$q^T_{f, M}\rbr{x^T} = q_\theta^0\rbr{x_0}\prod_{t=1}^T\abr{\det\frac{\partial x^{t}}{\partial x^{t-1}}}$. The deterministic Langevin embedding has been exploited in variational auto-encoder~\citep{DaiDaiHeLiuetal18}, in which the variational technique has been applied to bypass the calculation of $\prod_{t=1}^T\abr{\det\frac{\partial x^{t}}{\partial x^{t-1}}}$.

Plug such parametrization of the dual distribution into~\eqref{eq:primal_CD}, we achieve the alternative objective 
\begin{equation}\label{eq:langevin_param}
\max_{f\in\Fcal} \min_{\theta, M, \eta} \ell\rbr{f; \theta, M, \eta} \defeq \widehat\EE_\Dcal\sbr{f}-\EE_{{x^0}\sim q_\theta^0\rbr{x}}\sbr{f\rbr{x^T} - \log q_\theta^0\rbr{x} - \sum_{t=1}^T\log \abr{\det\frac{\partial x^t}{\partial x^{t-1}}}}.
\end{equation}
For the $\log$-determinant term, $\log \abr{\det\frac{\partial x^t}{\partial x^{t-1}}} = \log \abr{\det \rbr{ I + \eta \Hb^f\rbr{x^t}}}$, where $\Hb^f_{i, j} = {\frac{\partial^2 f\rbr{x}}{\partial x_i \partial x_j}}$. Then, the gradient $\frac{\partial \log \abr{\det \rbr{ I + \eta \Hb^f\rbr{x^t}}}}{\partial f}= \eta\tr\rbr{\rbr{I + \eta\Hb^f\rbr{x_t}}^{-1}\frac{\partial H^f\rbr{x^t}}{\partial f}}$. However, the computation of the $\log$-determinant and its derivative w.r.t. $f$ are expensive. We can apply the polynomial expansion to approximate it. 

Denoting $\delta$ as the bound of the spectrum of $\Hb^f\rbr{x^t}$ and $C \defeq \frac{\eta\delta}{1 + \eta\delta} I - \frac{1}{1 + \eta\delta}\Hb^f\rbr{x^t}$, we have $\lambda\rbr{C}\in \rbr{-1, 1}$. Then, 
$$
\log \abr{\det \rbr{ I + \eta \Hb^f\rbr{x^t}}} = d\log\rbr{1 + \eta\delta} + \tr\rbr{\log\rbr{I-C}}.
$$
We can apply Taylor expansion or Chebyshev expansion to approximate the $ \tr\rbr{\log\rbr{I-C}}$. Specifically, we have
\begin{itemize}
  \item {Stochastic Taylor Expansion}~\citep{BouDriKamKonetal17} Recall $\log\rbr{1 - x} = -\sum_{k=1}^\infty \frac{x^k}{k}$, we have the Taylor expansion 
  $$
  \tr\rbr{\log\rbr{I-C}} = -\sum_{i=1}^k\frac{\tr\rbr{C^i}}{i}.
  $$
  To avoid the matrix-matrix multiplication, we further approximate the $\tr\rbr{C} = \EE_z\sbr{z^\top C z}$ with $z$ as Rademacher random variables, \ie, Bernoulli distribution with $p=\frac{1}{2}$. 

  Particularly, if we set $i=1$, recall the $\tr\rbr{\Hb^f\rbr{x}} = \nabla_x^2 f\rbr{x}$, we can directly calculate without the Hutchinson approximation. 

  \item {Stochastic Chebyshev Expansion}~\citep{HanMalShi15} We can approximate with Chebyshev polynomial, \ie, 
  $$
  \tr\rbr{\log\rbr{I-C}} = \sum_{i=1}^k c_i \tr\rbr{R_i\rbr{C}},
  $$
  where $R\rbr{\cdot}$ denotes the Chebshev polynomial as $R_i\rbr{x} = 2x R_{i-1}\rbr{x} - R_{i-2}\rbr{x}$ with $R_1\rbr{x} = x$ and $R_0\rbr{x} = 1$. The $c_i = \frac{2}{k+1}\sum_{j=0}^k\log\rbr{1-s_j}R_i\rbr{s_j}$ if $i\ge1$, otherwise $c_0 = \frac{1}{n+1}\sum_{j=0}^k\log\rbr{1-s_j}$ where $s_j = \cos\rbr{\frac{\pi\rbr{k + \frac{1}{2}}}{k+1}}$ for $j = 0, 1, \ldots, k$. 

  Similarly, we can use the Hutchinson approximation to avoid matrix-matrix multiplication. 

\end{itemize}

\subsection{Continuous-time Langevin Embedding}\label{subsec:cont_embedding}

We discuss several discretized dynamics embedding above. In this section, we take the continuous-time limit $\eta\rightarrow 0$ in the deterministic Langevin dynamics, \ie, $\frac{dx}{dt} = \nabla_x f\rbr{x}$. Follow the change-of-variable rule, we obtain
\begin{eqnarray*}
&&q\rbr{x'} = p\rbr{x}\det\rbr{I + \eta \Hb^f\rbr{x}}\\
&\Rightarrow& \log q\rbr{x'} - \log p\rbr{x} = -\tr\log\rbr{I + \eta \Hb^f\rbr{x}} = -\eta\nabla_x^2 f\rbr{x} +\Ocal\rbr{\eta^2}.
\end{eqnarray*}
As $\eta\rightarrow 0$, we have
\begin{equation}\label{eq:monge-ampere}
\frac{d \log q\rbr{x, t}}{dt} = -\nabla_x^2 f\rbr{x}.
\end{equation}

\noindent{\bf Remark (connections to Fokker-Planck equation)}
Consider the $\frac{dx}{dt} = \nabla_x f\rbr{x}$ as a SDE with zero diffusion term, by Fokker-Planck equation, we obtain the PDE w.r.t. $q\rbr{x, t}$ as
$$
\frac{\partial q\rbr{x, t}}{\partial t} = -\nabla\cdot\rbr{\nabla_x f\rbr{x} q\rbr{x, t}}.
$$
Alternatively, we can also derive the~\eqref{eq:monge-ampere} from the Fokker-Planck equation by explicitly writing the derivative. Specifically, 
\begin{eqnarray*}
\frac{d  q\rbr{x, t}}{dt} &=& \frac{\partial q\rbr{x, t}}{\partial x}\frac{\partial x}{\partial t} + \frac{\partial q\rbr{x, t}}{\partial t}\\
&=& \frac{\partial q\rbr{x, t}}{\partial x}\nabla_x f\rbr{x} -\nabla\cdot\rbr{\nabla_x f\rbr{x} q\rbr{x, t}}\\
&=& \frac{\partial q\rbr{x, t}}{\partial x}\nabla_x f\rbr{x} -{\nabla^2_x f\rbr{x} q\rbr{x, t}}- \nabla_x f\rbr{x}\frac{\partial q\rbr{x, t}}{\partial t}\\
&=&-{\nabla^2_x f\rbr{x} q\rbr{x, t}}.
\end{eqnarray*}
Therefore, we have
\begin{equation}
\frac{1}{q\rbr{x, t}}\frac{d  q\rbr{x, t}}{dt} = -{\nabla^2_x f\rbr{x} }\Rightarrow 
\begin{bmatrix}
\frac{d \log q\rbr{x, t}}{dt} = -\nabla_x^2 f\rbr{x}\\
\frac{dx}{dt} = \nabla_x f\rbr{x}
\end{bmatrix}
.
\end{equation}

Based on~\eqref{eq:monge-ampere}, we can obtain the samples and its density value by 
\begin{eqnarray}
\begin{bmatrix}
x^t\\
\log q\rbr{x^t}-\log p_\theta^0\rbr{x^0}
\end{bmatrix} = \int_{t_0}^{t_1}
\begin{bmatrix}
\nabla_x f\rbr{x\rbr{t}}\\
-\nabla_x^2 f\rbr{x(t)}
\end{bmatrix}
dt \defeq \Lb_{f, t_0, t_1}\rbr{x}.
\end{eqnarray}
We emphasize that this dynamics is different from the continuous-time flow proposed in~\citet{GraCheBetSutetal18}, where we have $\nabla_x^2 f\rbr{x}$ in the ODE rather than a trace operator, which requires one more Hutchinson stochastic approximation. We noticed that \citet{ZhaEWan18} also exploits the Monge-Amp\`{e}re equation to design the flow-based model for unsupervised learning. However, their learning algorithm is totally different from ours. They use the parameterization as a new flow and fit the model by matching a \emph{separate} distribution; while in our case, the exponential family and flow share the same parameters and match each other automatically.

We can approximate the integral using a numerical quadrature methods. One can approximate the $\nabla_{\rbr{f, t_0, t_1}}\ell\rbr{f; t_0, t_1}$ by the derivative through the numerical quadrature. 
Alternatively, we denote
$
g\rbr{t} = -\frac{\partial \ell\rbr{f, t_0, t_1}}{\partial x\rbr{t}},
$
by the adjoint method, the $\frac{\ell\rbr{f, t_0, t_1}}{\partial f}$ is also characterized by ODE
\begin{equation}
\frac{\partial \ell\rbr{f, t_0, t_1}}{\partial f} = \int_{t_0}^{t_1}- g\rbr{t}^\top {\nabla_f\cdot\nabla_x} f\rbr{x}dt,
\end{equation}
and can be approximated by numerical quadrature too.

We can combine the discretized and continuous-time Langevin dynamics by simply stacking several layers of $\Lb_{f, t_0, t_1}$. 

\subsection{Generalized Continuous-time Langevin Embedding}\label{subsec:generalized_cont_embedding}

We generalize the continuous-time Langevin dynamics by introducing more learnable space as
\begin{equation}\label{eq:generalized_langevin}
\frac{dx}{dt} = h\rbr{\xi_f\rbr{x}},
\end{equation}
where $h$ can be arbitrary smooth function and $\xi_f\rbr{x} = \rbr{\nabla_x f\rbr{x}, f\rbr{x}, x}$. We now derive the distributions formed by such flows following the change-of-variable rule, \ie, 
\begin{eqnarray*}
&&q\rbr{x'} = p\rbr{x}\det\rbr{I + \eta \nabla_x h\rbr{\xi_f\rbr{x}}}\\
&\Rightarrow& \log q\rbr{x'} - \log p\rbr{x} = -\tr\log\rbr{I + \eta \nabla_x h\rbr{\xi_f\rbr{x}} }= -\eta\tr\rbr{\nabla_x h\rbr{\xi_f\rbr{x}}} +\Ocal\rbr{\eta^2}.
\end{eqnarray*}
As $\eta\rightarrow 0$, we have
\begin{equation}
\frac{d\log q\rbr{x, t}}{dt} = -\tr\rbr{\nabla_x h\rbr{\xi_f\rbr{x}}}.
\end{equation}
Similarly, we can compute the samples and its density value by
\begin{eqnarray}
\begin{bmatrix}
x^t\\
\log q\rbr{x^t}-\log p_\theta^0\rbr{x^0}
\end{bmatrix} = \int_{t_0}^{t_1}
\begin{bmatrix}
h\rbr{\xi_f\rbr{x}}\\
-\tr\rbr{\nabla_x h\rbr{\xi_f\rbr{x}}}
\end{bmatrix}
dt \defeq \Lb_{f, t_0, t_1}\rbr{x}.
\end{eqnarray}

\section{Practical Algorithm}\label{appendix:practical}

In this section, we discuss several key components in the implementation of the~\algref{alg:mle_de}, including the gradient computation and the parametrization of the initialization $q_\theta\rbr{x,v}$. 

\subsection{Gradient Estimator}\label{subsec:grad_compute}

The gradient w.r.t. $f$ is illustrated in~\eqref{eq:hmc_grad_f}. The computation of the gradient needs to compute back-propagated through time, therefore, the computational cost is proportional to the number of sampling steps $T$. 

By Denskin's theorem~\citep{Bertsekas95}, if the samples $\rbr{x, v}$ from the optimal solution $p\rbr{x,v}\propto \exp\rbr{-\Hcal\rbr{x, v}}$, the third term in~\eqref{eq:hmc_grad_f} exactly vanish to zero, \ie,
\begin{equation}\label{eq:approx_hmc_grad_f}
\nabla_f \ell\rbr{f; \Theta} = \widehat\EE_\Dcal\sbr{\nabla_f f\rbr{x}} - \EE_{\rbr{x, v}\sim p\rbr{x, v}}\sbr{\nabla_f f\rbr{x}}, 
\end{equation}
whose computational cost is independent to $T$. 

Recall~\thmref{thm:grad_flow} that as $\eta\rightarrow 0$ and $T\rightarrow\infty$, the HMC embedding converges to the optimal solution. Therefore, we can approximate the BPTT estimator~\eqref{eq:hmc_grad_f} with the truncated gradient~\eqref{eq:approx_hmc_grad_f}. As $T$ increasing, the corresponding dual sampler approaches the optimal solution, and the truncation bias becomes smaller.

\subsection{Initialization Distribution Parametrization}\label{subsec:init_param}

In our algorithm, the dual distribution are parametrized via dynamics sampling method with an initial distribution $q_\theta^0\rbr{x, v}$, whose density value is available. There are several possible parametrization:
\begin{itemize}
  \item {\bf Flow-based model:} The most straightforward parametrization for $q_\theta^0\rbr{x, v}$ is utilizing flow-based model~\citep{RezMoh15,DinSohBen16,KinDha18}. For simplicity, we can decompose $q_\theta^0\rbr{x, v} = q_{\theta_1}^0\rbr{x}q_{\theta_2}^0\rbr{v}$ and parametrized both $q_{\theta_1}^0\rbr{x}$ and $q_{\theta_2}^0\rbr{v}$ separately. 
  \item {\bf Variants of deterministic Langevin embedding:} The expression ability of flow-based models is still restricted. We can exploit the deterministic Langevin embedding with separate potential function as the initialization. Specifically, we can also decompose $q_\theta^0\rbr{x, v} = q_{\theta_1}^0\rbr{x}q_{\theta_2}^0\rbr{v}$, for the sampler $x$, we exploit
  $$
  x^{t+1} =  x^t + \epsilon \phi^t\rbr{x^t}. 
  $$
  Although we do not have the explicit $\log q_{\theta_1}^0\rbr{x}$, we can approximate it via either Taylor expansion or Chebyshev expansion as~\secref{subsec:langevin_embedding}. It should be emphasized that in such parametrization, in each layer we use different $\phi^t$ for $t=\cbr{1,\ldots, T}$, which are all different from $\nabla_x f\rbr{x}$. 

  \item {\bf Deep latent variable model:} We can also consider the model
  \begin{eqnarray}
  v&\sim& q_{\theta_2}^0\rbr{v},\\
  x&=& \phi_{\theta_1}\rbr{v} + \epsilon,\quad \epsilon\sim \Ncal\rbr{0, \Sigma},
  \end{eqnarray}
  where $q_{\theta_2}^0\rbr{v}$ is some known distribution with $\theta_2$ as parameter and $\phi_{\theta_1}$ denotes the neural network with $\theta_1$ as parameter. Therefore, we have the distribution as
  $$
  q_{\theta}^0\rbr{x, v} = \Ncal\rbr{x; \phi_{\theta_1}^0\rbr{v}, \Sigma}q_{\theta_2}^0\rbr{v}.
  $$
  For vanilla HMC with leap-frog, the auxiliary variable $v$ should be the same size as $x$. However, for generalized HMC, the dimension of $v$ can be smaller than that of $x$. 

  \item {\bf Nonparametric model:} We can also prefix the $q^0\rbr{x, v} = q^0\rbr{x}q^0\rbr{v}$ without learning. Specifically, we set $q^0\rbr{x}$ as the empirical $p_\Dcal\rbr{x}$ and $q^0\rbr{v} = \Ncal\rbr{0, \Ib}$. Since the initial distribution is fixed, the learning objective~\eqref{eq:primal_dual_joint_MLE} reduces to
  \begin{equation}\label{eq:reduced_mle}
  \max_{f\in q}\min_\Theta \,\,\ell\rbr{f, \Theta}\propto\widehat\EE_\Dcal\sbr{f}-\EE_{\rbr{x^0, v^0}\sim q^0\rbr{x, v}}\sbr{f\rbr{x^T} -  \frac{1}{2}\nbr{v^T}_2^2 }. 
  \end{equation}
\end{itemize}

\section{Experiment Details}\label{sec:exp_details}

\subsection{Synthetic Experiments Details}\label{subsec:synthetic_exp_details}

We parametrize the potential function $f$ with fully connected multi-layer perceptron with 3 hidden layers. Each hidden layer has $128$ hidden units. We use ReLU to do the nonlinear activation in each hidden layer. We clip the norm of $\nabla_x f$ when updating $v$, and clip $v$ when updating $x$. The coefficient $\lambda$ in \eqref{eq:vanilla_hmc_param} is tuned in $\{0.1, 0.5, 1\}$. For the NF baseline, we tune the number of layers in $\{10, 15, 20\}$. For our~\algshort, we fix the number of normalizing flow layers to be $10$, and then perform at most $10$ steps of dynamics updates. So finally, the number of steps for sampling is comparable, while the~\algshort~maintains less memory cost. 

To make the training stable, we also tried several tricks, including:
\begin{enumerate}
  \item clip samples in HMC. This helps stabilize the training; We assume the final output has limited support over 2D space. 
  \item gradient penalty for $f(\cdot)$. We use a small penalty coefficient 0.01 for this, which is not very important though. 
  \item variance of proposal gaussian distribution. While we use 1 in general, a standard deviation of 0.5 would be more helpful in some cases. 
  \item penalty of momentum term in HMC. This is equivalent to the variance of prior of the latent variable we introduced. 
\end{enumerate}

The dataset generators are collect from several open-source projects~\footnote{\href{https://github.com/rtqichen/ffjord}{https://github.com/rtqichen/ffjord}.}~\footnote{\href{https://github.com/kevin-w-li/deep-kexpfam}{https://github.com/kevin-w-li/deep-kexpfam}.}. During training, we use this generator to generate the data from the true distribution on the fly. To get a quantitative comparison, we also generate 1,000 data samples for held-out evaluation. We illustrate the unnormalized model $\exp\rbr{c\cdot f}$ in~\figref{fig:synthetic} and~\ref{fig:more_synthetic}, where $c$ is a constant that is tuned within $[0.01, 10]$.

To compute the MMD, for NF and~\algshort, we use 1,000 samples from their sampler with Gaussian kernel. The kernel bandwidth is chosen using median trick~\citep{DaiHeDaiSon16}. For SM, since there is no such sampler available, we directly use vanilla HMC to get sample from the learned model $f$, and use them to estimate MMD. 

\paragraph{Parameter estimation experiments} In the experiment of recovering parameters of a given graphical model from data, we use high dimensional gaussian distribution with diagonal covariance. Here the energy function to be estimated $f(x) = -0.5 (x - \mu)^\top \Sigma^{-1} (x - \mu)$, where $\Sigma$ is a diagonal matrix. 

For our method, we use a 2-layer MLP as initial proposal with 3 of HMC steps afterwards. The step size in HMC is learned end-to-end. For CD, we use up to 15 steps of HMC, where the step size is adaptively adjusted according to the rejection rate. For all the methods, we average the parameters estimated in the last 5 epochs during training, and report the best results in this parameter estimation procedure.

\subsection{Real-world Experiments Details}\label{subsec:real_archi}

\begin{table*}[h]
  \centering

  \caption{Our architectures for both potential function $f\rbr{x}$ and initial dual sampler $p_\theta^0\rbr{x, v}$ used in \texttt{MNIST} and \texttt{CIFAR-10} experiments.}\label{tab:netarch}
  \begin{tabular}{ccc}
  \begin{tabular}{c}
  Potential function $f\rbr{\cdot}$\\
  \hline
  \hline
  3x3 conv, 64 \\
  \hline
  3x3 conv, 128 \\
  \hline
  2x2 avg pool \\
  \hline
  3x3 conv, 128 \\
  \hline
  3x3 conv, 256 \\
  \hline
  2x2 avg pool \\
  \hline
  3x3 conv, 256 \\
  \hline
  7x7 avg pool \\
  \hline
  fc, 256 $\rightarrow$ 1 \\
  \hline
  \\
  (a) Potential function $f\rbr{\cdot}$\\

  \end{tabular}
  & & 
  \begin{tabular}{c}
  \\
  \\
  Initial dual sampler\\
  \hline
  \hline
  fc, 512 $\rightarrow$ $4 \times 4 \times 512$ \\
  \hline
  Reshape to $4 \times 4$ Feature Map \\
  \hline
  2x2 Deconv, 256, stride 2 \\
  \hline
  2x2 Deconv, 128, stride 2 \\
  \hline
  2x2 Deconv, 64, stride 2 \\
  \hline
  3x3 Deconv, 3, stride 1 \\
  \hline
  \\
  (b) initial dual sampler
  \end{tabular}

  \end{tabular}

\end{table*}

We used the standard spectral normalization on the discriminator to stabilize the training process, and Adam with learning rate $10^{-4}$ and $\beta_1 = 0.0$ to optimize our model. For stability, we use a separate Adam optimizer for the hmc parameters and set the epsilon to $1e-5$. We trained the models with $200000$ iterations with batch size being $64$. For better performance, we used generalized HMC~\eqref{eq:generalized_leapfrog}, where we set $S_v(\cdot) = 0$, $S_x(\cdot) = 0$, $g_v(v) = \text{clip}(v, -0.01, 0.01)$ and $g_x(v^{1/2}) = v^{1/2}$. We fix $\eta$ to be $0.5$. The step sizes for our HMC sampler are independently learned for all HMC dimensions but shared among all time steps, and the values are all initialized to $10$. We set the number of HMC steps to $30$. The coefficient of the entropy regularization term is set to $10^{-5}$ and that of the $L_2$ regularization on the momentum vector in the last HMC step is set to $10^{-5}$.

We demonstrate the architectures of potential function $f$ and initial Deep LVM in~\tabref{tab:netarch}. A leaky ReLU follows each convolutional/deconvolutional layer in both the discriminator and generator. For the discriminator, we use spectral normalization for all layers in the discriminator. In addition, there is no activation function after the final fully-connected layer. For each deconvolution layer in the generator, we insert a batch normalization layer before passing the output to the leaky ReLU. 

We generate the image from the model and illustrated in~\figref{fig:real_images},~\figref{fig:more_mnist} and~\figref{fig:more_cifar}. We also compared in terms of inception score with other energy-model training algorithm and several state-of-the-art GAN algorithm in~\tabref{tab:inception_score}, where the~\algshort~achieves the best performances. Also, with simple importance sampling and proposal distribution being uniform distribution on $[-1, 1]^{n_d}$ ($n_d$ is the dimension of images), the log likelihood (in nats) on \texttt{CIFAR-10} is estimated to be around $2100$.

We also trained a non-parametric ADE on \texttt{MNIST} dataset for image completion to verify our algorithm. Specifically, we use with the same discriminator architecture used in parametric ADE for \texttt{MNIST}. The model is trained with fully observed images. We used generalized HMC~\eqref{eq:generalized_leapfrog}, where we set $S_v(v)$ being a learnable logit (so that $\exp(S_v(\cdot)) \in [0, 1]$), $g_v(v) = \text{clip}(v, -0.1, 0.1)$, $S_x(\cdot) = 0$ and $g_x(\cdot) = 1$. Both $S_v$ and $\eta$ will be learned, with $\eta$ initialized to $\sqrt{10}$ and $S_v$ initialized to a small number close to 0. We unfold $60$ steps of HMC in the dual samplers. As in~\citet{DuMor19}, we used a replay buffer of size $10000$. We added extra amount of noise into the dataset to make the training process more stable. We trained the model with Adam optimizer ($\beta_1 = 0.0, \beta_2 = 0.999$) for $60000$ iterations.

We tested the~\algshort~by image completion where we covered the lower half of images with uniform noise and used them as input to the learned HMC operators. We repeatedly apply the learned HMC with the learned model to lower half of these images for $20$ steps, with the upper half images fixed, and obtain $\text{HMC}^{(20)}(x_0; S_v, \eta)$.  We visualize the output from each of the $20$ HMC runs in~\figref{fig:mnist_completion}.

\section{More Experiment Results}\label{sec:more_exp_results}

\begin{figure*}[h]
\resizebox{1.0\textwidth}{!}{%
\centering
  \begin{tabular}{ccccccc}
    \includegraphics[width=0.14\textwidth]{figs/synthetic_v2/2spirals/2spirals-none} & 
    \includegraphics[width=0.14\textwidth]{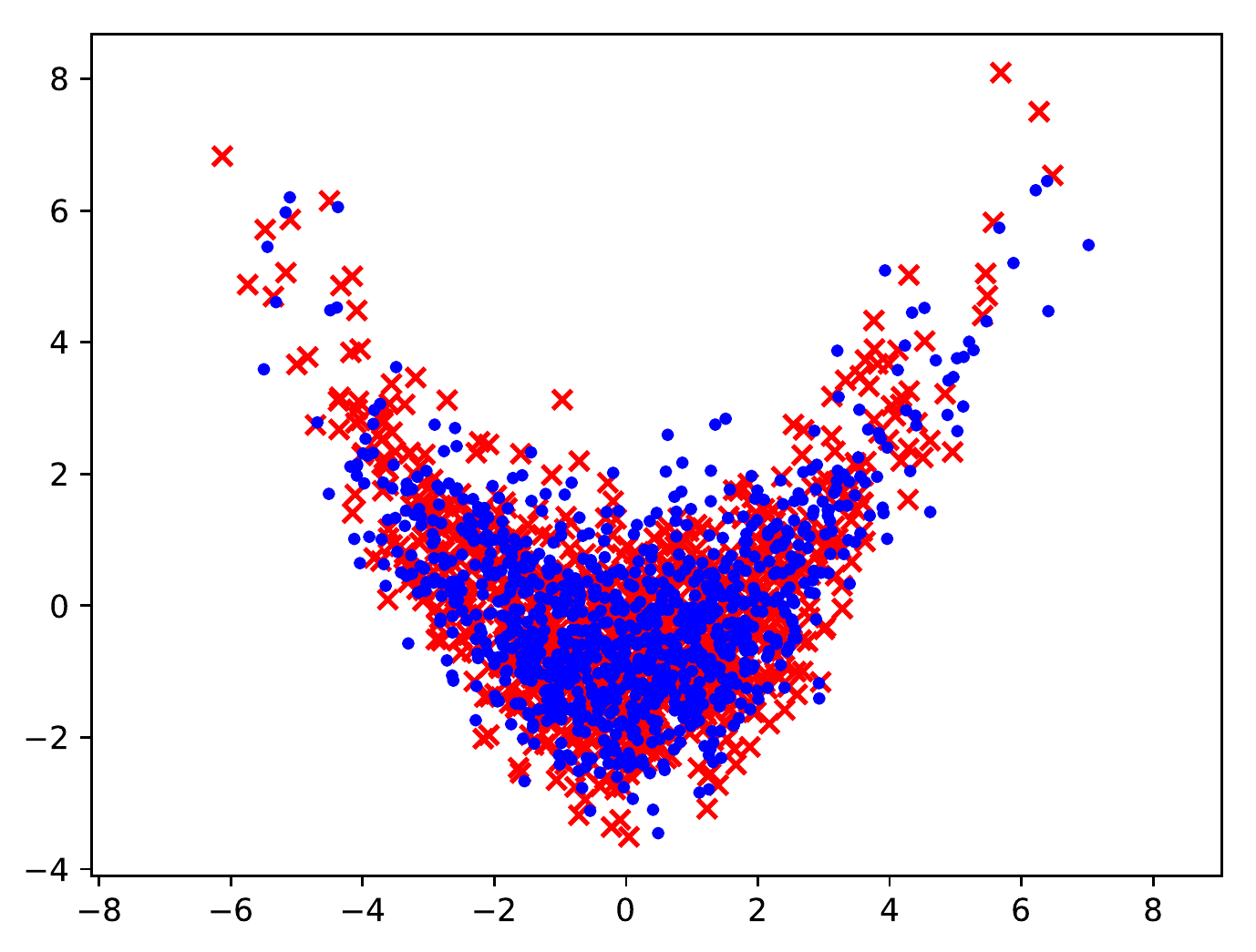} & 
    \includegraphics[width=0.14\textwidth]{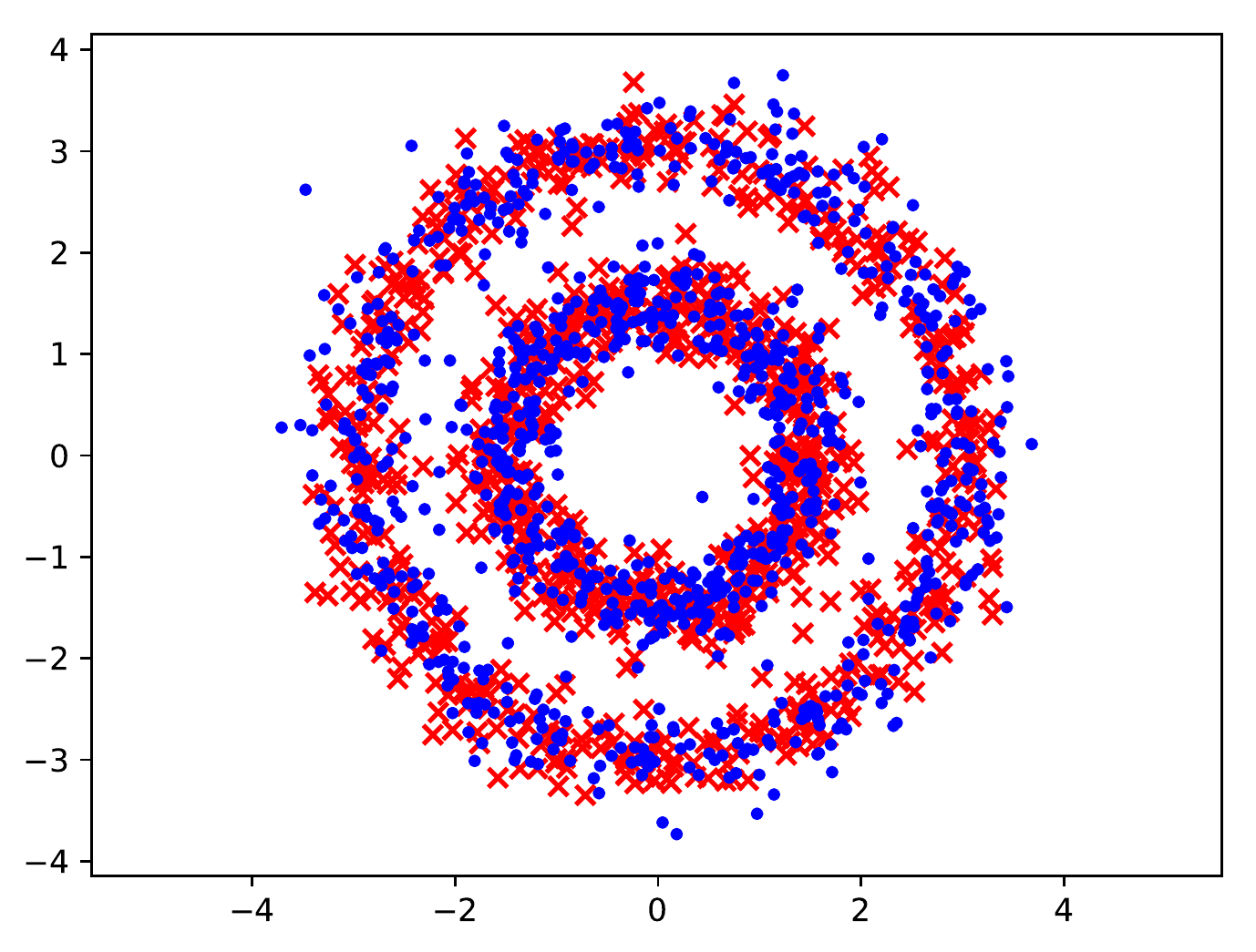} &
    \includegraphics[width=0.14\textwidth]{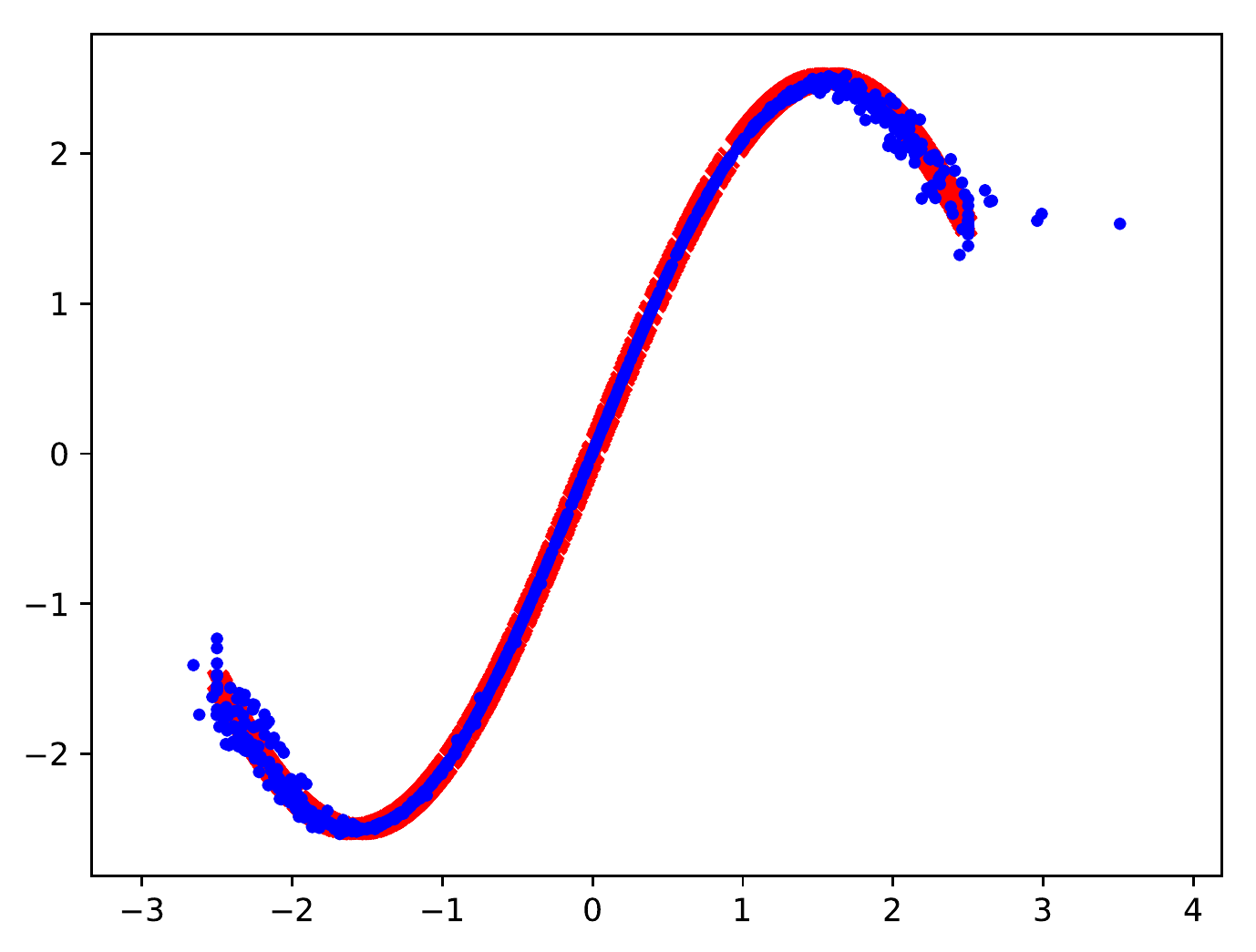} & 
    \includegraphics[width=0.14\textwidth]{figs/synthetic_v2/Cosine/Cosine-none} &
    \includegraphics[width=0.14\textwidth]{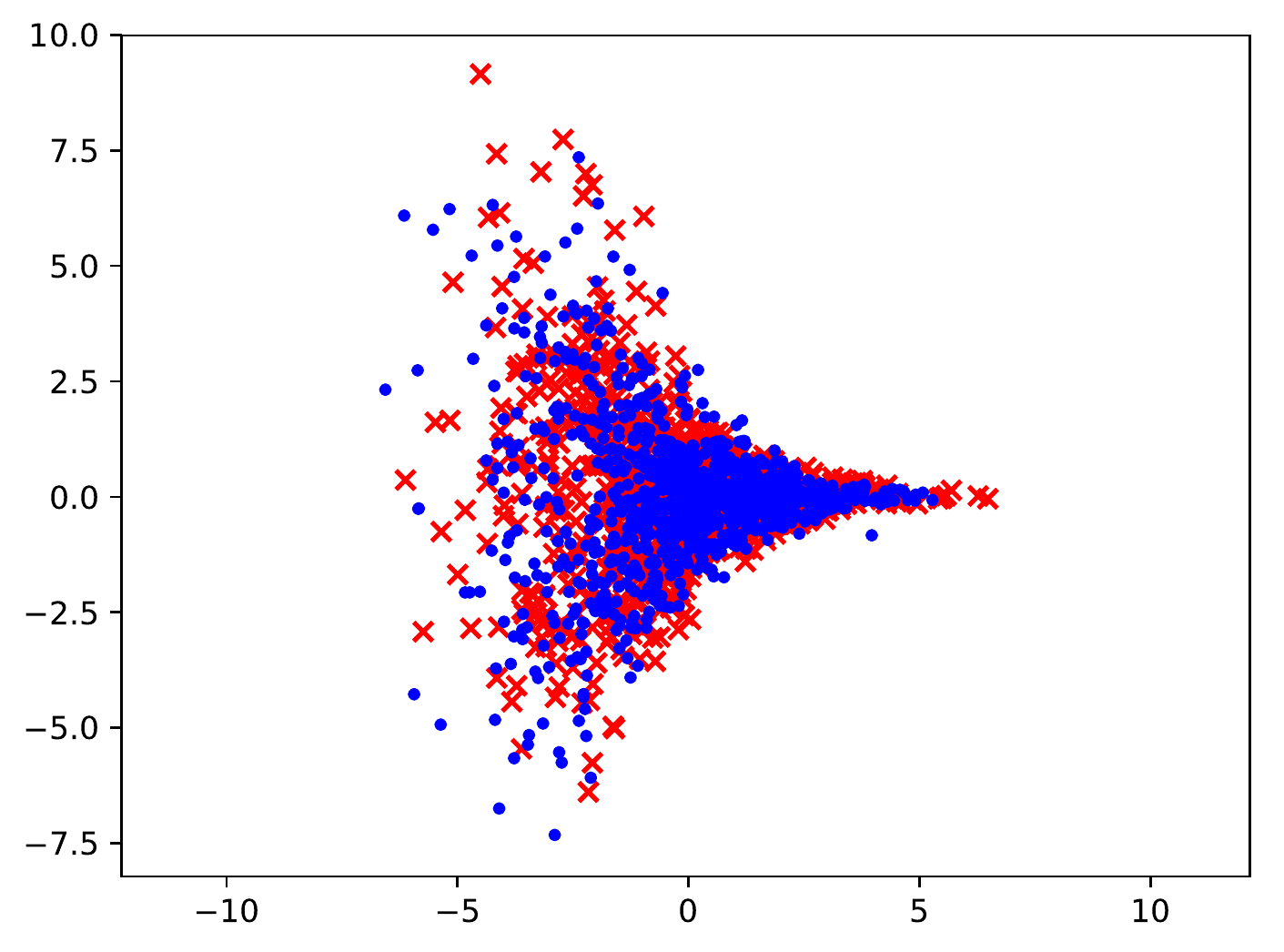} & 
    \includegraphics[width=0.14\textwidth]{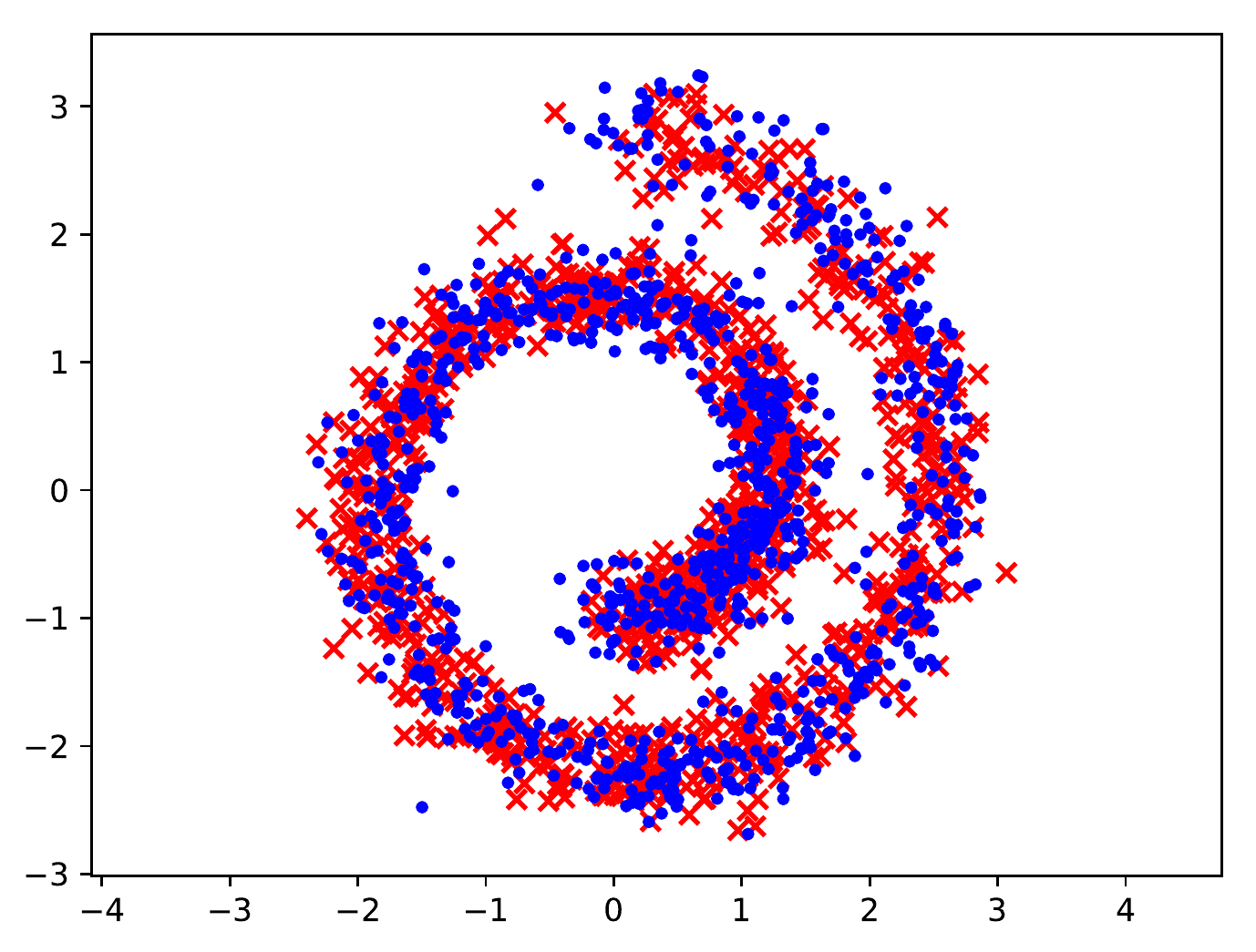} \\

    \tikz[baseline=(a.north)]\node[yscale=-1,inner sep=0,outer sep=0](a){\includegraphics[width=0.14\textwidth, trim={2cm 0cm 2cm 1cm},clip]{figs/synthetic_v2/2spirals/heat-93-0_2}}; &
    \tikz[baseline=(a.north)]\node[yscale=-1,inner sep=0,outer sep=0](a){\includegraphics[width=0.14\textwidth, trim={2cm 0cm 2cm 1cm},clip]{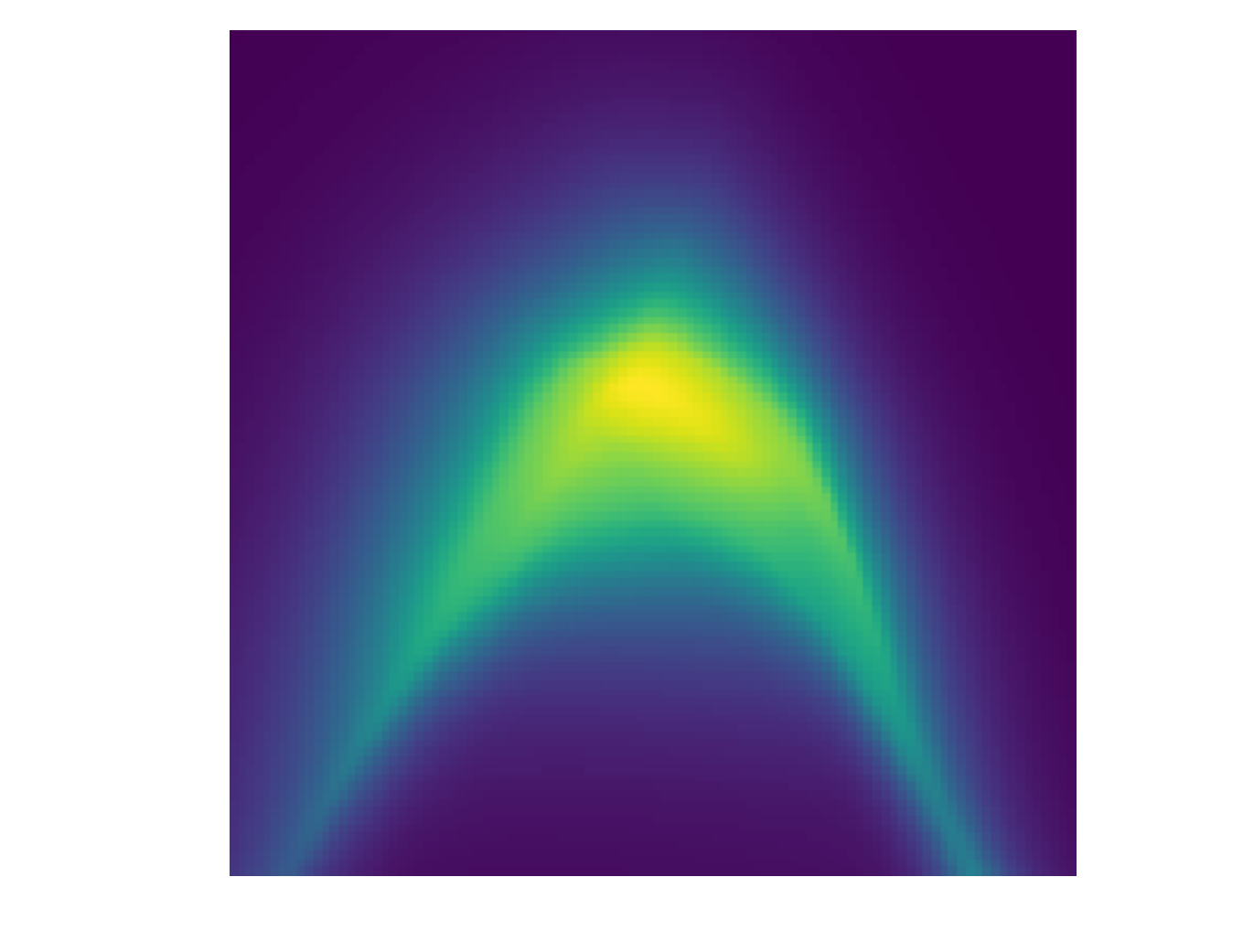}}; &
    \tikz[baseline=(a.north)]\node[yscale=-1,inner sep=0,outer sep=0](a){\includegraphics[width=0.14\textwidth, trim={2cm 0cm 2cm 1cm},clip]{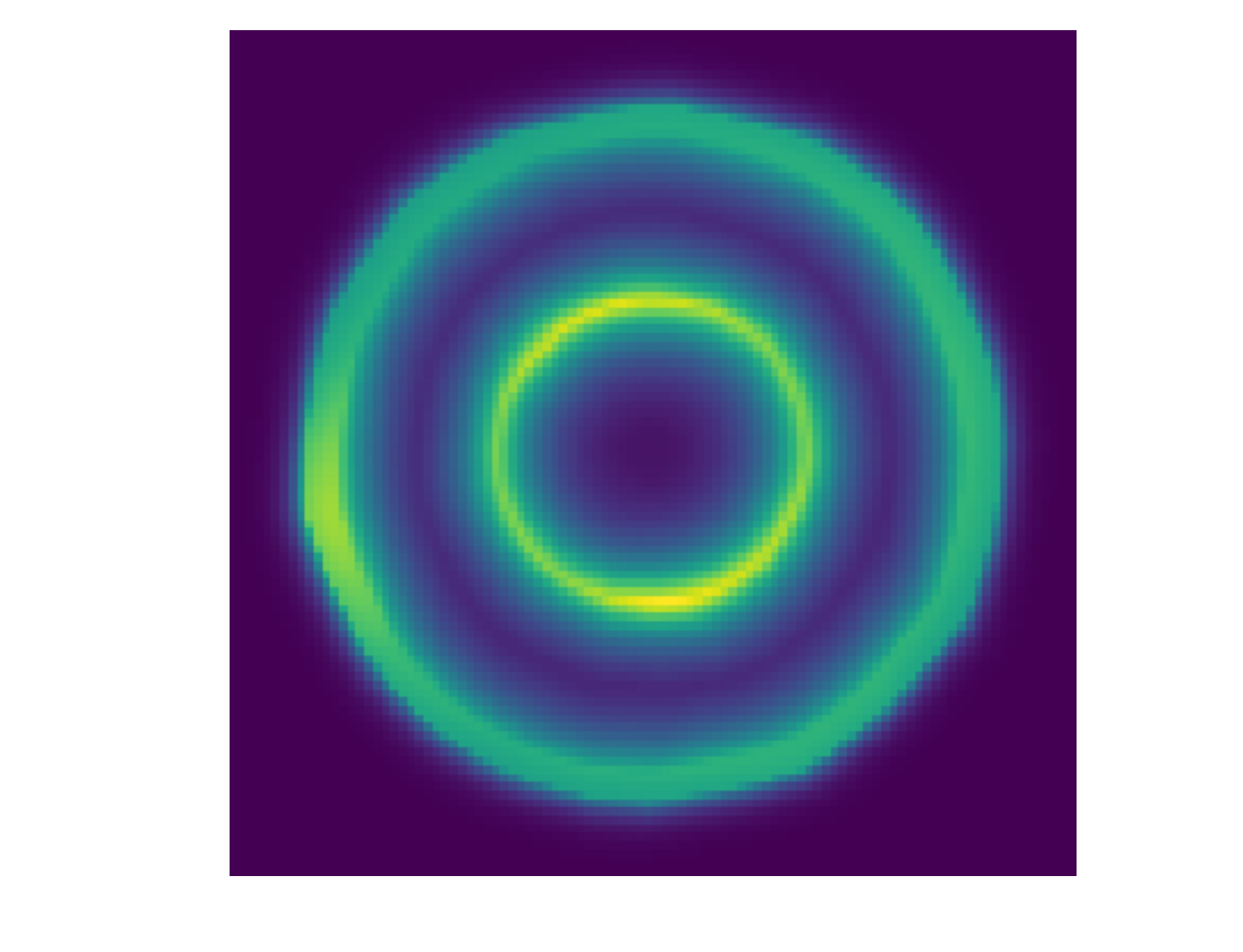}}; &
    \tikz[baseline=(a.north)]\node[yscale=-1,inner sep=0,outer sep=0](a){\includegraphics[width=0.14\textwidth, trim={2cm 0cm 2cm 1cm},clip]{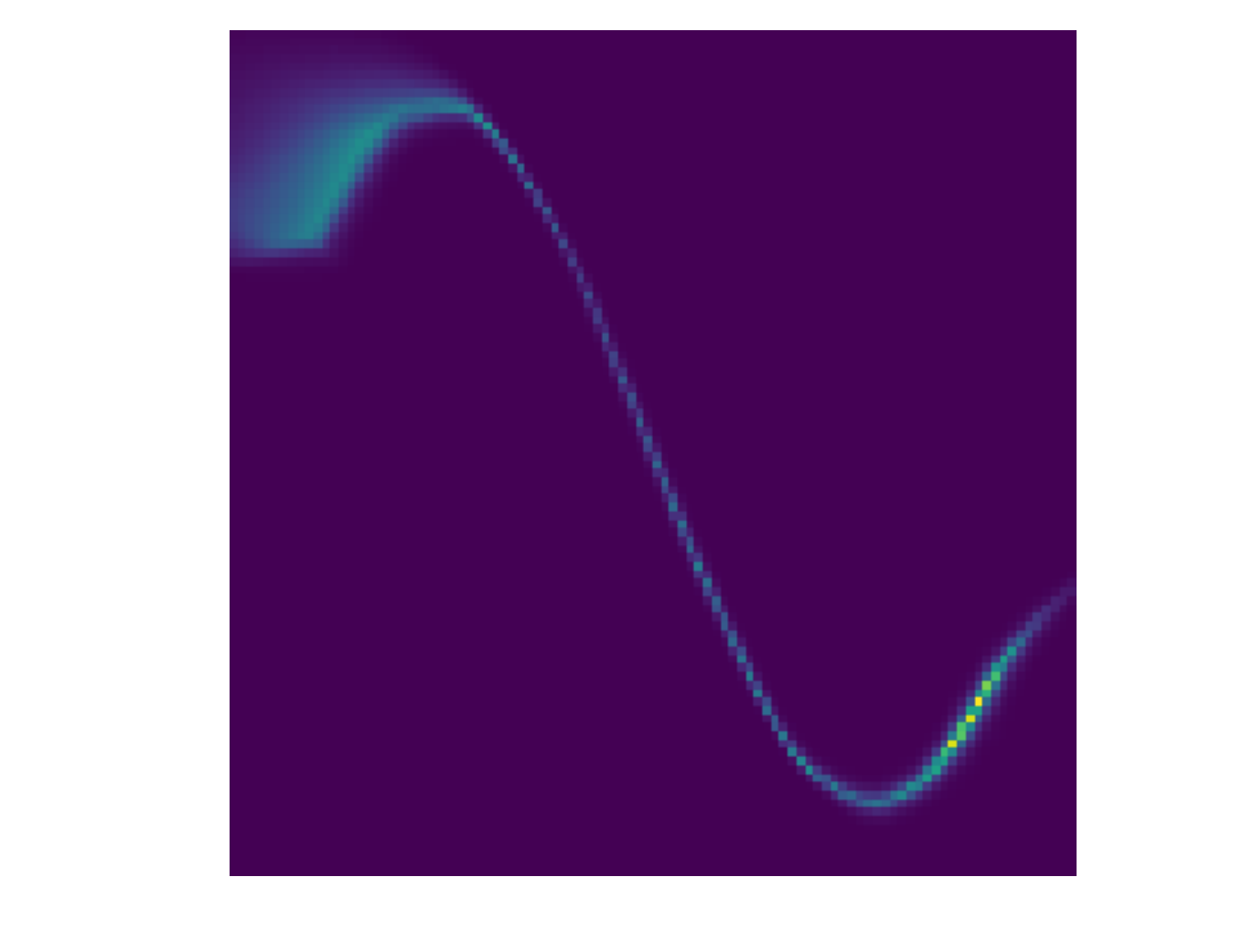}}; &
    \tikz[baseline=(a.north)]\node[yscale=-1,inner sep=0,outer sep=0](a){\includegraphics[width=0.14\textwidth, trim={2cm 0cm 2cm 1cm},clip]{figs/synthetic_v2/Cosine/heat-ema-98-0_1}}; & 
    \tikz[baseline=(a.north)]\node[yscale=-1,inner sep=0,outer sep=0](a){\includegraphics[width=0.14\textwidth, trim={2cm 0cm 2cm 1cm},clip]{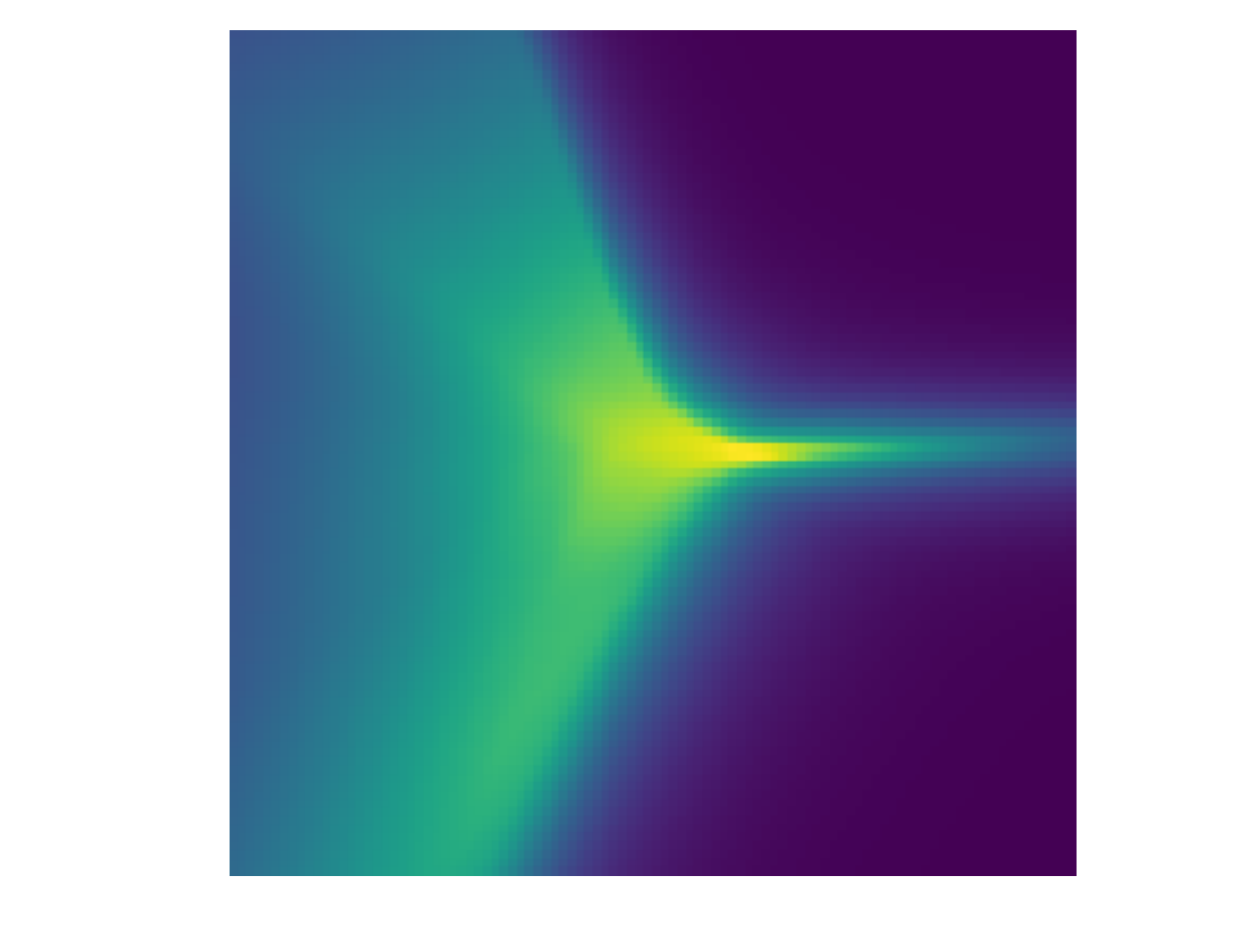}}; &
    \tikz[baseline=(a.north)]\node[yscale=-1,inner sep=0,outer sep=0](a){\includegraphics[width=0.14\textwidth, trim={2cm 0cm 2cm 1cm},clip]{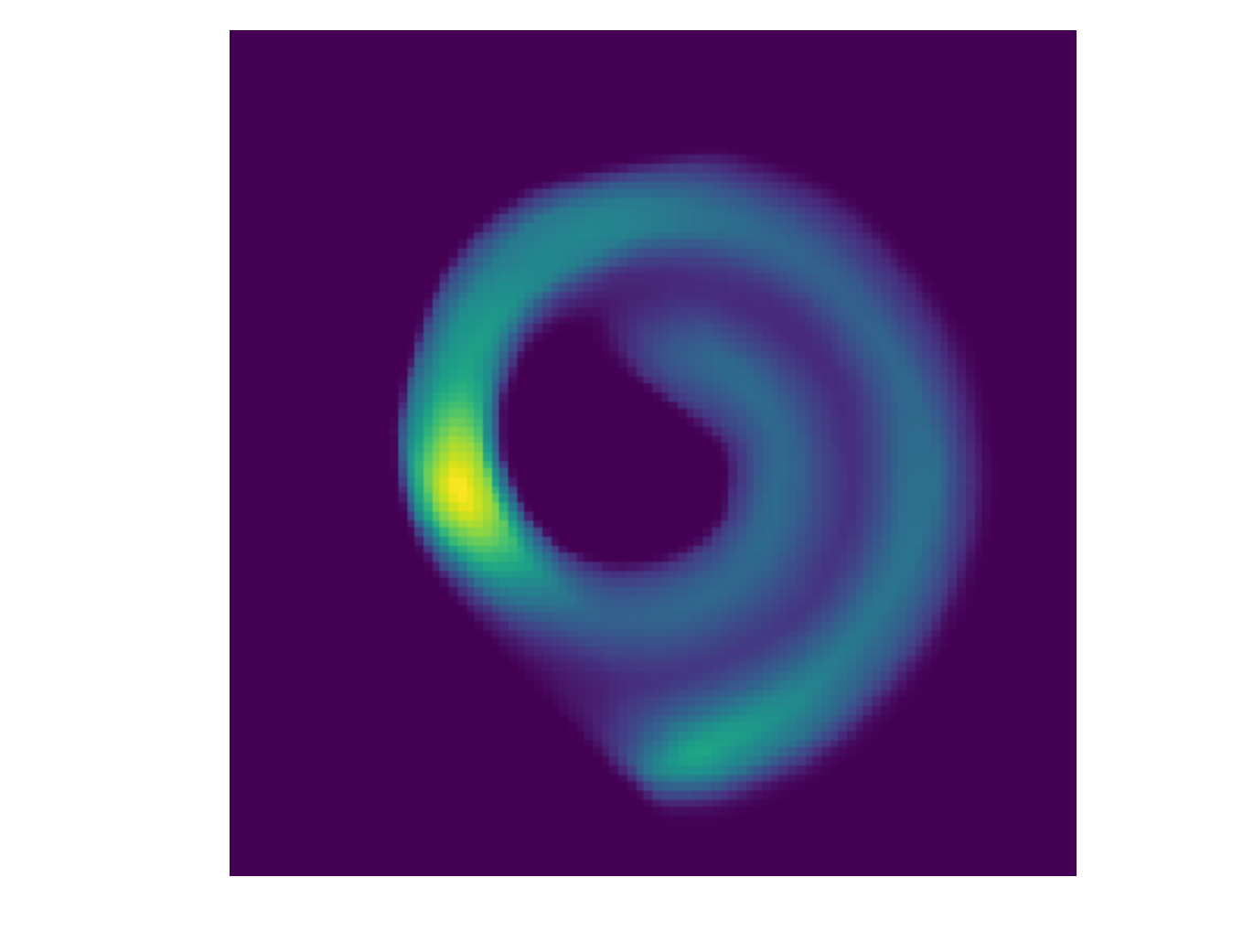}}; \\

    (a) \texttt{2spirals} &(b) \texttt{Banana}  &(c) \texttt{circles} &(d) \texttt{cos} &(e) \texttt{Cosine} &(f) \texttt{Funnel} &(g) \texttt{swissroll} \\

    \includegraphics[width=0.14\textwidth]{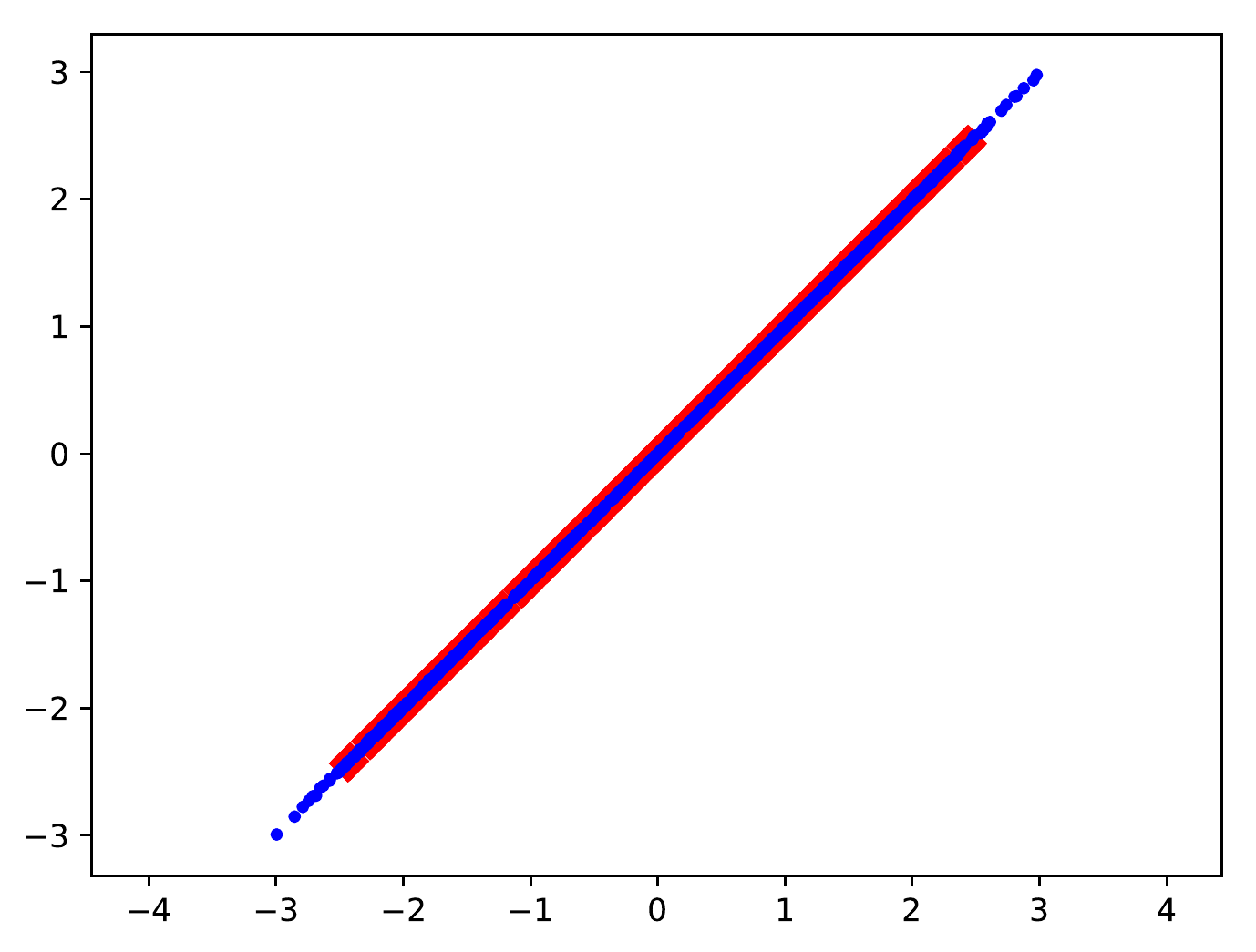} & 
    \includegraphics[width=0.14\textwidth]{figs/synthetic_v2/moons/moons-none} & 
    \includegraphics[width=0.14\textwidth]{figs/synthetic_v2/Multiring/Multiring-none} & 
    \includegraphics[width=0.14\textwidth]{figs/synthetic_v2/pinwheel/pinwheel-none} & 
    \includegraphics[width=0.14\textwidth]{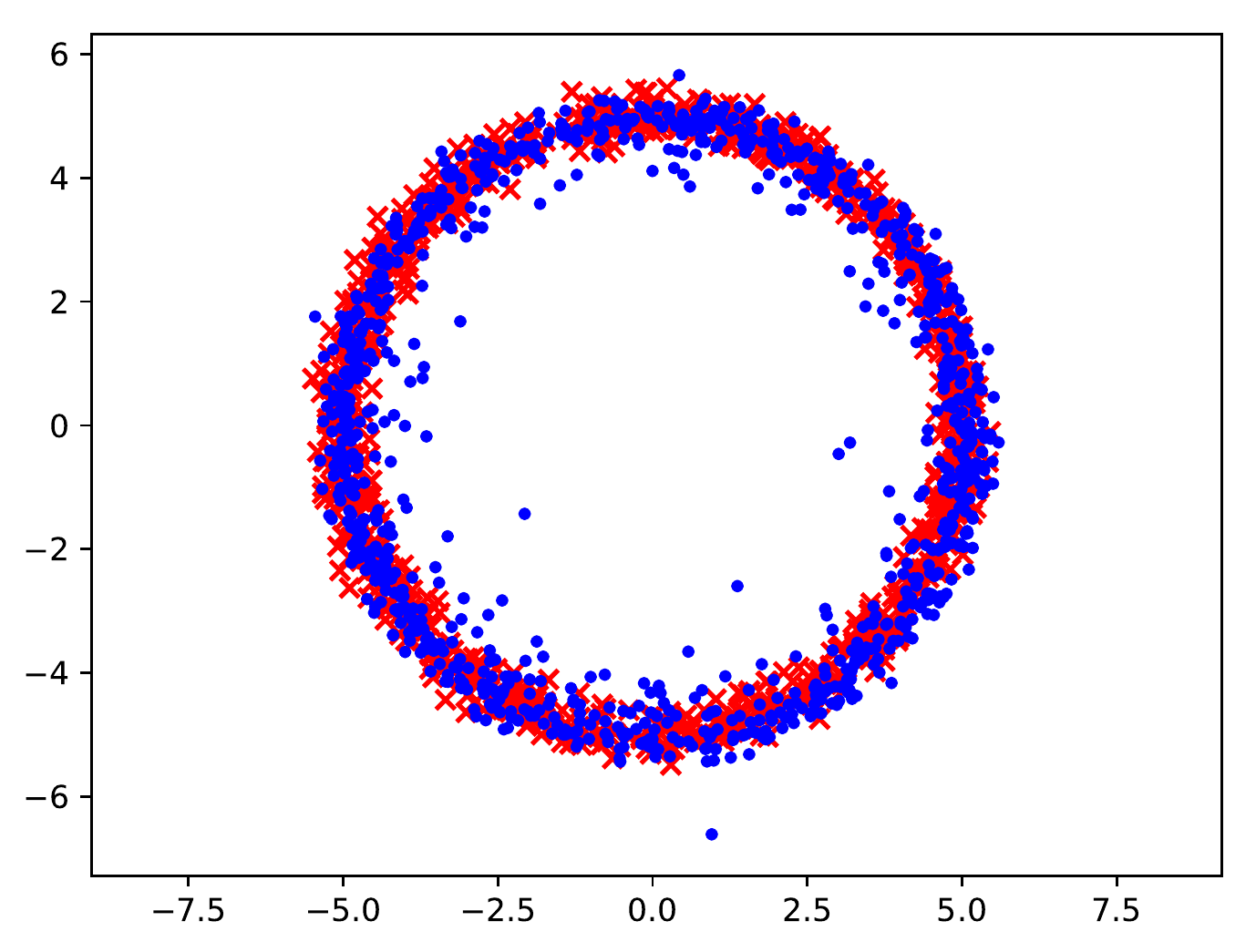} & 
    \includegraphics[width=0.14\textwidth]{figs/synthetic_v2/Spiral/Spiral-none} &
    \includegraphics[width=0.14\textwidth]{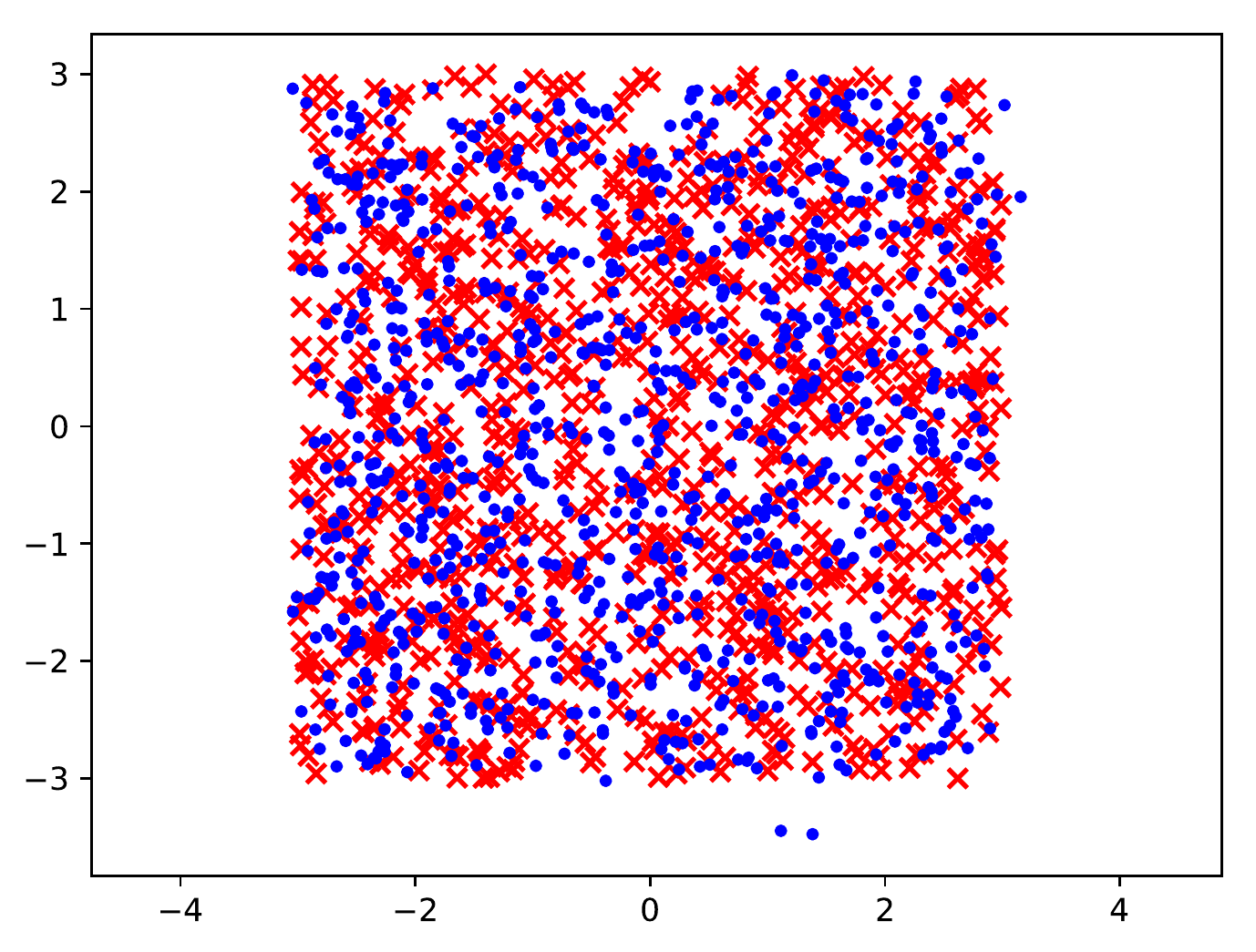} \\
    \tikz[baseline=(a.north)]\node[yscale=-1,inner sep=0,outer sep=0](a){\includegraphics[width=0.14\textwidth, trim={2cm 0cm 2cm 1cm},clip]{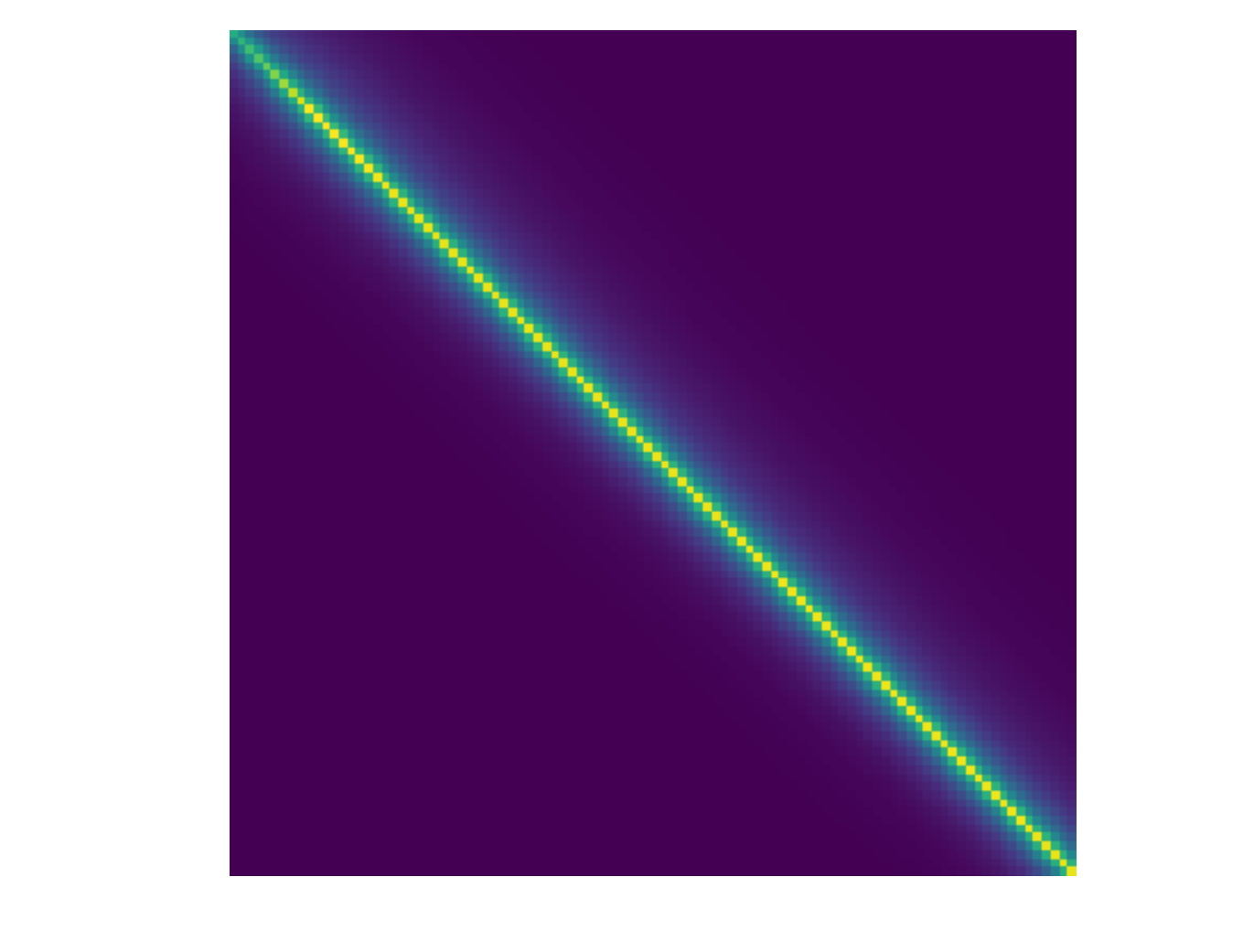}}; & 
    \tikz[baseline=(a.north)]\node[yscale=-1,inner sep=0,outer sep=0](a){\includegraphics[width=0.14\textwidth, trim={2cm 0cm 2cm 1cm},clip]{figs/synthetic_v2/moons/heat-79-0_2}}; &
    \tikz[baseline=(a.north)]\node[yscale=-1,inner sep=0,outer sep=0](a){\includegraphics[width=0.14\textwidth, trim={2cm 0cm 2cm 1cm},clip]{figs/synthetic_v2/Multiring/heat-ema-87-0_2}}; & 
    \tikz[baseline=(a.north)]\node[yscale=-1,inner sep=0,outer sep=0](a){\includegraphics[width=0.14\textwidth, trim={2cm 0cm 2cm 1cm},clip]{figs/synthetic_v2/pinwheel/heat-ema-54-0_01}}; & 
    \tikz[baseline=(a.north)]\node[yscale=-1,inner sep=0,outer sep=0](a){\includegraphics[width=0.14\textwidth, trim={2cm 0cm 2cm 1cm},clip]{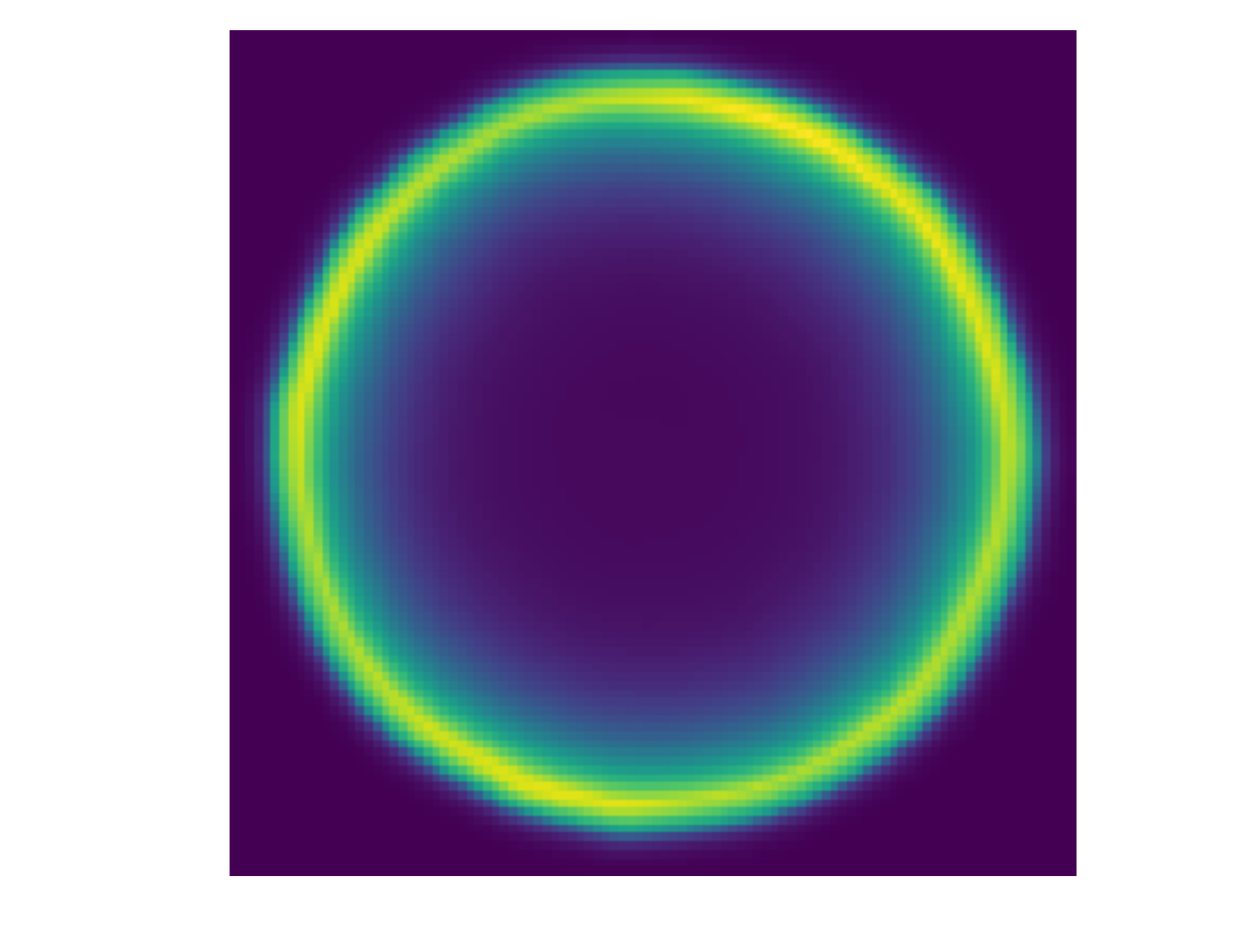}}; & 
    \tikz[baseline=(a.north)]\node[yscale=-1,inner sep=0,outer sep=0](a){\includegraphics[width=0.14\textwidth, trim={2cm 0cm 2cm 1cm},clip]{figs/synthetic_v2/Spiral/heat-ema-69-0_2}}; &
    \tikz[baseline=(a.north)]\node[yscale=-1,inner sep=0,outer sep=0](a){\includegraphics[width=0.14\textwidth, trim={2cm 0cm 2cm 1cm},clip]{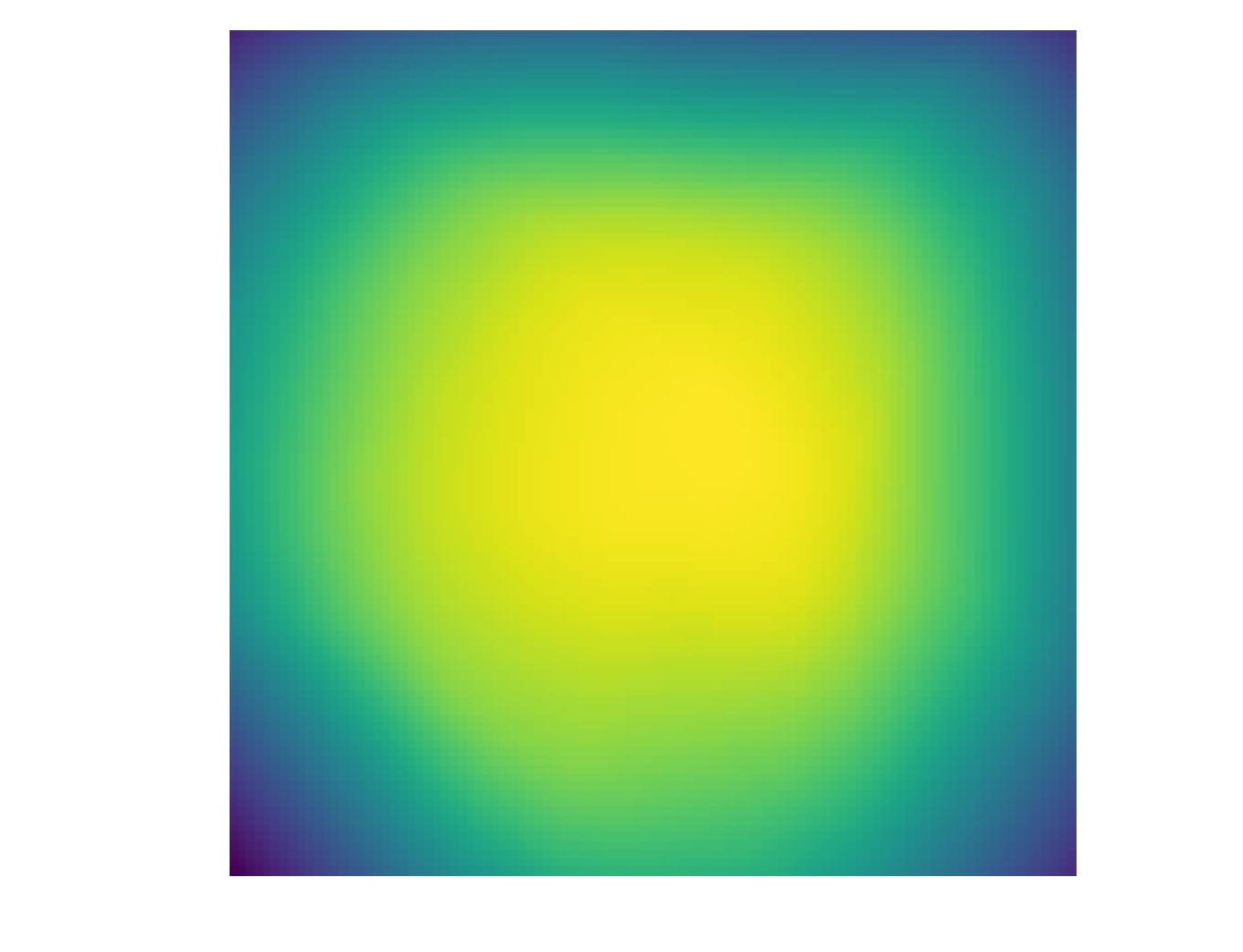}}; \\

    (h) \texttt{line} &(i) \texttt{moons} &(j) \texttt{Multiring} &(k) \texttt{pinwheel} &(l) \texttt{Ring} &(m) \texttt{Spiral} &(n) \texttt{Uniform}
  \end{tabular}
  }
  \caption{Learned samplers in odd row and potential function $f$ in even row from different synthetic datasets. In the sampler illustration in odd rows, the {\color{red}$\times$} denotes training data and {\color{blue}$\bullet$} denotes the ADE samplers. }
  \label{fig:more_synthetic}
\end{figure*}

\paragraph{More results on synthetic datasets} We visualized the learned models and samplers on all the synthetic datasets in~\figref{fig:more_synthetic}. 

\begin{figure}[h!]
\centering
\resizebox{0.8\textwidth}{!}{%
\includegraphics{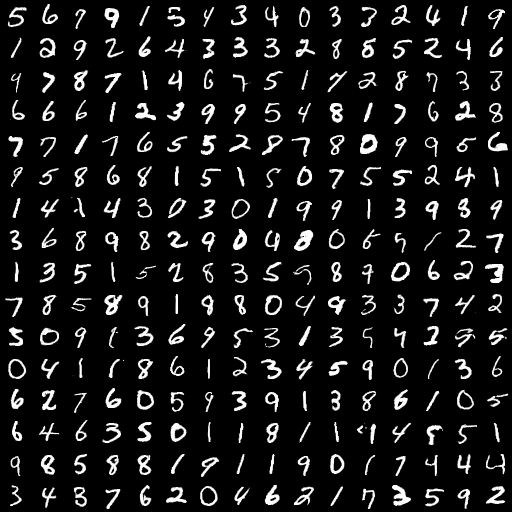} 
}
\caption{Generated images for \texttt{MNIST} by~\algshort.}
\label{fig:more_mnist}
\end{figure}

\begin{figure}[h!]
\centering
\resizebox{0.8\textwidth}{!}{%
\includegraphics{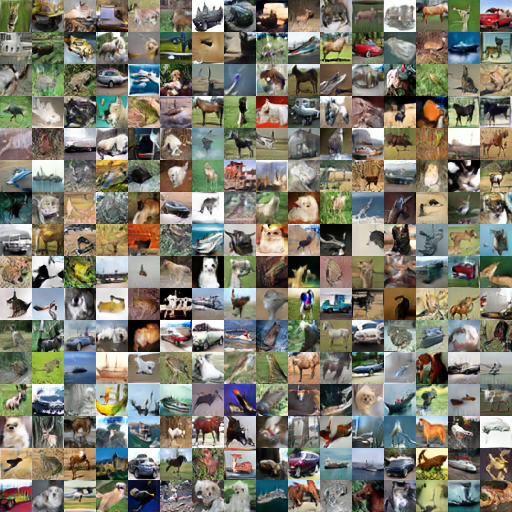} 
}
\caption{Generated images for \texttt{CIFAR-10} by~\algshort.}
\label{fig:more_cifar}
\end{figure}

\paragraph{More results on real-world image generation}
We illustrated additional generated images by the proposed~\algshort~on~\texttt{MNIST} and~\texttt{CIFAR-10} in~\figref{fig:more_mnist} and~\figref{fig:more_cifar}, respectively.

\end{appendix}

\end{document}